%% file: paper.tex
\documentclass{article}
\usepackage[margin=1.02in]{geometry}
\input{preamble.tex}
\usepackage{enumitem}
\setlist{leftmargin=5.5mm}
\setlist[itemize]{noitemsep}

\newcommand{\sm}[1]{{\color{blue}Song:{#1}}}
\newcommand{\yuchen}[1]{{\color{red}Yuchen:{#1}}}

\title{Deep Networks as Denoising Algorithms: Sample-Efficient Learning of Diffusion Models in High-Dimensional Graphical Models}

\date{}
\author{Song Mei\thanks{Department of Statistics, University of California, Berkeley. Email: \tt{songmei@berkeley.edu}.} \and Yuchen Wu\thanks{Department of Statistics and Data Science, University of Pennsylvania. Email: \tt{wuyc14@wharton.upenn.edu}}}

\begin{document}

\maketitle

\begin{abstract}

We investigate the efficiency of deep neural networks for approximating scoring functions in diffusion-based generative modeling. While existing approximation theories leverage the smoothness of score functions, they suffer from the curse of dimensionality for intrinsically high-dimensional data. This limitation is pronounced in graphical models such as Markov random fields, where the approximation efficiency of score functions remains unestablished. 

To address this, we note score functions can often be well-approximated in graphical models through variational inference denoising algorithms. Furthermore, these algorithms can be efficiently represented by neural networks. We demonstrate this through examples, including Ising models, conditional Ising models, restricted Boltzmann machines, and sparse encoding models. Combined with off-the-shelf discretization error bounds for diffusion-based sampling, we provide an efficient sample complexity bound for diffusion-based generative modeling when the score function is learned by deep neural networks. 
\end{abstract}

\tableofcontents

\section{Introduction}

In recent years, diffusion models  \cite{sohl2015deep, ho2020denoising,song2019generative,song2020score} have emerged as a leading approach for generative modeling, achieving state-of-the-art results across diverse domains.  These include breakthroughs in image generation, text generation, and speech synthesis \cite{dhariwal2021diffusion, ramesh2022hierarchical, austin2021structured, popov2021grad}. Diffusion models now power many of the most advanced generative AI systems like DALL·E 2, Stable Diffusion, and Imagen \cite{dalle2, rombach2022high, saharia2022photorealistic}. For overviews of the rapid progress in this exciting new field, see the recent surveys  \cite{yang2022diffusion, croitoru2023diffusion}.


Given a dataset of $n$ independent and identically distributed samples $\{\bx_i \}_{i=1}^n$ drawn from an unknown $d$-dimensional data distribution $\mu \in \cP(\R^d)$, diffusion models aim to learn a generative model that produces new samples $\hat{\bx} \sim \hat{\mu}$ that match this distribution. Popular diffusion models such as DDPM \cite{ho2020denoising} achieve this through a two-step procedure: 
\begin{itemize}
\item {\bf Step 1.} Fit approximate score functions $\hat \bs_t: \R^d \to \R^d$ for $t \in [0, T]$ by minimizing the following empirical risk over a neural network class $\cF$: 
\begin{equation}\label{eqn:score_ERM}\tag{ERM}
\hat \bs_t = \arg\min_{\NN \in \cF} \frac{1}{n} \sum_{i = 1}^n \big\| \sigma_t^{-1} \bg_i+ \NN(\lambda_t \bx_i + \sigma_t \bg_i) \big\|_2^2,~~~~ \bg_i \sim_{iid} \cN(\bzero, \id_d), ~~~ (\lambda_t, \sigma_t^2) = (e^{-t}, 1 - e^{-2t}). 
\end{equation}
\item {\bf Step 2.} Discretize the following stochastic differential equation (SDE) from Gaussian initialization, whose drift term is given by the fitted approximate score functions:  
\begin{equation}\label{eqn:diffusion_SDE}\tag{SDE}
\de \bY_t = \big(\bY_t + 2 \hat \bs_{T - t}(\bY_t)\big) \de t + \sqrt{2} \de \bB_t, ~~~ t \in [0, T], ~~~ \bY_0 \sim \cN(\bzero, \id_d), 
\end{equation}
and take the approximate sample $\hat \bx = \bY_T \in \R^d$.
\end{itemize}
\textit{Score functions} $\bs_t(\bz)$ are central to the diffusion model framework. Given infinite data and model capacity, the minimizer of the empirical risk in Eq.~(\ref{eqn:score_ERM}) yields the score function, 
\begin{equation}\tag{Score}\label{eqn:score}
\begin{aligned}
\bs_t(\bz) =&~ \nabla_{\bz} \log \mu_t(\bz), ~~~~~ \mu_t(\bz): \text{marginal density of $\bz$, when $\bx \sim \mu$ and $[\bz | \bx] \sim \cN( \lambda_t \bx, \sigma_t^2 \id_d ).$ }
\end{aligned}
\end{equation}
Standard analysis shows that replacing the fitted score functions $\hat \bs_t(\bz)$ with the true score functions $\bs_t(\bz)$ in Eq.~(\ref{eqn:diffusion_SDE}), and sending $T \rightarrow \infty$, recovers the original data distribution $\mu$. Therefore, the sample quality from diffusion models relies on two key factors: (1) how well the fitted score functions $\hat \bs_t$ approximate the true score functions $\bs_t$; and (2) how accurately the SDE discretization scheme approximates the continuous process in Eq.~(\ref{eqn:diffusion_SDE}).

Recent work has made substantial progress on controlling the SDE discretization error in diffusion models, assuming access to a good score function estimator \cite{chen2022sampling, chen2023improved, lee2023convergence, li2023towards, benton2023linear}. However, understanding when neural networks can accurately estimate the score function itself remains less explored. Some analyses rely on strong distributional assumptions for score function realizability \cite{shah2023learning, yuan2023reward}, while others exploit the smoothness of score functions, incurring the curse of dimensionality \cite{oko2023diffusion, chen2023score}. These results do not cover many common high-dimensional graphical models for images and text, such as Markov random fields or restricted Boltzmann machines \cite{geman1986markov, ranzato2010factored, conroy2001text}.

\paragraph{A new perspective on score function approximation.} We provide a new perspective on approximating diffusion model score functions with neural networks. First, we observe that by Tweedie's formula, score functions $\bs_t: \R^d \to \R^d$ are related to denoising functions $\bbm_t: \R^d \to \R^d$, which give the posterior expectation of data $\bx$ given noisy observations $\bz$:
\begin{equation}\label{eqn:score-denoiser}\tag{Denoiser}
\bs_t(\bz) = (\lambda_t \cdot \bbm_t(\bz) - \bz) / \sigma_t^2, ~~~~~~\bbm_t(\bz) = \E_{(\bx, \bg) \sim \mu \otimes \cN(\bzero, \id_d)}[\bx | \lambda_t \bx + \sigma_t \bg = \bz ]. 
\end{equation}
Our key insight is that if the data distribution $\mu$ arises from a graphical model, these denoisers $\bbm_t(\bz)$ can often be approximated by variational inference algorithms, which takes the form $\{ \bmf_\ell: \cU_{\ell -1} \to \cU_{\ell}\}_{\ell \in [L]} \cup\{ \bmf_{\inside}: \R^d \to \cU_{0}, \bmf_{\out}: \cU_{L} \to \R^d\}$, 
\begin{equation}\label{eqn:VI}\tag{VI}
\bbm_t(\bz) \approx \bmf_{\out}(\bu^{(L)}),~~~~ \bu^{(\ell)} = \bmf_\ell(\bu^{(\ell-1)}), ~~~~ \ell \in \{ 1, \ldots, L\},~~~~ \bu^{(0)} = \bmf_{\inside}(\bz). 
\end{equation}
For instance, when $\mu$ is an Ising model, the denoising function $\bbm_t$ can be approximated by an iterative algorithm that minimizes a variational inference objective \cite{jordan1999introduction,wainwright2008graphical}. Each update step $\bmf_\ell$ is composed of simple operations, including matrix-vector multiplication and pointwise nonlinearity, which can be captured by a two-layer neural network $\bmf_{\ell}(\bu) \approx \bu + \bW_1 \cdot \relu(\bW_2 \bu)$. This allows efficient approximation of Ising model score functions using a residual network \cite{he2016deep}: By comparing the variational inference updates (\ref{eqn:VI}) and residual network forms (\ref{eqn:relu-resnet}), we can see how the iterative variational inference steps directly translate to residual block approximations. This establishes a clear connection between variational inference in graphical models and score approximation in diffusion models. 



\paragraph{Our contributions.} In this paper, we analyze several examples of high-dimensional graphical models, including Ising models (Section~\ref{sec:Ising-new}), latent variable Ising models (Section~\ref{sec:latent-Ising}), conditional Ising models (Section~\ref{sec:conditional-Ising}), and sparse coding models (Section~\ref{sec:SC}). Assuming the denoising function can be approximated by the minimizer of a variational inference objective, verified in our examples, we derive an efficient score estimation error bound. This modular score estimation result is a valuable step toward sampling guarantees for diverse diffusion sampling schemes. For instance, combined with existing discretization error analyses for the DDPM sampling scheme, we provide a sample complexity guarantee for diffusion-based generative modeling and conditional generative modeling using neural network-based score learning.

\section{Preliminaries: the DDPM sampling scheme}\label{sec:prelim}

\begin{algorithm}
\caption{The DDPM sampling scheme}\label{alg:sampling-resnet}
\begin{algorithmic}[1]
\REQUIRE Samples $\{ \bx_i \}_{i \in [n]} \subseteq \R^d$. ResNet parameters $(d, D, L, M, B)$. Discretization scheme parameters $(N, T, \delta, \{ t_k \}_{0 \le k \le N})$ with $0 = t_0 < \cdots < t_{N} = T - \delta$. Denote $\gamma_k = t_{k+1} - t_k$. 
\STATE {\color{blue} // Computing the approximate score function}
\STATE Sample $\{\bg_i \}_{i \in [n]} \sim_{iid} \cN(\bzero, \id_d)$. 
\FOR{$t \in \{ T - t_k \}_{0 \le k \le N - 1}$ }
\STATE Solve the ERM problem below for $t = T -  t_k$: 
\begin{equation}\label{eqn:ERM-in-alg}
\what \bW_t = \arg\min_{\bW \in \cW_{d, D, L, M, B} } \frac{1}{n} \sum_{i = 1}^n \Big\| \sigma_t^{-1} \bg_i  + \sP_{t} [ \resnet_{\bW}](\lambda_t \bx_i + \sigma_t \bg_i)\Big\|_2^2,~~~ (\lambda_t, \sigma_t^2) = 
(e^{-t}, 1 - e^{-2t}). 
\end{equation}
\STATE Take the approximate score function to be $\hat \bs_t(\bz) = \sP_t[ \resnet_{\what \bW_t}] (\bz)$.  
\ENDFOR
\STATE {\color{blue} // Sampling by discretizing the stochastic differential equation}
\STATE  Sample $\what\bY_0 \sim \cN(\bzero, \id_d)$. 
\FOR{$k = 0, \cdots, N - 1$}
\STATE Sample $\bG_k \sim \cN(\bzero, \id_d)$. Calculate $\what \bY_{k+1}$ using the exponential integrator scheme:  
\begin{equation}\label{eqn:exponential-integrator-discrete}
\what \bY_{k + 1} = e^{\gamma_k} \cdot \what\bY_{k} + 2 (e^{\gamma_k} - 1) \cdot  \hat \bs_{T - t_k}(\what\bY_{k})  + \sqrt{e^{2\gamma_k} - 1} \cdot \bG_k. 
\end{equation}
\ENDFOR
\end{algorithmic}
{\bf Return}: $\hat \bx = \what \bY_N$.
\end{algorithm}


This section provides details on the two-step DDPM sampling scheme in Algorithm \ref{alg:sampling-resnet}. The inputs of the algorithm are $n$ IID samples $\{ \bx_i \}_{i \in [n]} \subseteq \R^d$ from a distribution $\mu$. The algorithm also receives parameters $(d, D, L, M, B)$ for specifying the ResNet class, and $(N, T, \delta, \{ t_k \}_{0 \le k \le N} )$ for specifying the time discretization scheme. The first step of the algorithm performs empirical risk minimization to compute the approximate score functions $\hat \bs_t$ (lines 2-5). The second step generates a sample by discretizing the reverse-time SDE using the fitted score functions (lines 7-9). We discuss the score learning and SDE discretization steps in more detail below.

\paragraph{Empirical risk minimization and the ResNet class.} The first step of Algorithm \ref{alg:sampling-resnet} solves an empirical risk minimization problem (\ref{eqn:ERM-in-alg}) to fit the score functions. This regresses manually-generated standard Gaussian noises $\{ \bg_i \}_{i \in [n]}$ on the noisy samples $\{ \lambda_t \bx_i + \sigma_t \bg_i \}_{i \in [n]}$, for $(\lambda_t, \sigma_t^2) = (e^{-t}, 1 - e^{-2t})$, using a standard residual network architecture $\ResN_{\bW}: \R^d \to \R^d$. The ResNet is parameterized by a set of weight matrices $\bW = \{\bW_1^{(\ell)} \in \R^{D \times M}, \bW_2^{(\ell)} \in \R^{M \times D}\}_{\ell \in [L]} \cup \{\bW_{\inside} \in \R^{(d + 1) \times D}, \bW_{\out} \in \R^{D \times d}\}$ with embedding dimension $D$, number of layers $L$, and hidden-layer width $M$. It applies iterative residual blocks with ReLU nonlinearities ($\relu(x) = x\cdot 1\{ x > 0\}$) to map an input $\bz$ to an output in $\R^d$: 
\begin{equation}\label{eqn:relu-resnet}\tag{ResNet}
\resnet_{\bW}(\bz) = \bW_{\out} \bu^{(L)},~~~~ \bu^{(\ell)} = \bu^{(\ell - 1)} + \bW_1^{(\ell)} \relu(\bW_2^{(\ell)} \bu^{(\ell - 1)}), ~~~~ \bu^{(0)} = \bW_{\inside} [\bz; 1]. 
\end{equation}
The minimization in (\ref{eqn:ERM-in-alg}) is over the ResNets whose weights are contained in a $B$-bounded set, specified by parameters $(d, D, L, M, B)$ 
\begin{align}\label{eqn:ResNet-class}
\cW_{d, D, L, M, B} := \Big\{ \bW = \{ \bW_1^{(\ell)}, \bW_2^{(\ell)} \}_{\ell \in [L]} \cup \{\bW_{\inside}, \bW_{\out} \}:  \nrmps{\bW} \leq B \Big\}.
\end{align}
Here the norm of ResNet weights is defined as 
\begin{align}
\nrmps{\bW} := \max_{\ell \in [L]} \big\{ \|\bW_1^{(\ell)}\|_{\op} + \|\bW_2^{(\ell)}\|_{\op} \big\} \vee \max\big\{ \|\bW_{\inside}\|_{\op}, \|\bW_{\out}\|_{\op}  \big\}. 
\end{align}

For technical reasons, we truncate the ResNet output using the operator $\sP_{t}$, which constrains the norm of the approximated denoising function. Given a function $f: \R^d \to \R^d$, we define $\sP_{t}[f] (\bz) = \proj_{\lambda_t \sigma_t^{-2} \sqrt{d}}(f(\bz) + \sigma_t^{-2} \bz ) - \sigma_t^{-2} \bz$, where $\proj_R(\bz) = \bz \cdot \ones\{ \| \bz \|_2 \le R\} + R \cdot ( \bz / \| \bz \|_2) \cdot  \ones\{ \| \bz \|_2 > R\}$ is the projector of $\bz \in \R^d$ into the $R$-Euclidean ball. Note that when $f(\bz)$ is a score function, $f(\bz) + \sigma_t^{-2} \bz$ is a rescaled denoising function and should be bounded for data distribution with compact support. This operator is a technical detail that could be eliminated in practice --- it is only used to control the generalization error of the empirical risk minimization problem.

\paragraph{Choice of the discretization scheme.} The second step of Algorithm \ref{alg:sampling-resnet} discretizes the backward SDE using the exponential integrator scheme. While multiple options exist for the time discretization $\{ t_k \}_{0 \le k \le N}$, we choose a particular scheme that uses a uniform grid in the first phase and an exponential decaying grid in the second phase. As shown in \cite{benton2023linear}, such a scheme provides a sharp sampling error control. 

\begin{definition}[Two-phase discretization scheme \cite{benton2023linear}]\label{def:two-phase-discretization-scheme}
The two-phase discretization scheme has parameters $(\kappa, N_0, N, T, \delta) \in (0, 1) \times \N \times \N \times \R \times (0, 1)$, where $(\kappa, N_0, N)$ are free parameters and $(T, \delta)$ are fully determined by $(\kappa, N_0, N)$. In the first uniform phase, the $N_0$ time steps have equal length $\kappa$. In the second exponential phase, the $N - N_0$ steps decay with rate $1/(1+\kappa) \in (0,1)$. The last time step $t_N$ has a gap $\delta =  (1 + \kappa)^{N_0 - N} \in (0, 1)$ to $T$. 

Specifically, we take $t_0 = 0$, $t_k = k \kappa$ for $k \le N_0$, $t_{N_0} = N_0 \kappa = T - 1$, $t_{N_0 + k} = T - (1 + \kappa)^{-k}$ for $0\le k \le N - N_0$, and $t_{N} = T - (1 + \kappa)^{N_0 - N} = T - \delta$. Defining $\gamma_k = t_{k + 1} - t_k$, we have $\gamma_k = \kappa$ for $k \le N_0 - 1$, and $\gamma_
{N_0 + k} = \kappa / (1 + \kappa)^{k+1}$ for $0 \le k \le N - N_0 - 1$. See \cite[Figure 1]{benton2023linear} for a pictorial illustration of this scheme. 
\end{definition}

\paragraph{The conditional diffusion model.} In conditional generative modeling tasks, we observe IID samples $\{ (\bx_i, \btheta_i) \}_{i \in [n]} \sim_{iid} \mu$, and our goal is to learn a model to generate new samples $\hat \bx$ from the conditional distribution $\mu(\bx | \btheta)$ for a given $\btheta$. 

The DDPM sampling scheme can be simply adapted to solve conditional generative modeling tasks, as per Algorithm \ref{alg:sampling-resnet-conditional}. Specifically, we modify the ResNet in empirical risk minimization to take the form (\ref{eqn:relu-resnet-conditional}), admitting inputs $(\lambda_t \bx_i + \sigma_t \bg_i, \btheta_i) \in \R^d \times \R^m$. The approximated score functions $\hat \bs_t (\bz)$ become conditional $\hat \bs_t (\bz; \btheta) = \sP_t[\ResN_{\what \bW_t}](\bz, \btheta)$, estimating the conditional score functions $\bs_t(\bz; \btheta) = \nabla_{\bz}\log \mu_t(\bz, \btheta)$, where $\mu_t$ is the joint density of $(\bz, \btheta)$ when $(\bx, \btheta) \sim \mu$ and $[\bz | \bx] \sim \cN( \lambda_t \bx, \sigma_t^2 \id_d)$. Details of the conditional algorithm are provided in Appendix \ref{app:ddpm-conditional}.

\section{Diffusion models for Ising models}\label{sec:Ising-new}

The Ising model $\mu \in \cP(\{ \pm 1\}^d)$ is a distribution over the discrete hypercube, with probability mass function characterized by an energy function of spin configurations. Specifically,
\begin{equation}\label{eqn:Ising}\tag{Ising}
\textstyle \mu(\bx) = Z^{-1}\exp\{  \< \bx, \bA \bx\> / 2 \},~~~~ \bx \in \{ \pm 1\}^d,~~~~ Z = \sum_{\bx \in \{ \pm 1\}^d} \exp\{ \< \bx, \bA \bx\> / 2 \}. 
\end{equation}
The Ising model stands as one of the most fundamental graphical models; it belongs to the exponential family, yet its normalizing constant, $Z$, does not possess an analytic expression. Computational complexities in parameter estimation and sampling from Ising models have remained a focal point of research for several decades \cite{el2022sampling, mezard2009information, jerrum1993polynomial}. A notable variant of the Ising model is the Sherrington-Kirkpatrick (SK) model \cite{sherrington1975solvable}, where $\bA$ is a Gaussian random matrix. This model has attracted extensive research due to its ties with statistical physics and high-dimensional statistics \cite{mezard2009information, wainwright2008graphical}.

Consider the task of generative modeling where the input consists of IID samples $\{ \bx_i \}_{i \in [n]} \sim \mu$ derived from the Ising model. To demonstrate that the DDPM sampling scheme (Algorithm \ref{alg:sampling-resnet}) provides a sample $\hat \bx \sim \hat \mu$ with $\hat \mu \approx \mu$, we need to control the score estimation error $\E[\| \hat \bs_{t}(\bz) - \bs_t(\bz) \|_2^2]$ \cite{benton2023linear}. Here,  $\hat \bs_t$ is the trained ResNet and $\bs_t(\bz)$ relates to the denoiser $\bbm_t(\bz) = \E_{(\bx, \bg) \sim \mu \otimes \cN(\bzero, \id_d)}[\bx | \lambda_t \bx + \sigma_t \bg = \bz ]$ per (\ref{eqn:score-denoiser}). To calculate $\bbm_t(\bz)$, a naive variational Bayes approximation \cite{wainwright2008graphical} suggests using the minimizer of the naive variational Bayes (VB) free energy $\cF^{\rm naive}_t$: 
\[
\begin{aligned}
\cF^{\rm naive}_t(\bbm; \bz) =&~ \sum_{i = 1}^d -\sh_{\rm bin}(m_i) - \frac{1}{2} \< \bbm, \bA \bbm\>  - \frac{\lambda_t}{\sigma_t^2} \< \bz, \bbm\>, \\
- \sh_{\rm bin}(m) =&~ \Big[ \frac{1 + m}{2} \log\Big(\frac{1 + m}{2} \Big) + \frac{1 - m}{2} \log\Big(\frac{1 - m}{2} \Big) \Big]. 
\end{aligned}
\]
However, in many cases, a correction to the VB free energy is needed for its minimizer to be consistent with the denoiser $\bbm_t(\bz)$ \cite{thouless1977solution, ghorbani2019instability, fan2021tap, fan2022tap}. To establish our main result, we will assume the consistency of a free energy minimizer with the denoiser. 

\begin{assumption}[Consistency of the free energy minimizer in Ising models]\label{ass:Ising-free-energy}
Let $\bx \sim \mu(\bsigma) \propto \exp\{ \< \bsigma, \bA \bsigma\> / 2 \}$ and $\bz \sim \cN( \lambda_t \bx, \sigma_t^2 \id_d)$. Denote the marginal distribution of $\bz$ by $\mu_t$. For any fixed $t$, assume that there exists $\eps_{{\rm VI}, t}^2(\bA) < \infty$ and $\bK = \bK(\bA, t)$ with $\| \bK - \bA \|_{\op} \le A < 1$, such that
\[
\begin{aligned}
&~ \E_{\bz \sim \mu_t} [\| \hat \bbm_t(\bz) - \bbm_t(\bz) \|_2^2 ] / d \le \eps_{{\rm VI}, t}^2(\bA),\\
&~ \hat \bbm_t(\bz) = \argmin_{\bbm \in [-1, 1]^d} \Big\{ \cF_t^{\rm VI}(\bbm; \bz) = \sum_{i = 1}^d -\sh_{\rm bin}(m_i) - \frac{1}{2} \< \bbm, \bA \bbm\>  - \frac{\lambda_t}{\sigma_t^2} \< \bz, \bbm\> + \frac{1}{2} \< \bbm, \bK \bbm\>\Big\}. \\
\end{aligned}
\]
\end{assumption}

In our assumption, the free energy  $\cF_t^{\rm VI}(\bbm; \bz) = \cF_t^{\rm naive}(\bbm; \bz) + \< \bbm, \bK \bbm\>/2$ includes a correction term $\< \bbm, \bK \bbm\>/2$ added to the naive VB free energy. In the statistical physics and variational inference literature, this correction term is often called the Onsager's reaction term, and the resulting VI free energy is often called the Thouless-Anderson-Palmer (TAP) free energy \cite{thouless1977solution}. In Section \ref{sec:Ising-examples}, we will discuss cases in which the VI approximation error $\eps_{{\rm VI}, t}^2(\bA)$ can be well-controlled. 

Given Assumption~\ref{ass:Ising-free-energy} holds, we are ready to provide a bound on the estimation error of the approximate score function. We give a proof outline in Section~\ref{sec:proof-outline-Ising} and the full proof in Appendix~\ref{app:Ising-new}.


\begin{theorem}\label{thm:Ising-score-approximation}
Let Assumption~\ref{ass:Ising-free-energy} hold. Let $\{ \hat \bs_{T - t_k} \}_{0 \le k \le N - 1}$ be the approximate score function given by Algorithm \ref{alg:sampling-resnet} in which we take
\[
D = 3d, ~~~~~ M \ge 4 d,~~~~~ B \ge 7 \cdot (M/d) \cdot \log (M) + 1 / \min_{k} 
\{ T - t_k \} + \sqrt{d}. 
\]
Then with probability at least $1 - \eta$, for any $t \in \{ T - t_k\}_{0 \le k \le N-1}$, we have
\begin{equation}\label{eqn:score-estimation-error-Ising}
\begin{aligned}
&~\E_{\bz \sim \mu_t}[\|  \hat \bs_t(\bz) - \bs_t(\bz) \|_2^2]/d
\lesssim \lambda_t^2 \sigma_t^{-4} \cdot \Big( \eps_{{\rm VI}, t}^2(\bA) + \eps^2_{{\rm ResN}}  + \eps^2_{{\rm gen}} \Big),
\end{aligned}
\end{equation}
where 
\begin{equation}\label{eqn:VI-ResN-gen-bound-Ising}
\begin{aligned}
\eps^2_{{\rm ResN}} =  \frac{d^2}{M^2(1 - A)^2} + A^{2L},~~~~
\eps^2_{{\rm gen}} =&~ \sqrt{\frac{ (M d L + d^2) [ T + L \log (B L) ] + \log(N/\eta)}{n}}.
\end{aligned}
\end{equation}
\end{theorem}

Theorem~\ref{thm:Ising-score-approximation} shows the score estimation error can be bounded by three terms: $\eps_{\rm VI}^2$, $\eps^2_{\rm ResN}$, and $\eps^2_{\rm gen}$. The first term $\eps_{\rm VI}^2$ is the error from Assumption~\ref{ass:Ising-free-energy}, independent of the ResNet architecture and sample size $n$, and often small when dimension $d$ is large. We will discuss this term in more detail in Section \ref{sec:Ising-examples}. The second term $\eps^2_{\rm ResN}$ is the ResNet approximation error that vanishes as depth $L$ and width $M$ increase. The third term $\eps^2_{\rm gen}$ is the generalization error controlled for sample size $n$ polynomial in $(d, M, L, T, \log B, \log (N / \eta))$. We remark that our generalization error bound $\eps_{\rm gen}^2$ may not be tight; localization techniques could possibly yield a $1/n$ convergence rate \cite{bartlett2002localized, wainwright_2019}. 

Combining Theorem~\ref{thm:Ising-score-approximation} with off-the-shelf results on the DDPM discretization error \cite{benton2023linear}, we obtain the following bound on the sampling error in terms of KL divergence: 
\begin{corollary}\label{cor:KL-sampling-bound-Ising}
Let Assumption~\ref{ass:Ising-free-energy} hold. Consider the two-phase discretization scheme as in Definition \ref{def:two-phase-discretization-scheme}. Denote the distribution of the output of Algorithm \ref{alg:sampling-resnet} as $\hat \mu$. Then, with probability at least $1 - \eta$, we have
\begin{equation}\label{eqn:KL-cor-Ising}
{\rm KL}( \mu_\delta, \hat \mu ) / d \lesssim  \eps_{\rm score}^2  + \eps_{\rm disc}^2, 
\end{equation}
where 
\begin{equation}\label{eqn:score-disc-bound-cor-Ising}
\eps_{\rm score}^2 \le \delta^{-1} \cdot \Big( \sup_{0 \le k \le N-1} \eps_{{\rm VI}, T - t_k}^2 + \eps^2_{{\rm ResN}}  + \eps^2_{{\rm gen}} \Big), ~~~~\eps_{\rm disc}^2 \le \kappa^2 N + \kappa T + e^{- 2 T}.
\end{equation}
\end{corollary}

Equation (\ref{eqn:KL-cor-Ising}) provides control on the KL divergence between $\mu_\delta$ and $\hat{\mu}$ normalized by dimension $d$. If the right-hand side is small, this guarantees the two distributions are close in an average per-coordinate sense: for two $d$-dimensional product distributions $\mu = \cN(0, 1)^{\otimes d}$ and $\nu = \cN(\eps, 1)^{\otimes d}$ that are close per coordinate, their KL divergence scales as ${\rm KL}(\mu, \nu) \asymp d \cdot \eps^2$, growing linearly with $d$.  Furthermore, it is possible to derive bounds on the distance between the original distribution $\mu$ (instead of $\mu_\delta$) and the learned distribution $\hat{\mu}$ using other DDPM discretization analyses such as \cite{chen2022sampling, chen2023improved, li2023towards}.

\subsection{Verifying the assumption in examples}\label{sec:Ising-examples}

This section provides examples that admit controlled VI approximation error $\eps_{{\rm VI}, t}^2$. The results in this section are proved in Appendix~\ref{app:Ising-examples}.

\paragraph{Ising model in the VB consistency regime.} There is a line of work studying the consistency of the naive mean-field variational Bayes (VB) free energy in Ising models under high-temperature conditions \cite{chatterjee2016nonlinear, eldan2018gaussian, jain2018mean, mukherjee2022variational}. We build on this by providing a quantitative bound on the variational inference approximation error for a general coupling matrix $\bA$ in this regime.

\begin{lemma}\label{lem:Ising-VB-consistency}
Assume $\| \bA \|_{\op} < 1/2$. Then for any $t$, Assumption~\ref{ass:Ising-free-energy} is satisfied for $\bK = \bzero$, and 
\begin{equation}\label{eqn:VI-error-bound-VB}
\eps_{{\rm VI}, t}^2( \bA) \le \frac{4}{1 - 2 \| \bA \|_{\op}} \frac{\| \bA \|_F^2}{ d}. 
\end{equation}
\end{lemma}

As an example, for the ferromagnetic Ising model we have $\bA = \beta \ones \ones^\sT / d$, giving $\eps_{{\rm VI}, t}^2( \bA) \le [4\beta^2 /(1 - 2 \beta)] / d$. This shows the VI approximation error vanishes as $\beta < 1/2$ and $d \to \infty$. However, this is not a particularly interesting regime for Ising models, since they can be well-approximated by a product distribution when $\beta$ is small \cite{chatterjee2016nonlinear, eldan2018gaussian}.

\paragraph{The Sherrington-Kirkpatrick model.} The Sherrington-Kirkpatrick model assumes $\bA = \beta \bJ$, where $\bJ \sim \GOE(d)$ is a symmetric Gaussian random matrix with off-diagonal entries that are IID Gaussian with variance $1/d$. Prior work has shown that the VB free energy does not provide consistent estimation in this model \cite{ghorbani2019instability,fan2021tap}. Instead, the variational objective that yields a consistent estimator of the Gibbs mean is the Thouless-Anderson-Palmer (TAP) free energy \cite{thouless1977solution,fan2021tap,el2022sampling}. Using results on the TAP free energy, the variational inference (VI) approximation error can be controlled for this model when $\beta < 1/4$. 

\begin{lemma}\label{lem:SK-TAP-consistency}[Corollary of Lemma 4.10 of \cite{el2022sampling}]
Assume $\bA = \beta \bJ$ where $\bJ \sim \GOE(d)$ and $\beta < 1/4$. Then for any $t$, there exists matrices $\bK = c_t \id_d$ for some $c_t$, such that with high probability, $\| \bA - c_t \id_d \|_{\op} \le A < 1$ and
\[
\eps_{{ \rm VI}, t}(\beta \bJ) \gotop 0,~~~~ \text{ as }d \to \infty.
\]
\end{lemma}

Lemma~\ref{lem:SK-TAP-consistency} provides a qualitative result on the consistency of variational inference (VI) for the Sherrington-Kirkpatrick model, but does not give a non-asymptotic error bound. To establish a non-asymptotic guarantee, one could potentially leverage tools like the smart path method \cite[Theorem 2.4.20]{talagrand2003spin} or Stein's method \cite{chatterjee2010spin}. We conjecture it is possible to prove a quantitative error bound of order $C(\beta)/d$ using these techniques, as illustrated in \cite[Theorem 2.4.20]{talagrand2003spin}. 

\paragraph{Other Ising models.} We conjecture that Lemma~\ref{lem:SK-TAP-consistency} could extend to a variety of other models including non-Gaussian Wigner matrices \cite{carmona2006universality}, heterogeneous variances \cite{wu2023thouless}, orthogonally invariant spin glasses \cite{fan2022tap}, and spiked matrix models with non-Rademacher priors \cite{fan2021tap, lelarge2019fundamental}: 
\begin{itemize}
\item {\it Non-Gaussian Wigner matrices.} We have $\bA = \beta \bJ$ where $\bJ$ is a symmetric random matrix whose off-diagonal elements are independent with variance $1/d$ and satisfy some moment condition. This generalizes GOE matrices to non-Gaussian distributions. Since these matrices have similar properties to GOE matrices \cite{carmona2006universality}, we conjecture Lemma~\ref{lem:SK-TAP-consistency} should hold. 
\item {\it Heterogeneous variance: multi-species Sherrington-Kirkpatrick models.} We have $\bA = \beta \bJ$ where $\bJ$ is a random matrix with independent entries but heterogeneous variance. An example is the bipartite Sherrington-Kirkpatrick model specified by a set $S \subseteq [d]$, with $J_{ij} = J_{ji} \sim \cN(0, 1/d)$ for $i \in S$ and $j \in S^c$, and $J_{ij} = 0$ for $i, j \in S$ or $i, j \in S^c$. The TAP equations verifying Assumption~\ref{ass:Ising-free-energy} has been shown to hold in similar models \cite{wu2023thouless} in the high-temperature regime $\beta \le \beta_0$. 
\item {\it Orthogonally invariant spin glass models.} We have $\bA = \beta \bJ$, where $\bJ  = \bO \bE \bO^\sT \in \R^{d \times d}$. Here, $\bO \sim {\rm Haar}({\rm SO}(d))$ is a uniform random orthogonal matrix and $\bE = \diag(e_1, \ldots, e_d) \in \R^{d \times d}$ is a diagonal matrix. The TAP equations have been shown for related models \cite{fan2022tap} in the high-temperature regime. 
\item {\it Spiked matrix models.} Suppose we observe $\bY = \bu \bu^\sT + \bJ$ where $\bJ \sim \GOE(d)$ and $\bu \in \R^{d}$ with $u_i \sim_{iid} \pi_0$ for some distribution $\pi_0 \in \cP(\R)$. The posterior distribution of $\bu$ given observation $\bY$ is given by $\mu(\bx) \propto \exp\{ \< \bx, \bY \bx\>/ 2 - \| \bx \|_2^4/(4n) \} \pi_0^d(\bx)$. Taking this $\mu$ as the sample distribution, we conjecture that Assumption~\ref{ass:Ising-free-energy} can be verified for this model \cite{fan2021tap, lelarge2019fundamental}. 
\end{itemize}

\subsection{Discussions}

\paragraph{More explicit sample complexity bounds.} Corollary~\ref{cor:KL-sampling-bound-Ising} provides a sampling error bound in terms of the KL divergence of $\mu_\delta$ and $\hat \mu$. To interpret this bound, assume $\hat \mu$ satisfies a dimension-free transportation-information inequality, i.e., $W_1^2(\mu_\delta, \hat \mu) \lesssim {\rm KL}(\mu_\delta, \hat \mu)$. Further assume $\sup_{t} \eps_{{\rm VI}, t}^2(\bA) \lesssim 1/ d$ (conjectured to hold for the SK model when $\beta < 1$). Since $W_1^2(\mu_\delta, \mu) / d \lesssim \delta$, this implies
\[
W_1^2(\mu, \hat \mu) /d \lesssim W_1^2(\mu, \mu_\delta)/ d + {\rm KL}(\mu_\delta, \hat \mu)/d \lesssim \delta + \eps_{\rm score}^2  + \eps_{\rm disc}^2. 
\]
By the formulation of $\eps_{\rm score}^2$ and $\eps_{\rm disc}^2$ in Eq.~(\ref{eqn:VI-ResN-gen-bound-Ising}) and (\ref{eqn:score-disc-bound-cor-Ising}) and by $\sup_{t} \eps_{{\rm VI}, t}^2(\bA) \lesssim 1/ d$, to ensure $W_1^2(\mu, \hat \mu) / d \lesssim \eps^2$, it suffices to take 
\[
\begin{aligned}
&~\delta \asymp \eps^2,&~~~~ T \asymp&~ \log(1/\eps),&~~~\kappa \asymp&~ \eps^2 / \log(1/\eps),&~~~ N \asymp&~ \log^2(1/\eps) /\eps^2,\\
&~d \asymp 1/\eps^4,&~~~M \asymp&~ 1 / \eps^{6},& ~~~L \asymp&~ \log(1/\eps),&~~~n \asymp&~ \log^3(1/\eps)/\eps^{18}.
\end{aligned}
\]

\paragraph{The role of dimensionality.} In contrast to existing results \cite{oko2023diffusion, chen2023score} in which the score estimation error bounds exhibit a curse of dimensionality, our result seems to demonstrate a ``blessing of dimensionality''. Specifically, the term $\eps_{\rm VI, t}^2$ in Theorem~\ref{thm:Ising-score-approximation} is independent of ResNet size, sample size, and will typically vanish as dimension $d$ goes to infinity. However, we cannot conclude that score estimation actually becomes easier for higher-dimensional Ising models, since our result only provides an upper bound on the estimation error. Whether score approximation truly simplifies with increasing dimensions is an open question deserving further investigation. 

\paragraph{Generalizing Assumption~\ref{ass:Ising-free-energy}.} While Assumption~\ref{ass:Ising-free-energy} provides a sufficient condition for efficient score approximation, it is stronger than necessary. For example, in the Sherrington-Kirkpatrick model when $\bA = \beta \bJ$ where $\bJ \sim \GOE(d)$, an efficient sampling algorithm is known when $\beta < 1$ \cite{celentano2022sudakov}. However, we can only verify Assumption~\ref{ass:Ising-free-energy} for $\beta \le \beta_0$ for some $1/4 < \beta_0 < 1/2$. Nevertheless, we believe one can weaken our assumption to show score estimation is efficient for any $\beta < 1$ by leveraging local convexity of the TAP free energy of the SK model, proved in \cite{el2022sampling, celentano2022sudakov}.

\paragraph{The choice of sampling scheme and discretization scheme.} Importantly, our score estimation error bound in Theorem~\ref{thm:Ising-score-approximation} can combine with sampling schemes beyond DDPM, as it does not rely on a specific diffusion model. For instance, stochastic localization schemes \cite{eldan2013thin, el2022sampling, montanari2023posterior, montanari2023sampling} estimate the denoiser rather than the score, and our analysis can be adapted to bound the denoiser estimation error, enabling sampling guarantees for stochastic localization. Additionally, the discretization scheme and sampling error bound in Corollary~\ref{cor:KL-sampling-bound-Ising} may not be optimal. The analysis could likely be sharpened, or the discretization improved, to provide tighter error guarantees. 




\subsection{Proof outline of Theorem~\ref{thm:Ising-score-approximation}}\label{sec:proof-outline-Ising}

Here, we outline the proof of Theorem~\ref{thm:Ising-score-approximation}, with full details in Appendix~\ref{app:Ising-new}.

Recall that we have $\hat \bs_t(\bz) = \sP_t[\ResN_{\what \bW}](\bz)$, where $\what \bW = \argmin_{\bW \in \cW} \hat \E[\| \sP_t \ResN_{\bW}(\bz) + \sigma_t^{-1} \bg  \|_2^2 ]$ for $\cW = \cW_{d, D, L, M, B}$. Here, $\hat \E$ denotes averaging over the empirical data distribution. By standard error decomposition analysis in empirical risk minimization theory, we have:
\[
\begin{aligned}
&~\E[\| \sP_t [\ResN_{\what \bW}](\bz)  + \sigma_t^{-1} \bg\|_2^2] / d 
\le  \inf_{\bW \in \cW} \E[\| \sP_t [\ResN_{\bW}](\bz) + \sigma_t^{-1} \bg  \|_2^2] / d \\
&~~~~~+ 2 \sup_{\bW \in \cW} \Big| \hat \E [ \| \sP_t [\ResN_{\bW}](\bz) + \bsigma_t^{-1} \bg \|_2^2] / d - \E [\| \sP_t [\ResN_{\bW}](\bz) + \bsigma_t^{-1} \bg \|_2^2] / d \Big|. 
\end{aligned}
\]
Furthermore, a standard identity in diffusion model theory shows:
\[
\begin{aligned}
\E[\| \hat \bs_t(\bz)  - \bs_t(\bz)\|_2^2 ] / d = \E[\| \hat \bs_t(\bz)  + \sigma_t^{-1} \bg \|_2^2] / d  + C,~~~~ C = \E[\| \bs_t(\bz) \|_2^2] / d - \E[\| \sigma_t^{-1} \bg \|_2^2]/d.
\end{aligned}
\]
Combining the above yields:
\[
\E [\| \hat \bs_t(\bz) - \bs_t(\bz) \|_2^2] / d  \le \bar\eps_{\rm app}^2 + \bar \eps_{\rm gen}^2, 
\]
where $\bar\eps_{\rm app}^2$ is the approximation error and $\bar \eps_{\rm gen}^2$ is the generalization error, 
\[
\begin{aligned}
\bar\eps_{\rm app}^2 =&~ \inf_{\bW \in \cW} \E[\| \sP_t [\ResN_{\bW}](\bz) - \bs_t(\bz)  \|_2^2] / d, \\
\bar \eps_{\rm gen}^2 =&~ 2 \sup_{\bW \in \cW} \Big| \hat \E [ \| \sP_t [\ResN_{\bW}](\bz) + \bsigma_t^{-1} \bg \|_2^2] / d - \E [\| \sP_t [\ResN_{\bW}](\bz) + \bsigma_t^{-1} \bg \|_2^2] / d \Big|. 
\end{aligned}
\]

The generalization error $\bar \eps_{\rm gen}^2$ can be controlled by a standard empirical process analysis. We simply use a parameter counting argument to control this term, which can be found in Proposition~\ref{prop:uniform2}. This gives rise to the term $\eps_{\rm gen}^2$ in (\ref{eqn:VI-ResN-gen-bound-Ising}). 

To control the approximation error $\bar\eps_{\rm app}^2$, we note that $\bs_t(\bz) = (\lambda_t \cdot \bbm_t(\bz) - \bz) / \sigma_t^2$, where $\bbm_t(\bz) = \E_{(\bx, \bg) \sim \mu \otimes \cN(\bzero, \id_d)}[\bx | \lambda_t \bx + \sigma_t \bg = \bz ]$ is the denoiser. Thus, approximating the score function reduces to approximating $\bbm_t(\bz)$ using a ResNet. By Assumption~\ref{ass:Ising-free-energy}, the denoiser $\bbm_t$ can be approximated by the minimizer of a variational free energy $\cF_t^{\rm VI}$. This minimizer can be found by a fixed point iteration, which can further be approximated by a ResNet. 

More specifically, simple calculus shows that the minimizer $\hat \bbm = \hat \bbm_t$ of the variational free energy $\cF_t^{\rm VI}$ satisfies the fixed point equation 
\[
\hat \bbm = \tanh(\bU \hat \bbm + \bh),~~~~~\bU = \bA - \bK,~~~~~ \bh = \lambda_t \sigma_t^{-2} \bz.
\]
When $\| \bU \|_{\op} < 1$, this can be efficiently solved by fixed point iteration 
\[
\hat \bbm \approx \bbm^L,~~~~~~~~\bbm^{\ell +1} = \tanh(\bU  \bbm^\ell + \bh),~~~~~~~~ \bbm^0 = \bzero.
\]
This fixed point iteration can further be approximated by the ResNet structure (\ref{eqn:relu-resnet}), where $\tanh$ is approximated by a linear combination of ReLU activations. Lemma~\ref{lemma:approx-tanh} and \ref{lem:approximate_TAP_iteration_Ising} analyze this approximation error $\eps_{\text{ResN}}^2$. Our analysis shows that the total approximation error $\bar\eps_{\rm app}^2$ is controlled by $\eps_{\rm VI}^2 + \eps_{\rm ResN}^2$. Adding the generalization error yields the overall score estimation error bound in Eq.~(\ref{eqn:score-estimation-error-Ising}). 


%

\section{Generalization to other high-dimensional graphical models}
\label{sec:latent-variable-ising}

To demonstrate the flexibility of our proposed framework, we now generalize the results from Section~\ref{sec:Ising-new} to other high-dimensional graphical models. Specifically, we consider latent variable Ising models (Section~\ref{sec:latent-Ising}), the conditional Ising models for the conditional generative modeling task (Section~\ref{sec:conditional-Ising}), and the sparse coding models (Section~\ref{sec:SC}). 



\subsection{Diffusion models for latent variable Ising models}\label{sec:latent-Ising}

In the latent variable Ising model $\mu$, we have a coupling matrix $\bA = [\bA_{11}, \bA_{12}; \bA_{12}^\sT, \bA_{22}] \in \R^{ (d+m) \times (d + m)}$ (where $\bA_{11} \in \R^{d \times d}$, $\bA_{12} \in \R^{d \times m}$, and $\bA_{22} \in \R^{m \times m}$), specifying a joint distribution over $(\bx, \btheta) \in \{ \pm 1\}^{d + m}$, 
\begin{equation}\label{eqn:latent-variable-Ising}
\mu(\bx, \btheta) \propto \exp\{ \< \bx, \bA_{11} \bx\>/2 + \< \bx, \bA_{12} \btheta\> + \< \btheta, \bA_{22}\btheta\>/2 \},~~~~ \bx \in \{ \pm 1\}^d, \btheta \in \{ \pm 1\}^m. 
\end{equation}
Note that the joint distribution over $(\bx, \btheta)$ is still an Ising model. However, here we will treat $\btheta$ as a latent variable and consider generative modeling for the marginal distribution $\mu(\bx) = \sum_{\btheta} \mu(\bx,\btheta)$ when $\btheta$ is unobserved. When $\bA_{11} = \bzero$ and $\bA_{22} = \bzero$, this model reduces to a restricted Boltzmann machine, which is often used to model natural image distributions \cite{ranzato2010factored}.

We still consider the generative modeling task where we observe $\{ \bx_i \}_{i \in [n]} \sim_{iid} \mu$, and our goal is to sample a new $\hat \bx \sim \hat \mu$ with $\hat \mu \approx \mu$. To show the DDPM scheme (Algorithm \ref{alg:sampling-resnet}) provides a controlled error bound,  we need to bound the score estimation error \cite{benton2023linear}. This estimation error can be controlled if we assume the denoiser minimizes a VI objective. 

\begin{assumption}[Consistency of the free energy minimizer in marginal Ising models]\label{ass:Ising-free-energy-marginal}
Let $\bsigma = (\bx, \btheta) \sim \mu(\bx, \btheta) \propto \exp\{ \< \bsigma, \bA \bsigma\> / 2 \}$ and $\bz \sim \cN( \lambda_t \bx, \sigma_t^2 \id_d)$. For any fixed $t$, assume that there exists $\eps_{{\rm VI}, t}^2(\bA) < \infty$ and $\bK = \bK(\bA, t) \in \R^{(d+m) \times (d+m)}$ with $\| \bK - \bA \|_{\op} \le A < 1$, such that
\[
\begin{aligned}
&~ \E_{\bz \sim \mu_t} [\| \hat \bbm_t(\bz) - \bbm_t(\bz) \|_2^2 ] / d \le \eps_{{\rm VI}, t}^2(\bA),~~~~~~~~~~~~~~~~~ \hat \bbm_t(\bz) = [\hat \bomega_t(\bz)]_{1:d},\\
&~ \hat \bomega_t(\bz) = \argmin_{\bomega \in [-1, 1]^{d+m}} \Big\{ \sum_{i = 1}^{d+m} -\sh_{\rm bin}(\omega_i) - \frac{1}{2} \< \bomega, \bA \bomega\>  - \frac{\lambda_t}{\sigma_t^2} \< \bz, \bomega_{1:d}\> + \frac{1}{2} \< \bomega, \bK \bomega \>\Big\}. \\
\end{aligned}
\]
\end{assumption}
Assumption~\ref{ass:Ising-free-energy-marginal} can be verified in concrete examples. Lemma \ref{lem:Ising-VB-consistency} still applies in this model: when $\| \bA \|_{\op} < 1/2$, taking $\bK = \bzero$ gives $\eps_{{\rm VI}, t}^2(\bA) \le 4 d^{-1} (1 - 2 \| \bA \|_{\op})^{-1} \| \bA \|_F^2$. We conjecture that for $\bA$ being spin glass models like the Sherrington-Kirkpatrick model at high temperature, there exists $\bK$ such that $\E[\eps_{{\rm VI}, t}^2(\bA)] \to 0$ as $d, m \to \infty$. Given Assumption~\ref{ass:Ising-free-energy-marginal}, the following theorem provides a score estimation error bound and a sampling error bound in latent variable Ising models, proved in Appendix~\ref{app:Ising-score-approximation-marginal}. 

\begin{theorem}\label{thm:Ising-score-approximation-marginal}
Let Assumption~\ref{ass:Ising-free-energy-marginal} hold. Let $\{ \hat \bs_{T - t_k} \}_{0 \le k \le N-1}$ be the approximate score function given by Algorithm \ref{alg:sampling-resnet} in which we take 
\[
D = 3(d + m), ~~~~~ M \ge 4 (d + m),~~~~~ B \ge 7 \cdot (M/(d + m)) \cdot \log (M) + \sqrt{d + m} + 1 / \min_{k} 
\{ T - t_k \}. 
\]
Then with probability at least $1 - \eta$, for any $t \in \{ T - t_k\}_{0 \le k \le N-1}$, we have
\begin{align}\label{eqn:score-estimation-bound-latent-Ising}
&~\E_{\bz \sim \mu_t}[\|  \hat \bs_t(\bz) - \bs_t(\bz) \|_2^2]/d
\lesssim \lambda_t^2 \sigma_t^{-4} \cdot \Big( \eps_{{\rm VI}, t}^2(\bA) + \eps^2_{{\rm ResN}}  + \eps^2_{{\rm gen}} \Big),
\end{align}
where $\eps_{{\rm VI}, t}^2$ is given in Assumption~\ref{ass:Ising-free-energy-marginal}, and
\begin{equation}\label{eqn:VI-ResN-gen-bound-Ising-latent-variable}
\begin{aligned}
\eps^2_{{\rm ResN}} = \frac{d + m}{d}  \Big( \frac{(d + m)^2}{M^2(1 - A)^2} + A^{2L} \Big), ~~~~~ \eps^2_{{\rm gen}} = \sqrt{\frac{ (M  L + d) (d + m) [ T + L \log (B L ) ] + \log(N/\eta)}{n}}.
\end{aligned}
\end{equation}
%
Furthermore, consider the two-phase discretization scheme as in Definition \ref{def:two-phase-discretization-scheme}, we have with probability $1 - \eta$ that
\begin{equation}\label{eqn:KL-error-latent-Ising}
{\rm KL}( \mu_\delta, \hat \mu ) / d \lesssim \delta^{-1} \cdot \Big( \sup_{0 \le k \le N-1} \eps_{{\rm VI}, T - t_k}^2 + \eps^2_{{\rm ResN}}  + \eps^2_{{\rm gen}} \Big) + \kappa^2 N + \kappa T + e^{- 2 T}. 
\end{equation}
\end{theorem}


\subsection{Conditional diffusion models for Ising models}\label{sec:conditional-Ising}

In the conditional Ising model, we also have a coupling matrix $\bA = [\bA_{11}, \bA_{12}; \bA_{12}^\sT, \bA_{22}] \in \R^{ (d+m) \times (d + m)}$, specifying a joint distribution over $(\bx, \btheta) \in \{ \pm 1\}^{d + m}$ as in Eq.~(\ref{eqn:latent-variable-Ising}). However, we now consider the conditional generative modeling task where we observe $\{ (\bx_i, \btheta_i) \}_{i \in [n]} \sim_{iid} \mu$. The goal is to sample $\hat \bx \sim \hat \mu(\cdot | \btheta) \approx \mu(\cdot | \btheta)$ for a given $\btheta$. Such problems naturally arise in image imputation tasks, where $(\bx, \btheta)$ represents a full image, $\btheta$ is the observed part, and $\bx$ is the missing part to impute.

The conditional generative modeling task can be solved using the conditional DDPM scheme (Algorithm \ref{alg:sampling-resnet-conditional} as described in Appendix \ref{app:ddpm-conditional}). To bound the error, we need to control the estimation error of the conditional score $\bs_t(\bz; \btheta) = \nabla_{\bz}\log \mu_t(\bz, \btheta)$. By Tweedie's formula, we have $\bs_t(\bz; \btheta) = (\lambda_t \bbm_t(\bz; \btheta) - \bz) / \sigma_t^2$, where $\bbm_t(\bz; \btheta) := \E_{(\bx, \btheta, \bg) \sim \mu \otimes \cN(0, 1)}[\bx | \btheta, \bz = \lambda_t \bx + \sigma_t \bg]$ is the conditional denoiser. We assume the following about $\bbm_t(\bz; \btheta)$.  

\begin{assumption}[Consistency of the free energy minimizer in conditional Ising models]\label{ass:Ising-free-energy-conditional}
Let $(\bx, \btheta) \sim \mu(\bx, \btheta) \propto \exp\{ \< \bsigma, \bA \bsigma\> / 2 \}$ and $\bz \sim \cN( \lambda_t \bx, \sigma_t^2 \id_d)$. For any fixed $t$, assume that there exists $\eps_{{\rm VI}, t}^2(\bA) < \infty$ and $\bK = \bK(\bA, t) \in \R^{d \times d}$ with $\| \bK - \bA_{11} \|_{\op} \le A < 1$, such that
\[
\begin{aligned}
&~ \E_{(\btheta, \bz)} [\| \hat \bbm_t(\bz; \btheta) - \bbm_t(\bz; \btheta) \|_2^2 ] / d \le \eps_{{\rm VI}, t}^2(\bA), \\
&~ \hat \bbm_t(\bz;\btheta) = \argmin_{\bbm \in [-1, 1]^d} \Big\{ \sum_{i = 1}^{d} -\sh_{\rm bin}(m_i) - \frac{1}{2} \< \bbm, \bA_{11} \bbm \> - \< \bbm, \bA_{12} \btheta\>- \frac{\lambda_t}{\sigma_t^2} \< \bz, \bbm \> + \frac{1}{2} \< \bbm, \bK \bbm \>\Big\}. \\
\end{aligned}
\]
\end{assumption}


Assumption~\ref{ass:Ising-free-energy-conditional} can be verified in concrete examples. Lemma \ref{lem:Ising-VB-consistency} still applies in this model: when $\| \bA_{11} \|_{\op} < 1/2$, taking $\bK = \bzero$ gives $\eps_{{\rm VI}, t}^2(\bA) \le 4 d^{-1} (1 - 2 \| \bA_{11} \|_{\op})^{-1} \| \bA_{11} \|_F^2$. We conjecture that $\E[\eps_{{\rm VI}, t}^2(\bA)] \to 0$ as $d, m \to \infty$ for $\bA$ being spin glass models at high temperature. Given Assumption~\ref{ass:Ising-free-energy-conditional}, the following theorem provides a conditional score estimation error bound and a conditional sampling error bound in conditional Ising models, proved in Appendix~\ref{app:Ising-score-approximation-conditional}. 

\begin{theorem}\label{thm:Ising-score-approximation-conditional}
Let Assumption~\ref{ass:Ising-free-energy-conditional} hold. Let $\{ \hat \bs_{T - t_k} \}_{0 \le k \le N-1}$ be the approximate score function given by Algorithm \ref{alg:sampling-resnet-conditional} in which we take
\[
D = 4d, ~~~~~ M \ge 4d,~~~~~ B \ge 7 \cdot (M/d) \cdot \log (M) + \sqrt{d} + 1 / \min_{k} 
\{ T - t_k\} + \|\bA_{12}\|_{\op} \cdot (M / d + 1). 
\]
Then with probability at least $1 - \eta$, for any $t \in \{ T - t_k\}_{0 \le k \le N-1}$, we have
\[
\begin{aligned}
&~\E_{(\btheta, \bz)}[\|  \hat \bs_t(\bz; \btheta) - \bs_t(\bz; \btheta) \|_2^2]/d
\lesssim \lambda_t^2 \sigma_t^{-4} \cdot \Big( \eps_{{\rm VI}, t}^2(\bA) + \eps^2_{{\rm ResN}}  + \eps^2_{{\rm gen}} \Big),
\end{aligned}
\]
where $\eps_{{\rm VI}, t}^2$ is given in Assumption~\ref{ass:Ising-free-energy-conditional}, and
\begin{equation}\label{eqn:VI-ResN-gen-bound-Ising-conditional}
\begin{aligned}
\eps^2_{{\rm ResN}} =  \frac{d^2}{M^2(1 - A)^2} + A^{2L},~~~~
\eps^2_{{\rm gen}} =&~ \sqrt{\frac{ (M d L + d (d+ m)) [ T + L \log (B Ld^{-1}(m + d)) ] + \log(N/\eta)}{n}}.
\end{aligned} 
\end{equation}
Furthermore, consider the two-phase discretization scheme as in Definition \ref{def:two-phase-discretization-scheme}, we have with probability $1 - \eta$ that
\[
\E_{\btheta \sim \mu} [{\rm KL}( \mu_\delta(\cdot | \btheta), \hat \mu( \cdot | \btheta) ) / d] \lesssim \delta^{-1} \cdot \Big( \sup_{0 \le k \le N-1} \eps_{{\rm VI}, T - t_k}^2 + \eps^2_{{\rm ResN}}  + \eps^2_{{\rm gen}} \Big) + \kappa^2 N + \kappa T + e^{- 2 T}. 
\]
\end{theorem}
We note the score estimation and sampling error bounds in Theorem~\ref{thm:Ising-score-approximation-conditional} are averaged over $\btheta \sim \mu(\btheta) = \sum_{\bx \in \{ \pm 1\}^d} \mu(\bx, \btheta)$, the marginal of $\btheta$. These do not ensure error bounds for any fixed $\btheta$.



\subsection{Diffusion models for sparse coding}\label{sec:SC}

In sparse coding, there is a fixed dictionary $\bA \in \R^{d \times m}$. Our observations are noisy, sparse linear combinations of the columns of the dictionary: $\bx_i = \bA \btheta_i + \beps_i$ for $i \in [n]$. Here $\beps_i \sim_{iid} \cN(\bzero, \tau^2 \id_d)$ are noise vectors, and $\btheta_i \sim_{iid} \pi_0^{\otimes m}$ are sparse coefficient vectors, with $\pi_0 \in \cP(\R)$ having a Dirac delta mass at $0$. Given observations $\{ \bx_i \}_{i \in [n]}$, sparse coding typically aims to recover $\bA$ and estimate $\{ \btheta_i\}_{i \in [n]}$. Instead, we consider the generative modeling problem --- learning a model to generate new samples $\hat \bx$ resembling the observations $\{ \bx_i \}_{i \in [n]}$.

The generative modeling task for sparse coding can be solved by the DDPM sampling scheme (Algorithm~\ref{alg:sampling-resnet}). To control the score estimation error, we make the following assumption on the following denoising function $\be_t$, which requires a little modification in the sparse coding setting: 
%
\begin{align}\label{eqn:be-z-ast}
\be_t(\bz_{\ast}) := \E_{(\bz_{\ast}, \btheta)} \left[ \btheta \mid \bz_{\ast} \right], 
\qquad \bz_{\ast} = \bA \btheta + \bar \beps, \qquad \bar\eps_j \iidsim  \cN(0, \tau^2 + \sigma_t^2 / \lambda_t^2).
\end{align}

\begin{assumption}[Consistency of the free energy minimizer in sparse coding]\label{ass:SC-free-energy}
Fix $\bA \in \R^{d \times m}$. Consider the Bayesian linear model $\bz_{\ast} = \bA \btheta + \bar\beps \in \R^d$, $\bar\eps_j \sim_{iid} \cN(0, \bar \tau_t^2)$ where $\bar \tau_t^2 = \tau^2 + \sigma_t^2/\lambda_t^2$ and $\theta_i \sim_{iid} \pi_0$ where $\pi_0 \in \cP([- \Pi, \Pi])$. Assume that for any $t > 0$, there exist $(\nu_t, \bK_t, \eps_{{\rm VI}, t}^2)$ that depend on $(\pi_0, \bA, \tau, t)$ with $\| \bA^\sT \bA / \bar \tau_t^2 - \bK_t \|_{\op} \leq A < 1/\Pi^2$, such that 
\[
\begin{aligned}
&~ \E_{\bz \sim \mu_t} [\| \hat \be_t(\bz_{\ast}) - \be_t(\bz_{\ast}) \|_2^2 ] / m \le \eps_{{\rm VI}, t}^2( \bA),\\
&~ \hat \be_t(\bz_{\ast}) = \argmin_{\be \in [- \Pi, \Pi]^m} \Big\{ \sum_{i = 1}^m \max_{\lambda} \Big[  \lambda m_i - \log \E_{\beta \sim \pi_0}[e^{\lambda \beta - \beta^2 \nu_t / 2}] \Big] + \frac{1}{2 \bar \tau_t^2 }\| \bz_{\ast} - \bA \be \|_2^2 -  \frac{1}{2} \< \be, \bK_t \be\> \Big\}. \\
\end{aligned}
\]
\end{assumption}

We also use a different truncation operator in Algorithm~\ref{alg:sampling-resnet}, replacing $\sP_t$ by $\bar \sP_t$: 
$$\bar \sP_{t}[f] (\bz) = \proj_{\sqrt{m} \, \|\bA\|_{\op} \Pi  \lambda_t (\sigma_t^2 + \tau^2 \lambda_t^2)^{-1} }(f(\bz) + (\sigma_t^2 + \tau^2 \lambda_t^2)^{-1} \bz ) - (\sigma_t^2 + \tau^2 \lambda_t^2)^{-1} \bz.
$$
Given Assumption~\ref{ass:SC-free-energy}, the following theorem provides a score estimation error bound in sparse coding models, proved in Appendix~\ref{app:SC-score-approximation}. 

\begin{theorem}\label{thm:SC-score-approximation}
Let Assumption~\ref{ass:SC-free-energy} hold. Let $\{ \hat \bs_{T - t_k} \}_{0 \le k \le N-1}$ be the approximate score function given by Algorithm \ref{alg:sampling-resnet} in which we take 
\begin{align*}
& D = 3m + d, ~~~~~ M \ge 4m,\\
& B \ge (M/m)  \cdot \left( A + 1 + 2\Pi^2 + w_\star \right) + 2\Pi + 6  + (\|\bA\|_{\op} + 1) / \min_k\{ T - t_k\} +\tau^{-2} \|\bA\|_{\op} + \sqrt{m}, 
\end{align*}
where $w_\star$ is defined in Eq.~(\ref{eqn:w-star-definition-in-proof}). Then with probability at least $1 - \eta$, when $n \ge \log (2 / \eta)$, for any $t \in \{ T - t_k\}_{0 \le k \le N-1}$, we have the following score estimation error bound
\[
\begin{aligned}
&~\E_{(\btheta, \bz)}[\|  \hat \bs_t(\bz; \btheta) - \bs_t(\bz; \btheta) \|_2^2]/d\\
\lesssim&~   \lambda_t^2 \|\bA\|_{\op}^2 (1 + \tau^{-4}) \cdot \frac{m}{d} \cdot \Big( \eps_{{\rm VI}, t}^2(\bA) + \eps^2_{{\rm ResN}} \Big)  + \Big( \lambda_t^2 \|\bA\|_{\op}^2 (1 + \tau^{-4}) \Pi^2 \cdot \frac{m}{d} + \frac{\lambda_t^2}{ \sigma_t^2}  (1 + \tau^2) \Big) \eps^2_{{\rm gen}},
\end{aligned}
\] 
for $\eps_{{\rm VI}, t}^2$ as given in Assumption~\ref{ass:SC-free-energy}, and 
\begin{equation}\label{eqn:VI-ResN-gen-bound-SC}
\begin{aligned}
\eps^2_{{\rm ResN}} = \Pi^2 \cdot (\Pi^2 A)^{2L} + \frac{m^2\Pi^2}{(1 - \Pi^2A)^2 M^2},~~~~~ \eps^2_{{\rm gen}} = \sqrt{\frac{(dD + LDM) \cdot (T +  L ) \cdot \iota}{n}}.
\end{aligned}
\end{equation}
where $\iota = \log(LBnmT(1 + \tau) (1 + \| \bA \|_{\op} \Pi) \tau^{-1} N \eta^{-1})$. 

\end{theorem}


Theorem~\ref{thm:SC-score-approximation} can be further combined with an off-the-shelf discretization bound as in Theorem~\ref{thm:benton} to derive a sampling error bound. 

\paragraph{Verifying the assumption.} The VI approximation error $\eps_{\rm VI}^2$ in Assumption~\ref{ass:SC-free-energy} converges to $0$ as $d, m \to \infty$ when $\bA$ is a rotationally invariant design matrix, by choosing the variational objective to be the TAP free energy \cite{thouless1977solution}. Specifically, assume the SVD decomposition $\bA = \bQ \bD \bO^\sT$ where $\bQ \in \R^{d \times d}$ and $\bO \in \R^{m \times m}$ are orthonormal, and $\bD \in \R^{d \times m}$ is diagonal. Assume that $\bO \sim {\rm Haar}({\rm SO}(m))$ is independent of everything else, and the diagonal elements of $\bD$ have certain empirical distribution converging to a bounded distribution $\sD$. As an example, $\bA$ with IID Gaussian entries of variance $1/m$ is rotationally invariant. Under the assumption that $\bA$ is rotationally invariant, a corollary of \cite[Theorem 1.11]{li2023random} gives the following lemma, with proof contained in Appendix~\ref{app:proof-lemma-SC-orthogonal}. 


\begin{lemma}[Corollary of \cite{li2023random} Theorem 1.11] \label{lem:SC-TAP-consistency}
Let $\bA \in \R^{d \times m}$ be a rotationally invariant design matrix and let Assumption \ref{ass:SC-in-proof} hold.
Then for any $\pi_0$, $\alpha = d / m$, and limiting distribution $\sD$, there exists $\tau^2 > 0$, such that for any $t$, there exists matrices $\bK = c_t \id_d$ for some $c_t$, such that  
\[
\eps_{{\rm VI}, t}(\bA) \stackrel{a.s.}{\to} 0,~~~~ d, m \to \infty,~~~ d / m \to \alpha.
\]
\end{lemma}

Although Lemma~\ref{lem:SC-TAP-consistency} does not provide non-asymptotic control of the VI approximation error, we believe this could be obtained through more refined analysis. 

\section{Other related work}

%

\paragraph{Score function approximation in diffusion models.} Neural network-based score function approximation has been recently studied in \cite{oko2023diffusion, chen2023score, yuan2023reward, shah2023learning}. \cite{oko2023diffusion} assumes that the data distribution $\mu \in \cP(\R^d)$ has a density with $s$-order bounded derivatives and shows that estimating the score to precision $\eps$ requires network size and sample complexity at least $\eps^{-d/s}$. This suffers from the curse of dimensionality unless the data distribution is very smooth ($s \asymp d$). \cite{oko2023diffusion, chen2023score} avoid the curse of dimensionality by assuming that the data distribution has a low-dimensional structure, but this assumption does not apply to high-dimensional graphical models. \cite{shah2023learning} considers Gaussian mixture models where the score function has a closed form, enabling parameterized by a small shallow network.

In contrast, we assume the data distribution is a graphical model, common for images and text \cite{blei2003latent, mnih2007three, geman1986markov}. Assuming the efficiency of variational inference approximation, we show that the score can be well-approximated by a network polynomial in dimension, enabling efficient learning from polynomial samples. Our graphical model assumption and algorithm unrolling of variational inference perspective circumvent dimensionality issues faced by prior work.


\paragraph{Discretizing the diffusion process.} 
Recent work has studied the convergence rates of the discretized reverse SDEs/ODEs for diffusion models \cite{liu2022let, li2023towards, lee2023convergence, chen2022improved, chen2023restoration, chen2022sampling, chen2023probability, chen2023improved, benton2023linear}. In particular, \cite{chen2023improved, benton2023linear} provide minimal assumptions to quantitatively control the KL divergence between the perturbed and data distributions. These assumptions include the second moment bound and the controlled score estimation error. Our work focuses on controlling the score estimation error, a goal that is orthogonal to analyzing discretization schemes. Specifically, we directly leverage the result of \cite{benton2023linear} to provide an end-to-end error bound. 

\paragraph{Stochastic localization.}

Stochastic localization, proposed by \cite{eldan2013thin, eldan2022analysis}, is another sampling scheme similar to diffusion models. Recent works have developed algorithmic sampling techniques based on stochastic localization \cite{el2022sampling, montanari2023posterior, celentano2022sudakov}. \cite{montanari2023sampling} shows the equivalence of stochastic localization to the DDPM sampling scheme in the Gaussian setting and proposes various ways of generalizing stochastic localization schemes. While we present our results in the diffusion model framework, our methods can also provide sampling error bound for stochastic localization schemes. 

\paragraph{Neural network approximation theory.}

Classical neural network approximation theory typically relies on assumptions that the target function is smooth or hierarchically smooth \cite{cybenko1989approximation, hornik1989multilayer, hornik1993some, pinkus1999approximation, devore2011approximation, weinan2019barron, yarotsky2017error, barron1993universal, bach2017breaking, devore2021neural}. These enable overcoming the curse of dimensionality for higher-order smooth or low-dimensional target functions \cite{barron1993universal, weinan2019barron, bach2017breaking}. However, when applying them to score function approximation in diffusion models, it is unclear whether such assumptions hold for the score function of high-dimensional graphical models. 

A recent line of work investigated the expressiveness of neural networks through an algorithm approximation viewpoint \cite{wei2022statistically, bai2023transformers, giannou2023looped, liu2022transformers, marwah2021parametric, marwah2023neural}. \cite{wei2022statistically, bai2023transformers, giannou2023looped, liu2022transformers} show that transformers can efficiently approximate several algorithm classes, such as gradient descent and Turing machines. \cite{marwah2021parametric, marwah2023neural} demonstrate that deep networks can efficiently approximate PDE solutions by approximating the gradient dynamics. We also adopt this algorithmic perspective for neural network approximation but apply it to score function approximation for diffusion models.


\paragraph{Variational inference in graphical models.}

Variational inference is commonly used to approximate the marginal statistics of graphical models \cite{pearl2022reverend, jordan1999introduction, minka2013expectation, mezard2009information, wainwright2008graphical, blei2017variational}. In certain regimes, such as graphical models in the high temperature, naive variational Bayes has been shown to yield consistent posterior estimates \cite{chatterjee2016nonlinear, eldan2018gaussian, jain2018mean, mukherjee2022variational}. For high dimensional statistical models in the low signal-to-noise ratio regime, approximate message passing \cite{donoho2009message, feng2022unifying} and equivalently TAP variational inference \cite{thouless1977solution, ghorbani2019instability, fan2021tap, celentano2021local, celentano2022sudakov, celentano2023mean}, can achieve consistent estimation of the Bayes posterior. Our paper directly adopts results developed for variational inference methods in spin glass models and Bayesian linear models \cite{talagrand2003spin, chatterjee2010spin, barbier2019optimal, barbier2016mutual, fan2021tap, fan2022tap, li2023random, celentano2021local, celentano2022sudakov, celentano2023mean}.



\paragraph{Algorithm unrolling.}

A line of work has focused on neural network denoising by unrolling iterative denoising algorithms into deep networks \cite{gregor2010learning, zheng2015conditional, zhang2018ista, papyan2017convolutional, ma2021unified, chen2018theoretical, borgerding2017amp, monga2021algorithm, yu2023white, yu2023emergence}. These approaches include unrolling ISTA for LASSO into recurrent nets \cite{gregor2010learning, zhang2018ista, papyan2017convolutional, borgerding2017amp}, unrolling belief propagation for Markov random fields into recurrent nets \cite{zheng2015conditional}, and unrolling graph denoising algorithms into graph neural nets \cite{ma2021unified}. Our work also adopts this algorithm unrolling viewpoint, but with a different goal: while the prior literature has mainly focused on devising better denoising algorithms, our work uses this perspective to provide neural network approximation theories for diffusion-based generative models. 


\section{Discussions}



\paragraph{Algorithmic hard phase.} The algorithm unrolling perspective can also shed light on the failure mode of score approximation, namely when score functions cannot be efficiently represented by neural networks. For example, we can conclude that the score function of the Sherrinton-Kirkpatrick model with $\beta > 1$ cannot be efficiently represented by a neural network, as it was proven in \cite{el2022sampling} that there is no stable algorithm to sample the SK Gibbs measure for $\beta > 1$. More generally, the relationship between hardness of sampling, hardness of diffusion-based sampling, and hardness of score approximation deserves further investigation. Recent work such as \cite{ghio2023sampling} provides a valuable discussion on this important topic.

\paragraph{Future directions.} Our work leaves open several interesting questions. One issue is that for fixed dimension $d$, our score approximation error does not decay as the network size and sample size increase, and is lower bounded by the variational inference approximation error $\eps_{{\rm VI}}^2$. To resolve this, one approach could consider a hierarchy of variational inference algorithms, such as Plefka's expansion \cite{plefka1982convergence, maillard2019high}, which provide increasingly accurate approximations. Using these hierarchical approximations within our framework could potentially reduce the score approximation error. 

Another open question is understanding the algorithms that diffusion neural networks like U-nets and transformers implement in diffusion models for image tasks. One hypothesis is that U-nets with convolution layers are implementing some form of variational inference denoising on graphical models with certain locality and invariance structures. It would be interesting to test this hypothesis on real image datasets. 

Finally, an exciting direction is leveraging the algorithmic unrolling perspective to design improved neural network architectures for diffusion models. The resulting architectures could potentially be more interpretable and achieve better emergent capabilities, as illustrated by recent works like  \cite{yu2023white, yu2023emergence}. 

%

\section*{Acknowledgement}

Song Mei is supported in part by NSF DMS-2210827 and NSF CCF-2315725. 

\bibliographystyle{amsalpha}
\bibliography{reference.bib}

\clearpage

\appendix

\section{Technical preliminaries}

\subsection{DDPM conditional sampling scheme}\label{app:ddpm-conditional}

We provide the details of the DDPM conditional sampling scheme 
(Algorithm \ref{alg:sampling-resnet-conditional}) as mentioned in Section~\ref{sec:prelim}. The algorithm still has two steps, with minor modifications from unconditional DDPM (Algorithm \ref{alg:sampling-resnet}). In the first step, empirical risk minimization (Eq.~\eqref{eqn:ERM-in-alg-conditional}) fits manually-generated noises $\{ \bg_i \}_{i \in [n]}$ using the noisy samples and conditioning variables $\{ (\lambda_t \bx_i + \sigma_t \bg_i; \btheta_i) \}_{i \in [n]}$. The ResNet $\ResN_{\bW}: \R^d \times \R^m \to \R^d$ is parameterized by $\bW = \{\bW_1^{(\ell)} \in \R^{D \times M}, \bW_2^{(\ell)} \in \R^{M \times D}\}_{\ell \in [L]} \cup \{\bW_{\inside} \in \R^{(d + m + 1) \times D}, \bW_{\out} \in \R^{D \times d}\}$ and is defined iteratively as 
\begin{equation}\label{eqn:relu-resnet-conditional}\tag{ResNet-Conditional}
\resnet_{\bW}(\bz, \btheta) = \bW_{\out} \bu^{(L)},~~~~ \bu^{(\ell)} = \bu^{(\ell - 1)} + \bW_1^{(\ell)} \relu(\bW_2^{(\ell)} \bu^{(\ell - 1)}), ~~~~ \bu^{(0)} = \bW_{\inside} [\bz; \btheta; 1]. 
\end{equation}
The only difference between (\ref{eqn:relu-resnet}) and (\ref{eqn:relu-resnet-conditional}) is the input dimension. Minimization is over the ResNets with weights in the set (for parameters $d,m,D,L,M,B$): 
\begin{equation}\label{eqn:conditional-ResNet}
\begin{aligned}
\cW_{d, m, D, L, M, B} :=&~ \Big\{ \bW = \{ \bW_1^{(\ell)}, \bW_2^{(\ell)} \}_{\ell \in [L]} \cup \{\bW_{\inside}, \bW_{\out} \}:  \nrmps{\bW} \leq B \Big\}, \\
\nrmps{\bW} :=&~ \max_{\ell \in [L]} \big\{ \|\bW_1^{(\ell)}\|_{\op} + \|\bW_2^{(\ell)}\|_{\op} \big\} \vee \max\big\{ \|\bW_{\inside}\|_{\op}, \|\bW_{\out}\|_{\op}  \big\}.
\end{aligned}
\end{equation}
We still truncate the ResNet output using  $\sP_t$: for $f: \R^d \times \R^m \to \R^d$, we define $\sP_t[f] (\bz, \btheta) = \proj_{\lambda_t \sigma_t^{-2} \sqrt{d}}(f(\bz, \btheta) + \sigma_t^{-2} \bz ) - \sigma_t^{-2} \bz$, where $\proj_R$ projects $\bz \in \R^d$ into the $R$-Euclidean ball.

The second step of Algorithm \ref{alg:sampling-resnet-conditional} still discretizes the backward SDE through the exponential integrator scheme (\ref{eqn:exponential-integrator-discrete-conditional}) and the two-phase discretization scheme (Definition \ref{def:two-phase-discretization-scheme}). However, we replace the score function $\hat \bs_{t}(\what \bY_k)$ with the conditional score function $\hat \bs_{t}(\what \bY_k; \btheta) = \sP_t[\ResN_{\what \bW_t}](\what \bY_k, \btheta)$.

\begin{algorithm}
\caption{The DDPM conditional sampling scheme}\label{alg:sampling-resnet-conditional}
\begin{algorithmic}[1]
\REQUIRE Samples $\{ (\bx_i, \btheta_i) \}_{i \in [n]} \subseteq \R^d \times \R^m$. Conditional latent variable $\btheta$. ResNet parameters $(d, m, D, L, M, B)$. Discretization scheme parameters $(N, T, \delta, \{ t_k \}_{0 \le k \le N})$ with $0 = t_0 < \cdots < t_{N} = T - \delta$. Denote $\gamma_k = t_{k+1} - t_k$. 
\STATE {\color{blue} // Computing the approximate conditional score function}
\STATE Sample $\{\bg_i \}_{i \in [n]} \sim_{iid} \cN(\bzero, \id_d)$. 
\FOR{$t \in \{ T - t_k \}_{0 \le k \le N}$ }
\STATE Solve the ERM problem below for $t = T -  t_k$: 
\begin{equation}\label{eqn:ERM-in-alg-conditional}
\what \bW_t = \arg\min_{\bW \in \cW_{d, m, D, L, M, B} } \frac{1}{n} \sum_{i = 1}^n \Big\| \sigma_t^{-1} \bg_i  + \sP_t [ \resnet_{\bW}]( \lambda_t \bx_i + \sigma_t \bg_i, \btheta_i)\Big\|_2^2,~~~ (\lambda_t, \sigma_t^2) = 
(e^{-t}, 1 - e^{-2t}). 
\end{equation}
\STATE Take the approximate score function to be $\hat \bs_t(\bz; \btheta) = \sP_t[ \resnet_{\what \bW_t}] (\bz, \btheta)$.  
\ENDFOR
\STATE {\color{blue} // Sampling by discretizing the stochastic differential equation}
\STATE  Sample $\what\bY_0 \sim \cN(\bzero, \id_d)$. 
\FOR{$k = 0, \cdots, N - 1$}
\STATE Sample $\bG_k \sim \cN(\bzero, \id_d)$. Calculate $\what \bY_{k+1}$ using the exponential integrator scheme: (here $\btheta$ is provided as an input)  
\begin{equation}\label{eqn:exponential-integrator-discrete-conditional}
\what \bY_{k + 1} = e^{\gamma_k} \cdot \what\bY_{k} + 2 (e^{\gamma_k} - 1) \cdot  \hat \bs_{T - t_k}(\what\bY_{k}; \btheta)  + \sqrt{e^{2\gamma_k} - 1} \cdot \bG_k. 
\end{equation}
\ENDFOR
\end{algorithmic}
{\bf Return}: $\hat \bx = \what \bY_N$.
\end{algorithm}

\subsection{Sampling error bound of the DDPM scheme}
\label{sec:DDPM-linear-convergence}

In this section, we state a result from \cite{benton2023linear}, which establishes the convergence of the DDPM discretization scheme, when evaluated using Kullback-Leibler (KL) divergence, with only minimal assumptions required. A slight generalization of the result in \cite{benton2023linear} is necessary, generalizing the identity covariance assumption to a general covariance matrix. The proof requires little modification, but we present a proof sketch here for completeness.

Suppose we are interested in drawing samples from $\mu$ in $\R^d$. The forward process that evolves according to the Ornstein-Uhlenbeck (OU) process is defined as the following SDE:
\begin{align}\label{eq:forward}
\de X_t = -X_t \de t + \sqrt{2} \de B_t, \qquad X_0 \sim \mu, \qquad  0 \leq t \leq T. 
\end{align}
In the above display, $(B_t)_{0 \leq t \leq T}$ is a standard Brownian motion in $\R^d$. We denote by $\mu_t$ the distribution of $X_t$. One can check that $X_t \overset{d}{=} e^{-t} X_0 + \sqrt{1 - e^{-2t}} \bg$ for $\bg \sim \cN(, \id_d)$ that is independent of $X_0$. The reverse process that corresponds to process \eqref{eq:forward} is defined via the SDE
\begin{align}\label{eq:reverse}
\de Y_t = \left\{ Y_t + 2 \nabla \mu_{T - t}(Y_t) \right\} \de t + \sqrt{2} \de B_t', \qquad Y_0 \sim \mu_T. 
\end{align}
An approximation to continuous-time process \eqref{eq:reverse} is obtained via performing time discretization, which directly leads to a sampling algorithm. More precisely, for $0 = t_0 < t_1 < \cdots < t_N = T - \delta$, we let 
\begin{align}\label{eq:B11}
\de \hat{Y}_t = \{\hat{Y}_t + 2 \hat s_{T - t_k}(\hat{Y}_{t_k})\} \de t + \de \hat{B}_t \,\,\,\,\, \mbox{ for }t_k \leq t \leq t_{k + 1}, \qquad \hat{Y}_0 \sim \cN(\bzero, \id_d), 
\end{align}
where $\hat s_{T - t}(\cdot)$ is an estimate of the true score function $s_{T - t}(\cdot) = \nabla \log \mu_{T - t}(\cdot)$. We denote by $p_t$ the marginal distribution of $\hat{Y}_t$, and set $\gamma_{k} = t_{k + 1} - t_k$. In addition, we assume there exists $\kappa > 0$, such that $\gamma_k \leq \kappa \cdot \min \{1, T - t_{k + 1}\}$. 

Next, we state the assumptions required to establish the discretization error bound of the DDPM sampling scheme. 
\begin{assumption}[Rescaled version of \cite{benton2023linear} Assumption 1]\label{assumption:score}
The score function estimator $\hat s_t$ satisfies
\begin{align*}
\sum_{k = 0}^{N - 1} \gamma_k \E_{\bx \sim \mu_{T - t_k}} \left[ \|\nabla \log \mu_{T - t_k}(\bx) - \hat s_{T - t_k}(\bx) \|_2^2 \right] \leq d \cdot \eps^2_{\score}. 
\end{align*}
\end{assumption}

\begin{assumption}\label{assumption:moment}
The data distribution $\mu$ has finite second moment: $\E_{\bx_0 \sim \mu}[\|\bx_0\|_2^2] \leq d \cdot B$, where $B \geq 1$ is a fixed constant. 
\end{assumption}

With Assumptions \ref{assumption:score} and \ref{assumption:moment}, we are ready to state the main theorem for this part. 

\begin{theorem}\label{thm:benton}[Theorem 1 of \cite{benton2023linear}]
Let Assumptions \ref{assumption:score} and \ref{assumption:moment} hold. Then there exists a numerical constant $C_0 > 0$, such that 
\begin{align*}
\KL (\mu_{\delta}, p_{t_N}) \leq C_0 \cdot d \cdot \left(  \eps_{\rm score}^2 + \kappa^2  N B + \kappa  T B + e^{-2T} B   \right). 
\end{align*}
\end{theorem}


\begin{proof}[Proof sketch of Theorem~\ref{thm:benton}]~

\noindent
{\bf Part 1.} We first control the quantity
\begin{align*}
E_{s, t} = \E\left[ \|\nabla \log \mu_{T - t}(Y_t) - \nabla \log \mu_{T - s}(Y_s)\|_2^2 \right],
\end{align*}
where $0 \leq s \leq t \leq T$. According to Lemma 2 of \cite{benton2023linear}, we have
\begin{align}
&~ \de \left( \|\nabla \log \mu_{T - t} (Y_t) - \nabla \log \mu_{T - s} (Y_s)\|_2^2 \right) \nonumber \\
=&~ -2 \|\nabla \log \mu_{T - t} (Y_t) - \nabla \log \mu_{T - s}(Y_s)\|^2_2 \de t - 2 \left\{ \nabla \log \mu_{T - t}(Y_t) - \nabla \log \mu_{T - s}(Y_s) \right\} \cdot \nabla \log \mu_{T - s}(Y_s) \de t \nonumber \\
&~ + 2 \|\nabla^2 \log \mu_{T - t} (Y_t)\|_F^2 \de t + 2\sqrt{2} \left\{ \nabla \log \mu_{T - t}(Y_t) - \nabla \log \mu_{T - s}(Y_s) \right\} \cdot \nabla^2 \log \mu_{T - t}(Y_t) \cdot \de B_t'. \label{eq:lemma2}
\end{align}
In the above display, $s$ is fixed and $t$ varies. 
Taking expectation and integrate over $[s, t]$,  we obtain
\begin{align*}
& \E\left[ \|\nabla \log \mu_{T - t} (Y_t) - \nabla \log \mu_{T - s}(Y_s)\|^2 \right]  = \E \int_s^t -2 \|\nabla \log \mu_{T - r} (Y_r) - \nabla \log \mu_{T - s}(Y_s)\|^2_2 \de r \\
& - \E \int_s^t 2 \left\{ \nabla \log \mu_{T - r}(Y_r) - \nabla \log \mu_{T - s}(Y_s) \right\} \cdot \nabla \log \mu_{T - s}(Y_s) \de r + \E \int_s^t 2 \|\nabla^2 \log \mu_{T - r} (Y_r)\|_F^2 \de r. 
\end{align*}
Observe that all terms above are integrable. Hence,
we may apply Fubini's theorem and interchange integration and expectation, which gives
\begin{align*}
&  \frac{\de E_{s, t}}{\de t} = -2 \E\left[ \|\nabla \log \mu_{T - t} (Y_t) - \nabla \log \mu_{T - s}(Y_s)\|^2_2 \right] \\
& + 2 \E\left[  \left\{ \nabla \log \mu_{T - s}(Y_s) - \nabla \log  \mu_{T - t}(Y_t) \right\} \cdot \nabla \log \mu_{T - s}(Y_s) \right] + 2 \E\left[ \|\nabla^2 \log \mu_{T - t}(Y_t)\|_F^2 \right]. 
\end{align*}
Invoking Cauchy-Schwartz inequality, we have
\begin{align}\label{eq:dEst}
\frac{\de E_{s, t}}{\de t} \leq \E\left[ \|\nabla \log \mu_{T - s}(Y_s)\|_2^2 \right] + 2\E\left[ \|\nabla^2 \log \mu_{T - t}(Y_t)\|_F^2 \right]. 
\end{align}
Next, we upper bound $\E\left[ \|\nabla \log \mu_{T - s}(Y_s)\|_2^2 \right]$ and $\E\left[ \|\nabla^2 \log \mu_{T - t}(Y_t)\|_F^2 \right]$, respectively.

Lemma 3 of \cite{benton2023linear} gives
\begin{align}\label{eq:lemma3}
\nabla \log \mu_t(\bx_t) = -\sigma_t^{-2} \bx_t + e^{-t} \sigma_t^{-2} \bbm_t(\bx_t), \qquad \nabla^2 \log \mu_t(\bx_t) = - \sigma_t^{-2} \id + e^{-2t} \sigma_t^{-4} \bSigma_t(\bx_t), 
\end{align}
where $\sigma_t^2 = 1 - e^{-2t}$, $\bbm_t(\bx_t) = \E_{\mu_0 \mid \mu_t (\bx_0 \mid \bx_t)}[\bx_0]$, and $\bSigma_t(\bx_t) = {\Cov}_{\mu_0 \mid \mu_t (\bx_0 \mid \bx_t)}[\bx_0]$. By Eq.~\eqref{eq:lemma3}, we see that
\begin{align*}
\E_{\bx_t \sim \mu_t} \left[ \|\nabla \log \mu_t(\bx_t)\|_2^2 \right] = \sigma_t^{-4} \E_{\bx_t \sim \mu_t}\left[ \|\bx_t\|_2^2 \right] - 2 e^{-t} \sigma_t^{-4} \E_{\bx_t \sim \mu_t}\left[ \bx_t \cdot \bbm_t(\bx_t) \right] + e^{-2t} \sigma_t^{-4} \E_{\bx_t \sim \mu_t}\left[ \|\bbm_t(\bx_t)\|_2^2 \right].
\end{align*}
Note that 
\begin{align*}
& \E_{\bx_t \sim \mu_t}[\bx_t \cdot \bbm_t(\bx_t)] = \E_{\bx_t \sim \mu_t}[\bx_t \cdot \bx_0] = e^{-t} \E_{\bx_0 \sim \mu_0}\left[ \|\bx_0\|^2\right] \leq dB e^{-t}, \\
& \Tr(\bSigma_t (\bx_t)) = \E[\|\bx_0\|^2 \mid \bx_t] - \|\bbm_t(\bx_t)\|_2^2, 
\end{align*}
hence 
\begin{align}\label{eq:B13}
&~ \E_{\bx_t \sim \mu_t} \left[\|\nabla \log  \mu_t(\bx_t)\|^2 \right] \nonumber \\ =&~  \sigma_t^{-4} \cdot \left( e^{-2t} \E[\|\bx_0\|^2] + \sigma_t^2 d \right) - 2e^{-2t} \sigma_t^{-4} \E[\|\bx_0\|^2] + e^{-2t} \sigma_t^{-4} \cdot \left( \E[\|\bx_0\|^2] - \E[\Tr(\bSigma_t(\bx_t))] \right)  \\
= &~ \sigma_t^{-2} d - e^{-2t} \sigma_t^{-4} \E[\Tr(\bSigma_t(\bx_t))] \leq d \sigma_t^{-2}. \nonumber
\end{align}
That is to say, we have $\E[\|\nabla \log \mu_{T - s}(Y_s)\|_2^2] \leq d \sigma_{T - s}^{-2}$. We write $\bSigma_t = \bSigma_t(\bx_t)$ for short.
The second part of Eq.~\eqref{eq:lemma3} implies that
\begin{align}\label{eq:B14}
\E_{\bx_t \sim \mu_t}\left[ \|\nabla^2 \log \mu_t(\bx_t)\|_F^2 \right] = \sigma_t^{-4} d- 2 \sigma_t^{-6} e^{-2t} \E\left[ \Tr(\bSigma_t) \right] + e^{-4t} \sigma_t^{-8} \E[\Tr(\bSigma_t^2)].\end{align} 
Lemma 1 of \cite{benton2023linear} gives 
\begin{align}\label{eq:B15}
\frac{e^{2t}\sigma_t^4}{2 } \frac{\de}{\de t} \E\left[\bSigma_t \right] = \E[\bSigma_t^2]. 
\end{align}
Putting together Eq.~\eqref{eq:B14} and \eqref{eq:B15}, we obtain
\begin{align}\label{eq:B16}
\begin{split}
\E_{\bx_t \sim \mu_t}\left[ \|\nabla^2 \log \mu_t(\bx_t)\|_F^2 \right] = &~ d \sigma_t^{-4} - 2 \sigma_t^{-6} e^{-2t} \E\left[ \Tr(\bSigma_t) \right] + \frac{e^{-2t} \sigma_t^{-4}}{2} \frac{\de}{\de t} \E[\Tr[\bSigma_t]] \\
\leq &~ d \sigma_t^{-4} + \frac{1}{2} \frac{\de}{\de t} \left( \sigma_t^{-4} \E[\Tr(\bSigma_t)] \right).
\end{split}
\end{align}
Putting together Eq.~\eqref{eq:B13} and \eqref{eq:B16}, we get
\begin{align*}
&~ \E\left[ \|\nabla \log \mu_{T - s}(Y_s)\|_2^2 \right] + 2\E\left[ \|\nabla^2 \log \mu_{T - t}(Y_t)\|_F^2 \right] \\
\leq &~ \sigma_{T - s}^{-2} d + 2d \sigma_{T - t}^{-4} - \frac{\de}{\de r} \left( \sigma_{T - r}^{-4} \E[\Tr(\bSigma_{T - r})]  \right) \big|_{r = t}. 
\end{align*}
We define
\begin{align*}
E_{s, t}^{(1)} := d \sigma_{T - s}^{-2} + 2d \sigma_{T - t}^{-4}, \qquad E_{s, t}^{(2)} := - \frac{\de}{\de r} \left( \sigma_{T - r}^{-4} \E[\Tr(\bSigma_{T - r})]  \right) \big|_{r = t}. 
\end{align*}
According to Eq.~\eqref{eq:dEst} and notice that $E_{t_k, t_k} = 0$, we have 
\begin{align*}
E_{t_k, t} \leq \int_{t_k}^t \left\{\E\left[ \|\nabla \log \mu_{T - t_k}(Y_{t_k})\|_2^2 \right] + 2\E\left[ \|\nabla^2 \log \mu_{T - s}(Y_s)\|_F^2 \right] \right\} \de s \leq \int_{t_k}^t \left( E_{t_k, s}^{(1)} + E_{t_k, s}^{(2)} \right) \de s. 
\end{align*}
Following exactly the same procedure as in \cite{benton2023linear}, we conclude that there exists a positive numerical constant $C_0$, such that 
\begin{align*}
\sum_{k = 0}^{N - 1} \int_{t_k}^{t_{k + 1}} \E\left[ \|\nabla \log  \mu_{T - t}(Y_{ t}) - \nabla \log \mu_{T - t_k}(Y_{t_k})\|^2 \right] \leq C_0 (\kappa^2 d N B  + \kappa d T B). 
\end{align*}

\noindent
{\bf Part 2.} We denote by $Q$ the distribution of $Y_{t_N}$ derived from process \eqref{eq:reverse}, and $P^{\mu_T}$ the distribution of process \eqref{eq:B11} at time $t_N$ initialized at $\mu_T$. By Proposition 3 of \cite{benton2023linear}, we obtain
\begin{align*}
\KL(Q \, || \, P^{\mu_T}) \leq \sum_{k = 0}^{N - 1} \int_{t_k}^{t_{k + 1}} \E\left[ \|\nabla \log \mu_{T - t}(Y_{t})- \hat s_{T - t_k}(Y_{t_k})\|^2_2 \right] \de t, 
\end{align*}
which by triangle inequality is no smaller than
\begin{align*}
&~ 2 \sum_{k = 0}^{N - 1} \gamma_k \E\left[ \|\nabla \log \mu_{T - t_k}(Y_{t_k})- \hat s_{T - t_k}(Y_{t_k})\|_2^2 \right] \de t + 2 \sum_{k = 0}^{N - 1} \int_{t_k}^{t_{k + 1}} \E\left[ \|\nabla \log \mu_{T - t_k}(Y_{ t_k}) - \nabla \log \mu_{T - t}(Y_{t})\|_2^2 \right] \\
\leq &~ \, 2 \, d \cdot \eps_{\score}^2 + 2 C_0 (\kappa^2 d N B + \kappa d T B). 
\end{align*}
We denote by $P$ the distribution of process \eqref{eq:B11} at time $t_N$ initialized at $\cN(\bzero, \id_d)$. By Eq.~(19) of \cite{benton2023linear}, we have
\begin{align*}
\KL(Q \, || \, P) = \KL(Q \, || \, P^{\mu_T}) + \KL(\mu_T \, || \, \cN(\bzero, \id_d)).  
\end{align*}
Proposition 4 of \cite{benton2023linear} gives $\KL (\mu_T \, || \, \cN(\bzero, \id_d)) \lesssim d B e^{-2T}$. Putting together the above upper bounds, we arrive at the following conclusion: 
\begin{align*}
\KL (\mu_{\delta} || \, p_{t_N}) \leq C_0 \cdot d \cdot \left(  B e^{-2T} + \kappa^2 N B + \kappa  T B + \eps^2_{\score} \right), 
\end{align*}
thus concluding the proof of Theorem~\ref{thm:benton}. 
\end{proof}

\subsection{Generalization error of empirical risk minimization over ResNets} 

%
%
%
%
%

\subsubsection{Result for Ising models}

Note that the conditional (and unconditional) DDPM methods estimate the score function $\hat\bs_t = \sP_t \resnet_{\what\bW_t}$ by solving the following ERM problem: 
\begin{equation}\label{eqn:ERM-appendix-Ising}
\begin{aligned}
\what\bW_t =&~ \argmin_{\bW \in \cW_{d, m, D, L, M, B}} \what{R}_n(\bW), \\
\what{R}_n(\bW) =&~ \frac{1}{nd}\sum_{i = 1}^n \big\| \sigma_t^{-1} \bg_i + \sP_t (\resnet_{\bW}( \lambda_t \bx_i + \sigma_t \bg_i, \btheta_i )) \big\|_2^2.
\end{aligned}
\end{equation}
Here, $\bx_i, \bg_i \in \R^d$, and $\btheta_i \in \R^m$ follow $\{(\bx_i, \btheta_i, \bz_i)\}_{i \in [n]} \sim_{iid} \mu \otimes \cN(\bzero, \id_d)$. Recall that the truncation operator gives $\sP_{t}[f] (\bz, \btheta) = \proj_{\lambda_t \sigma_t^{-2} \sqrt{d}}(f(\bz, \btheta) + \sigma_t^{-2} \bz ) - \sigma_t^{-2} \bz$. In cases where $\btheta_i$ does not exist (unconditional DDPM), we simply set $m = 0$. 
The population risk gives 
\begin{align*}
R(\bW) := \frac{1}{d} \E_{(\bx, \btheta, \bg) \sim \mu \otimes \cN(\bzero, \id_d)} \Big[ \big\| \sigma_t^{-1} \bg_0 + \sP_t (\resnet_{\bW}( [\lambda_t \bx + \sigma_t \bg, \btheta])) \big\|_2^2 \Big]. 
\end{align*}
%
%
In the proposition below, we provide a uniform upper bound for $| \what{R}(\bW) - R(\bW)|$ over $\cW_{d, m, D, L, M, B}$, where the ResNet class is given by Eq.~(\ref{eqn:conditional-ResNet}). 

\begin{proposition}\label{prop:uniform2}
Assume that $\mu \in \cP([- 1, 1]^{d + m})$. There exists a numerical constant $C > 0$, such that with probability at least $1 - \eta$, 
\begin{align*}
&~\sup_{\bW \in \cW_{d, m, D, L, M, B}} \Big|  
\what{R}(\bW) - R(\bW) \Big|\\ 
\leq &~ C \cdot \frac{ \lambda_t^2}{\sigma_t^4} \cdot  \sqrt{\frac{[ (d + m) D + L D M ] \cdot [ L \cdot \log ( L B (m + d)/ d) + \log(\lambda_t^{-1})] + \log (1 / \eta)}{n}}. 
\end{align*}
\end{proposition} 

\begin{proof}[Proof of Proposition~\ref{prop:uniform2}]

The proof of this proposition uses the following lemma. 
\begin{lemma}[Proposition A.4 of \cite{bai2023transformers}]
\label{lem:uniform-concen}
Suppose that $\{X_{w}\}_{w\in\Theta}$ is a zero-mean random process given by
\begin{align*}
X_w \equiv  \frac1n\sum_{i=1}^n f(z_i;w)-\E_z[f(z;w)],
\end{align*}
where $z_1,\cdots,z_n$ are i.i.d samples from a distribution $\P_z$ such that the following assumption holds:
\begin{enumerate}[topsep=0pt, leftmargin=2em]
\item[(a)] The index set $\Theta$ is equipped with a distance $\rho$ and diameter $B$. Further, assume that for some constant $A$, for any ball $\Theta'$ of radius $r$ in $\Theta$, the covering number admits upper bound $\log N(\Delta; \Theta',\rho)\leq d\log(2Ar/\Delta)$ for all $0<\Delta\leq 2r$.
\item[(b)] For any fixed $w\in\Theta$ and $z$ sampled from $\P_z$, the random variable $f(z;w) - \E_z[f(z; w)]$ is a $\sigma$-sub-Gaussian random variable ($\E[e^{\lambda [f(z; w) - \E_{z'}[f(z'; w)]] }] \le e^{\lambda^2 \sigma^2 / 2}$ for any $\lambda \in \R$).
\item[(c)] For any $w,w'\in\Theta$ and $z$ sampled from $\P_z$, the random variable $f(z;w)-f(z;w')$ is a $\sigma' \rho(w,w')$-sub-Gaussian random variable ($\E[e^{\lambda [f(z; w) - f(z; w')]}] \le e^{\lambda^2 (\sigma')^2 \rho^2(w, w') / 2}$ for any $\lambda \in \R$). 
\end{enumerate}
Then with probability at least $1-\eta$, it holds that
\begin{align*}
\sup_{w\in\Theta} | X_w | \leq C \sigma \sqrt{\frac{d \cdot \log(2 A (1 + B \sigma' / \sigma) )+\log(1/\eta)}{n}},
\end{align*}
where $C$ is a universal constant.
\end{lemma}

In Lemma~\ref{lem:uniform-concen}, we can take $z = (\bg, \bx, \btheta)$, $w = \bW$, $\Theta = \cW_{d, m, D, L, M, B}$, $\rho(w, w') = \nrmps{\bW - \bW'}$, and $f(z_i; w) = d^{-1} \| \sigma_t^{-1} \bg_i + \sP_t (\resnet_{\bW}(\lambda_t \bx_i + \sigma_t \bg_i, \btheta_i)) \|_2^2$. Therefore, to show Proposition~\ref{prop:uniform2}, we just need to apply Lemma~\ref{lem:uniform-concen} by checking (a), (b), (c).

\noindent
{\bf Check (a).} We note that the index set $\Theta = \cW_{d, m, D, L, M, B}$ equipped with $\rho(w, w') = \nrmps{\bW - \bW'}$ has diameter $2B$. Further note that $\cW_{d, m, D, L, M, B}$ has dimension bounded by $4(d + m)D + 2LDM$. According to Example 5.8 of \cite{wainwright_2019}, it holds that $\log N(\Delta; \cW_{d, m, D, L, M, r}, \nrmps{\cdot}) \leq [4(d + m)D + 2LDM] \cdot \log (1 + 2 r / \Delta)$ for any $0 < \Delta \le 2r$. This verifies (a).  

\noindent
{\bf Check (b).} By the definition of the projection operator that $\sP_t[f](\bz) = \proj_{\lambda_t \sigma_t^{-2} \sqrt{d}}(f(\bz) + \sigma_t^{-2} \bz) - \sigma_t^{-2} \bz$ and that $\bz = \lambda_t \bx + \sigma_t \bg$, we have 
\[
\begin{aligned}
0 \le f(z; w) =&~ d^{-1} \| \sigma_t^{-1} \bg + \sP_t (\resnet_{\bW}( \lambda_t \bx + \sigma_t \bg, \btheta)) \|_2^2 \\
=&~ d^{-1} \| - \lambda_t \sigma_t^{-2} \bx + \proj_{\lambda_t \sigma_t^{-2} \sqrt{d}}( \resnet_{\bW_1}( \lambda_t \bx + \sigma_t \bg, \btheta) + \sigma_t^{-2} \bz) \|_2^2 \\
\le&~  4 \lambda_t^2 \sigma_t^{-4}. 
\end{aligned}
\]
As a consequence, $f(z, w) - \E_z[f(z, w)]$ is a $\sigma= 4 \lambda_t^2 \sigma_t^{-4}$ sub-Gaussian random variable. 

\noindent
{\bf Check (c).} Direct calculation yields 
\begin{align*}
&~ | f(z; w_1) - f(z; w_2) |\\
=&~ \frac{1}{d}\Big| \|\sigma_t^{-1} \bg + \sP_t (\resnet_{\bW_1}(\lambda_t \bx + \sigma_t \bg, \btheta))\|_2^2 -  \|\sigma_t^{-1} \bg + \sP_t (\resnet_{\bW_2}(\lambda_t \bx + \sigma_t \bg, \btheta))\|_2^2 \Big| \\
= &~ \frac{1}{d} \Big| \| - \lambda_t \sigma_t^{-2} \bx + \proj_{\lambda_t \sigma_t^{-2} \sqrt{d}}( \resnet_{\bW_1}(\lambda_t \bx + \sigma_t \bg, \btheta) + \sigma_t^{-2} \bz)  \|_2^2  \\
&~ -\|- \lambda_t \sigma_t^{-2} \bx + \proj_{\lambda_t \sigma_t^{-2} \sqrt{d}}( \resnet_{\bW_2}(\lambda_t \bx + \sigma_t \bg, \btheta) + \sigma_t^{-2} \bz)  \|_2^2 \Big| \\
\leq &~ \frac{8 \lambda_t }{\sigma_t^2 \sqrt{d}} \cdot \big\|\proj_{\lambda_t \sigma_t^{-2} \sqrt{d}}( \resnet_{\bW_1}(\lambda_t \bx + \sigma_t \bg, \btheta) + \sigma_t^{-2} \bz) \\
&~- \proj_{\lambda_t \sigma_t^{-2} \sqrt{d}}( \resnet_{\bW_2}(\lambda_t \bx + \sigma_t \bg, \btheta) + \sigma_t^{-2} \bz) \big\|_2 \\
\lesssim &~ \frac{2 \lambda_t L (B^2 + 1)^L  }{\sigma_t^2} \cdot \frac{1}{\sqrt{d}} \Big( \lambda_t \|\bx \|_2 + \sigma_t \| \bg \|_2 + \| \btheta\|_2 \Big) \cdot \nrmps{\bW_1 - \bW_2}.
\end{align*}
Notice that $(\bx, \btheta, \bg) \sim \mu \otimes \cN(\bzero, \id_d)$ and note that $\mu \in \cP([- 1, 1]^{d + m})$, we have that $\| \btheta \|_2 / \sqrt{d}$ is $\sqrt{m / d}$-bounded and is thus $\cO(\sqrt{m / d})$-sub-Gaussian, $\| \bx \|_2 / \sqrt{d}$ is 1-bounded and is thus $\cO(1)$-sub-Gaussian, and $\| \bg \|_2 / \sqrt{d}$ is $\cO(1)$-sub-Gaussian. As a consequence, $f(z; w_1) - f(z; w_2)$ is $\sigma' \rho(w_1, w_2) = C \cdot \lambda_t \sigma_t^{-2} L (B^2 + 1)^L \sqrt{(m + d) / d} \cdot \nrmps{\bW_1 - \bW_2}$ sub-Gaussian.

Therefore, we apply Lemma~\ref{lem:uniform-concen}, and use the fact that 
\begin{align*}
\log(2 (1 + B \sigma' / \sigma) ) =  \log (2 (1 + (C/2)  B \lambda_t^{-1} \sigma_t^2 L (B^2 + 1)^L \sqrt{(m + d ) / d} )) \lesssim L \log(L B  (m + d)/ d) + \log(\lambda_t^{-1}). 
\end{align*}
This concludes the proof of Proposition \ref{prop:uniform2}. 
\end{proof}

\subsubsection{Result for Sparse coding}

In the setting of sparse coding, we assume a fixed dictionary $\bA \in \R^{d \times m}$. The model $\bx \sim \mu$ is given by $\bx = \bA \btheta + \beps$, where $\beps \sim \cN(\mathbf{0}, \tau^2 \id_d)$ is independent of anything else and $\theta_i \sim_{iid} \pi_{\theta} \in \cP([-\Pi, \Pi])$ for $i \in [m]$. Assume that we have $\{ (\bx_i, \bg_i) \}_{i \in [n]} \sim_{iid} \mu \otimes \cN(\bzero, \id_d)$. We are interested in estimating the score function $\hat\bs_t = \bar \sP_t \resnet_{\what\bW_t}$ by solving the following ERM problem: 
\begin{equation}\label{eqn:ERM-appendix-SC}
\begin{aligned}
\what\bW_t =&~ \argmin_{\bW \in \cW_{d, D, L, M, B}} \what{R}_n(\bW), \\
\what{R}_n(\bW) =&~ \frac{1}{nd}\sum_{i = 1}^n \big\| \sigma_t^{-1} \bg_i + \bar\sP_t (\resnet_{\bW}( \lambda_t \bx_i + \sigma_t \bg_i )) \big\|_2^2.
\end{aligned}
\end{equation}
Here, the truncation operator gives $\bar\sP_t[f] (\bz) = \proj_{\sqrt{m}\|\bA\|_{\op}\Pi \cdot \lambda_t (\sigma_t^2 + \tau^2 \lambda_t^2)^{-1} }(f(\bz) + (\sigma_t^2 + \tau^2 \lambda_t^2)^{-1} \bz ) - (\sigma_t^2 + \tau^2 \lambda_t^2)^{-1} \bz$. The corresponding population risk gives
\begin{align*}
R(\bW) := \frac{1}{d} \E_{(\bx, \bg) \sim \mu \otimes \cN(\bzero, \id_d)} \Big[ \big\| \sigma_t^{-1} \bg + \bar\sP_t (\resnet_{\bW}(\lambda_t \bx + \sigma_t \bg)) \big\|_2^2 \Big]. 
\end{align*}
In the proposition below, we provide a uniform upper bound for $| \what{R}(\bW) - R(\bW)|$ over $\cW_{d, D, L, M, B}$ in the sparse coding setup, where the ResNet class is given by Eq.~(\ref{eqn:ResNet-class}).
\begin{proposition}\label{prop:uniform3}
Under the setting of sparse coding stated above, there exists a numerical constant $C > 0$, such that with probability at least $1 - \eta$, for $n \ge \log(2 / \eta)$, we have 
\begin{align*}
&~ \sup_{\bW \in \cW_{d, D, L, M, B}} \left| \what{R}(\bW) - R(\bW) \right| 
\lesssim   \Big( \lambda_t^2\|\bA\|_{\op}^2 \Pi^2  (\tau^{-4} + 1) \frac{m}{d}  + \frac{\lambda_t^2}{\sigma_t^2} (1 + \tau^2) \Big) \\
&~~~~ \times \sqrt{\frac{(dD + LDM) \cdot \left[  T +  L\log (LB) + \log (n m T  (\tau + 1) (\|\bA\|_{\op} \Pi + 1) \tau^{-1}) \right] + \log (2 / \eta)}{n}}. 
\end{align*}
\end{proposition}



\begin{proof}[Proof of Proposition~\ref{prop:uniform3}]
Note that $\{ (\bx_i, \bg_i) \}_{i \in [n]} \sim_{iid} \mu \times \cN(\bzero, \id_d)$ where $\mu$ is the sparse coding model. Then we must have $\bx_i = \bA \btheta_i + \beps_i$ for some $(\btheta_i, \beps_i) \sim_{iid} \pi_0^m \times \cN(\bzero, \tau^2 \id_d)$. Denote $z = (\bg, \bx, \beps)$, $w = \bW$, and 
\[
f(z; w) = d^{-1} \|\sigma_t^{-1} \bg +  \bar\sP_t (\resnet_{\bW}(\lambda_t \bx + \sigma_t \bg))\|_2^2 - d^{-1} \|(\sigma_t^{-1} - \sigma_t (\sigma_t^2 + \tau^2 \lambda_t^2)^{-1}) \bg - \lambda_t  (\sigma_t^2 + \tau^2 \lambda_t^2)^{-1} \beps \|_2^2. 
\]
We further denote $\bz = \lambda_t \bx + \sigma_t \bg$. Note that we have
\begin{align*}
&~ | f(z; w_1) - f(z; w_2) |\\
=&~ \frac{1}{d}\Big| \|\sigma_t^{-1} \bg + \bar\sP_t (\resnet_{\bW_1}(\lambda_t \bx + \sigma_t \bg))\|_2^2 -  \|\sigma_t^{-1} \bg + \bar\sP_t (\resnet_{\bW_2}(\lambda_t \bx + \sigma_t \bg))\|_2^2 \Big| \\
\leq &~ \frac{1}{d} \Big| \| \proj_{\sqrt{m}\|\bA\|_{\op} \Pi \cdot \lambda_t (\sigma_t^2 + \tau^2 \lambda_t^2)^{-1}}(\resnet_{\bW_1}(\bz) + (\sigma_t^2 + \tau^2 \lambda_t^2)^{-1} \bz ) - (\sigma_t^2 + \tau^2 \lambda_t^2)^{-1} \bz + \sigma_t^{-1} \bg \|_2^2  \\
&~ -\| \proj_{\sqrt{m}\|\bA\|_{\op} \Pi \cdot \lambda_t (\sigma_t^2 + \tau^2 \lambda_t^2)^{-1}}(\resnet_{\bW_2}(\bz) + (\sigma_t^2 + \tau^2 \lambda_t^2)^{-1} \bz ) - (\sigma_t^2 + \tau^2 \lambda_t^2)^{-1} \bz + \sigma_t^{-1} \bg  \|_2^2 \Big| \\
\lesssim &~ \left( \frac{\sqrt{m}\lambda_t \Pi \|\bA\|_{\op}}{d(\sigma_t^2 + \tau^2\lambda_t^2)} + \frac{\lambda_t}{d(\sigma_t^2 + \tau^2\lambda_t^2)}\|\beps\|_2 +  \frac{\tau^2\lambda_t^2}{d \sigma_t(\sigma_t^2 + \tau^2 \lambda_t^2)}\|\bg\|_2 \right)  \\
&~ \times \big\|\proj_{\sqrt{m}\|\bA\|_{\op} \Pi \cdot \lambda_t (\sigma_t^2 + \tau^2 \lambda_t^2)^{-1}}(\resnet_{\bW_1}(\bz) + (\sigma_t^2 + \tau^2 \lambda_t^2)^{-1} \bz ) \\
&~- \proj_{\sqrt{m}\|\bA\|_{\op} \Pi \cdot \lambda_t (\sigma_t^2 + \tau^2 \lambda_t^2)^{-1}}(\resnet_{\bW_2}(\bz) + (\sigma_t^2 + \tau^2 \lambda_t^2)^{-1} \bz ) \big\|_2 \\
\lesssim &~ \frac{L(B^2 + 1)^L\lambda_t}{(\sigma_t^2 + \tau^2 \lambda_t^2)} \cdot \left( \frac{\sqrt{m} \Pi \|\bA\|_{\op}}{\sqrt{d}} + \frac{\|\beps\|_2}{\sqrt{d}} + \frac{\tau^2 \lambda_t \, \|\bg\|_2}{\sqrt{d}\sigma_t} \right) \cdot \frac{1}{\sqrt{d}} \Big( \lambda_t \|\bA \btheta \|_2 + \lambda_t \|\beps\|_2 + \sigma_t \| \bg \|_2 \Big) \cdot \nrmps{\bW_1 - \bW_2}.
\end{align*}
Therefore, we denote by $\cN( \Delta; \cW_{d, D, L, M, B}, \rho)$ a $\Delta$-covering of $\cW_{d, D, L, M, B}$ under metric $\rho(\bW_1, \bW_2) = \nrmps{\bW_1 - \bW_2}$ for some $\Delta > 0$. Then 
\begin{align*}
&~ \sup_{w \in \cW_{d, D, L, M, B}} \left| \frac{1}{n} \sum_{i = 1}^n f(z_i; w) - \E[f(z; w)] \right| \\
\leq &~ \sup_{w \in \cN( \Delta; \cW_{d, D, L, M, B}, \rho)} \left| \frac{1}{n} \sum_{i = 1}^n f(z_i; w) - \E[f(z; w)] \right| + \frac{L(B^2 + 1)^L\lambda_t}{(\sigma_t^2 + \tau^2 \lambda_t^2)} \cdot \Delta \cdot (L_n + \E[L_n]), 
\end{align*}
where 
\begin{align*}
L_n = \frac{1}{nd} \sum_{i = 1}^n \left( {\sqrt{m} \Pi \|\bA\|_{\op}} + {\|\beps_i\|_2} + {\sigma_t^{-1}\tau^2 \lambda_t \, \|\bg_i\|_2} \right) \cdot \Big( \lambda_t \|\bA \btheta_i \|_2 + \lambda_t \|\beps_i\|_2 + \sigma_t \| \bg_i \|_2 \Big). 
\end{align*}
Since $(\btheta_i, \beps_i, \bg_i) \sim \pi_0^m \otimes \cN(\bzero, \tau^2 \id_d) \otimes \cN(\bzero, \id_d)$, $\sigma_t^2 \le 1$ and $\lambda_t^2 \le 1$, we have $\E[L_n] \le \overline L$ and $L_n - \E[L_n]$ is ${\rm SE}(\overline L / \sqrt{n}, \overline L)$, for $\overline L = (m / d)\Pi^2 \|\bA\|_{\op}^2 +  \sigma_t^{-2} (\tau^4 + 1)$. 
By Bernstein's inequality, we conclude that with probability at least $1 - \eta / 2$, we have
\begin{align*}
L_n + \E[L_n] \leq&~ C 
\cdot \overline L (1 + \sqrt{\log(2/\eta) / n} + \log (2 / \eta) / n) \le C \cdot \overline L (1 + \log(2 / \eta)) \\
=&~ C \cdot \left( ( m / d) \Pi^2 \|\bA\|_{\op}^2 +  \sigma_t^{-2} (\tau^4 + 1) \right) \cdot ( 1 + \log(2 / \eta) ).
\end{align*}
for some numerical constant $C$. 

Furthermore, note that we have 
\begin{align*}
&~ f(z; w)  \\
= &~  d^{-1} \|\sigma_t^{-1} \bg +  \bar\sP_t (\resnet_{\bW}(\lambda_t \bx + \sigma_t \bg))\|_2^2 - d^{-1} \|(\sigma_t^{-1} - \sigma_t (\sigma_t^2 + \tau^2 \lambda_t^2)^{-1}) \bg - \lambda_t  (\sigma_t^2 + \tau^2 \lambda_t^2)^{-1} \beps\|_2^2  \\
= &~  d^{-1} \|\sigma_t^{-1} \bg +  \proj_{\sqrt{m}\|\bA\|_{\op} \Pi \cdot \lambda_t (\sigma_t^2 + \tau^2 \lambda_t^2)^{-1}}(\resnet_{\bW}(\bz) + (\sigma_t^2 + \tau^2 \lambda_t^2)^{-1} \bz ) - (\sigma_t^2 + \tau^2 \lambda_t^2)^{-1} \bz\|_2^2  \\
&~ - d^{-1} \|(\sigma_t^{-1} - \sigma_t (\sigma_t^2 + \tau^2 \lambda_t^2)^{-1}) \bg - \lambda_t  (\sigma_t^2 + \tau^2 \lambda_t^2)^{-1} \beps\|_2^2 \\
= &~ d^{-1} \|\proj_{\sqrt{m}\|\bA\|_{\op} \Pi \cdot \lambda_t (\sigma_t^2 + \tau^2 \lambda_t^2)^{-1}}(\resnet_{\bW}(\bz) + (\sigma_t^2 + \tau^2 \lambda_t^2)^{-1} \bz ) - \lambda_t(\sigma_t^2 + \tau^2 \lambda_t^2)^{-1}\bA \btheta \|_2^2 \\
&~ + 2d^{-1} \big\langle (\sigma_t^{-1} - \sigma_t (\sigma_t^2 + \tau^2 \lambda_t^2)^{-1}) \bg - \lambda_t  (\sigma_t^2 + \tau^2 \lambda_t^2)^{-1} \beps, \\
&~~~~ \proj_{\sqrt{m}\|\bA\|_{\op} \Pi \cdot \lambda_t (\sigma_t^2 + \tau^2 \lambda_t^2)^{-1}}(\resnet_{\bW}(\bz) + (\sigma_t^2 + \tau^2 \lambda_t^2)^{-1} \bz ) - \lambda_t(\sigma_t^2 + \tau^2 \lambda_t^2)^{-1}\bA \btheta \big\rangle. 
\end{align*}
As a consequence, $f(z; w) - \E_z[f(z, w)]$ is sub-Gaussian with variance proxy
\begin{align*}
C^2 \cdot \left( \frac{m \|\bA\|_{\op}^2 \Pi^2 \lambda_t^2}{d (\sigma_t^2 + \tau^2 \lambda_t^2)^2} + \frac{\tau^2 \lambda_t^2}{\sigma_t^2 (\sigma_t^2 + \lambda_t^2 \tau^2)}\right)^2
\end{align*}
for some other numerical constant $C$. Therefore, with probability at least $1 - \eta / 2$, by sub-Gaussian tail bound and by the bound $\log | \cN(\Delta; \cW_{d, D, L, M, B}, \rho)| \leq [4 d D + 2LDM] \cdot \log (1 + 2 B / \Delta)$, we have
\begin{align*}
&~ \sup_{\bW \in \cN( \Delta; \cW_{d, D, L, M, B}, \rho)} \left| \frac{1}{n} \sum_{i = 1}^n f(z_i; w_i) - \E[f(z; w)] \right| \\
\lesssim &~ \Big( \frac{m \|\bA\|_{\op}^2 \Pi^2 \lambda_t^2}{d (\sigma_t^2 + \tau^2 \lambda_t^2)^2} + \frac{\tau^2 \lambda_t^2}{\sigma_t^2 (\sigma_t^2 + \lambda_t^2 \tau^2)} \Big) \cdot \sqrt{\frac{   [4dD + 2LDM] \cdot \log (1 + 2 B / \Delta) + \log(2 / \eta)}{n}}. 
\end{align*}
Setting 
\begin{align*}
\Delta = \Big( \frac{m \|\bA\|_{\op}^2 \Pi^2 \lambda_t^2}{d (\sigma_t^2 + \tau^2 \lambda_t^2)^2} + \frac{\tau^2 \lambda_t^2}{\sigma_t^2 (\sigma_t^2 + \lambda_t^2 \tau^2)} \Big) \cdot \frac{(\sigma_t^2 + \tau^2 \lambda_t^2)}{n L(B^2 + 1)^L\lambda_t \cdot  \left( {md^{-1}\Pi^2 \|\bA\|_{\op}^2} +  \sigma_t^{-2} (\tau^4 + 1) \right) },
\end{align*}
we conclude that with probability at least $1 - \eta$, when $n \ge \log(2 / \eta)$, we have
\begin{align*}
&~  \sup_{\bW \in \cN(\cW_{d, D, L, M, B}, \rho, \Delta)} \left| \frac{1}{n} \sum_{i = 1}^n f(z_i; w_i) - \E[f(z; w)] \right| \\
\lesssim &~  n^{-1} \cdot \Big( \frac{m \|\bA\|_{\op}^2 \Pi^2 \lambda_t^2}{d (\sigma_t^2 + \tau^2 \lambda_t^2)^2} + \frac{\tau^2 \lambda_t^2}{\sigma_t^2 (\sigma_t^2 + \lambda_t^2 \tau^2)} \Big) \cdot (\log(2 / \eta) + 1)  + \left\{ \frac{m \|\bA\|_{\op}^2 \Pi^2 \lambda_t^2}{d (\sigma_t^2 + \tau^2 \lambda_t^2)^2} + \frac{\tau^2 \lambda_t^2}{\sigma_t^2 (\sigma_t^2 + \lambda_t^2 \tau^2)} \right\}  \\
&~ \times \sqrt{\frac{(dD + LDM) \cdot \left[  T +  L\log (LB) + \log (n m T  (\tau + 1) (\|\bA\|_{\op} \Pi + 1) \tau^{-1})  \right] + \log(2 / \eta)}{n}} \\
\lesssim&~  \Big( \lambda_t^2 \|\bA\|_{\op}^2 \Pi^2  (\tau^{-4} + 1) \frac{m}{d} + \frac{\lambda_t^2}{\sigma_t^2} (1 + \tau^2) \Big)  \\
&~ \times \sqrt{\frac{(dD + LDM) \cdot \left[  T +  L\log (LB) + \log (n m T  (\tau + 1) (\|\bA\|_{\op} \Pi + 1) \tau^{-1}) \right] + \log(2 / \eta)}{n}},
\end{align*}
where the inequalities above uses the definition that $\lambda_t = e^{-t}$, $\sigma_t^2 = 1 - e^{-2t}$ and $t \le T$. This concludes the proof of Proposition~\ref{prop:uniform3}. 
\end{proof}


\subsection{Uniform approximation of the denoiser}

The lemma below tells us that denoiser functions can be uniformly approximated with a linear combination of $\relu(\cdot)$ with changing intercepts. Furthermore, such approximation can achieve arbitrary precision. 

\begin{lemma}\label{lemma:approx-tanh}
Assume $\pi_0$ is a probability distribution over $\R$ that has bounded support, and $\gamma > 0$ is a fixed constant. 
Define $F(\lambda) := \E_{(\beta, z) \sim \pi_0 \otimes \cN(0,1)}[\, \beta \mid \beta + \gamma^{-1/2} z = \lambda \gamma^{-1}]$. 
Let $\Pi_{\min} := \inf_{\lambda} F(\lambda)$, $\Pi_{\max} := \sup_{\lambda} F(\lambda)$, $\Pi := \max\{\,|\Pi_{\max}|, \, |\Pi_{\min}|\,\}$, and $\Delta := \Pi_{\max} - \Pi_{\min}$. One can verify that $F(\cdot)$ is $\Pi^2$-Lipschitz continuous and non-decreasing.  
For any $\zeta > 0$, we define
\begin{equation}\label{eqn:w-zeta-definition}
w_{\zeta} := \inf\Big\{w: \mbox{ for all $\lambda_1 > \lambda_2 \geq w$ or $\lambda_1 < \lambda_2 \leq -w$ we have $|F(\lambda_1) - F(\lambda_2)| < \Delta / \lceil \Delta \zeta^{-1} \rceil$ } \Big\}. 
\end{equation}
Then there exists $\{a_j\}_{j \in \{0\} \cup [\lceil\Delta \zeta^{-1}\rceil - 1]}$ and $\{w_j\}_{j \in [\lceil\Delta \zeta^{-1}\rceil - 1]}$, such that 
\begin{align}\label{eq:uniform-G}
\sup_{\lambda \in \R} \left| F(\lambda) - f(\lambda) \right| \leq \zeta,~~~~~\text{where}~~~~ f(\lambda) = \sum_{j = 1}^{\lceil\Delta \zeta^{-1}\rceil - 1} a_j \relu (\lambda - w_j) + a_0. 
\end{align}
Furthermore, we have $\sup_{j \in [\lceil \Delta \zeta^{-1} \rceil - 1]}|w_j| \leq w_{\zeta}$, $|a_0| \leq \Pi$, and $|a_j| \leq 2 \Pi^2$ for all $j \in [\lceil \Delta \zeta^{-1}\rceil - 1]$.   
\end{lemma}
\begin{proof}[Proof of Lemma~\ref{lemma:approx-tanh}]
When $\pi_0$ is a Dirac measure, we simply take $a_0 = \E[\beta]$. In other cases, one can verify that $F(\cdot)$ is strictly increasing, hence $\Pi_{\max} > \Pi_{\min}$.  Then for any $\alpha \in (\Pi_{\min}, \Pi_{\max})$, there exists a unique $\mu_{\alpha} \in \R$, such that $F(\mu_{\alpha}) = \alpha$. 

Let $a_0 = \Pi_{\min} + \Delta \lceil \Delta \zeta^{-1} \rceil^{-1}$. For $j \in [[\Delta \zeta^{-1}] - 1]$, we let 
\begin{align*}
w_j = \mu_{-\Pi_{\min} + j \Delta / \lceil \Delta \zeta^{-1} \rceil}, \qquad a_j = \frac{\Delta}{\lceil \Delta \zeta^{-1} \rceil (w_{j + 1} - w_j)} - \frac{\Delta}{\lceil \Delta \zeta^{-1} \rceil(w_j - w_{j - 1})}.  
\end{align*}
In the above equations, we make the convention that $w_0 = w_{\lceil \Delta \zeta^{-1}\rceil} = \infty$. With $\{a_j\}_{j \in \{0\} \cup [\lceil \Delta \zeta^{-1} \rceil - 1]}$ and $\{w_j\}_{j \in [\lceil \Delta \zeta^{-1}\rceil - 1]}$ defined as above, one can verify that Eq.~\eqref{eq:uniform-G} is true. 
Furthermore, since $\|F'\|_{\infty} \leq \Pi^2$, we have $|{\Delta} / {\lceil \Delta \zeta^{-1} \rceil (w_{j + 1} - w_j)}| \leq \Pi^2$ for all possible $j$. This gives $|a_j| \leq 2 \Pi^2$ for every $j$. 
%

%

%
\end{proof}

\begin{remark}\label{rmk:approx-tanh-lem}
When $\pi_0 = \Unif(\{ \pm 1\})$, one can check that for any $\gamma > 0$, we have $F(x) = \tanh(x)$. In this case, one can verify that $|w_{\zeta}| \leq \log \lceil \zeta^{-1} \rceil$. In addition, we can further guarantee that $\sum_{j \in [\lceil\Delta \zeta^{-1}\rceil - 1]} | a_j | \le 2$. 
\end{remark}

\subsection{Approximation error of fixed point iteration}

\begin{lemma}\label{lem:approximate_TAP_iteration_Ising}
Assume that $\bh \in \R^d$, $\bU \in \R^{d \times d}$ with $\|\bU\|_{\op} \leq A < \Pi^{-2}$ for some $\Pi > 0$. Further assume that $f_{\ast}: \R \mapsto \R$ is $\Pi^2$-Lipschitz continuous and $f: \R \mapsto \R$ is a function satisfying 
\begin{align}\label{eq:914-1}
\sup_{u \in \R} |f(u) - f_{\ast}(u)| \leq \zeta. 
\end{align}
Let $\hat\bbm \in \R^d$ satisfying $\|\hat\bbm\|_2 \leq \Pi \sqrt{d}$ be the unique fixed point of 
\begin{align}\label{eq:914-2}
\hat\bbm = f_{\ast} (\bU \hat\bbm + \bh). 
\end{align}
Let $\tilde \bbm^0 = \bzero$ and 
\begin{align}\label{eq:914-3}
\tilde \bbm^k = f(\bU \tilde \bbm^{k - 1} + \bh).
\end{align}
Then we have
\begin{align}\label{eq:914-4}
\frac{1}{\sqrt{d}}\|\tilde \bbm^k - \hat\bbm\|_2 \leq \Pi \cdot (\Pi^2 A)^k + \frac{\zeta}{1 - \Pi^2 A}. 
\end{align}
\end{lemma}
\begin{proof}[Proof of Lemma \ref{lem:approximate_TAP_iteration_Ising}]
By Eq.~\eqref{eq:914-1} and \eqref{eq:914-3}, we have 
\begin{align*}
\tilde \bbm^k = f_{\ast} (\bU \tilde \bbm^{k - 1} + \bh) + \bzeta^k,
\end{align*}
where $\|\bzeta^k\|_2 \leq \sqrt{d} \zeta$. Comparing with Eq.~\eqref{eq:914-2}, we get
\begin{align*}
\|\tilde \bbm^k - \hat\bbm\|_2 \leq \Pi^2 \|\bU\|_{\op} \|\tilde \bbm^{k - 1} - \hat\bbm\|_2 + \|\bzeta^k\|_2 \leq \Pi^2 A \cdot \|\tilde \bbm^{k - 1} - \hat\bbm\|_2 + \sqrt{d} \zeta. 
\end{align*}
By the fact that $\|\tilde \bbm^0 - \hat\bbm\|_2 = \|\hat\bbm\|_2 \leq \Pi \sqrt{d}$, this gives Eq.~\eqref{eq:914-4}, which concludes the proof of the lemma. 
\end{proof}



\subsection{Properties of two-phase time discretization scheme}

The lemma below provides a bound related to the two-phase time discretization scheme that appears to be useful when deriving the sampling error bound. 

\begin{lemma}\label{lem:gamma-lambda-sigma-inequality}
Consider the two-phase discretization scheme $(\kappa, N_0, N, T, \delta, \{ t_k\}_{0 \le k \le N})$ and recall that $\gamma_k = t_{k+1} - t_k$ (Definition \ref{def:two-phase-discretization-scheme}). Recall the definition $\lambda_t = e^{-t}$ and $\sigma_t^2 = 1 - e^{-2t}$. Then we have
\begin{align}
\sum_{0 \le k \le N-1} \gamma_k \cdot \lambda_{T - t_k}^2 \sigma_{T - t_k}^{-4} \lesssim&~ 1 + \delta^{-1}. \label{eqn:gamma-lambda-sigma-inequality-1}
\end{align}
\end{lemma}

\begin{proof}[Proof of Lemma~\ref{lem:gamma-lambda-sigma-inequality}]

Simple algebra yields
\[
\sigma_t^{-2} = 1/ [1 - e^{-2t}] \le 10 \cdot  [ 1 \vee (1/ t)]. 
\]
Note that $T - t_k \le 1$ for all $k \ge N_0$ and $T - t_k \ge 1$ for all $k \le N_0 -1$ (c.f. Definition \ref{def:two-phase-discretization-scheme} for $N_0$). Then the summation in the first phase has bound (we use the fact that $\kappa < 1$)
\[
\sum_{0 \le k \le N_0 -1} \gamma_k \cdot \lambda_{T - t_k}^2 \sigma_{T - t_k}^{-4} \le 100 \kappa \sum_{0 \le k \le M - 1} e^{- 2 (T - t_k)} \le 100 \kappa e^{-2} \sum_{k \ge 0} e^{-2 k \kappa} \le 100 \kappa e^{-2} \frac{1}{1 - e^{-2 \kappa}} \le 100. 
\]
Furthermore, the summation in the second phase yields (recall from Definition \ref{def:two-phase-discretization-scheme} that for $k \ge N_0$, we have $T - t_{N_0 + k} = (1 + \kappa)^{-k}$, $\gamma_{N_0 + k} = \kappa / (1 + \kappa)^{k+1}$, and $\delta = (1 + \kappa)^{N_0 - N}$)
\[
\begin{aligned}
&~\sum_{N_0  \le k \le N -1} \gamma_k \lambda_{T - t_k}^2 \sigma_{T - t_k}^{-4} \le 100 \sum_{N_0 \le k \le N -1} \gamma_k /(T - t_k)^2 \\
=&~ 100 \sum_{0 \le k \le N -N_0 - 1} [\kappa / (1 + \kappa)^{k+1}] \cdot (1 + \kappa)^{2k} = 100 \frac{\kappa}{\delta} \sum_{0 \le k \le N -N_0 - 1} (1 + \kappa)^{-2 - k} \\
\le&~ 100 \frac{\kappa}{\delta}\sum_{k = 1}^\infty (1 + \kappa)^{-k} = 100/\delta. 
\end{aligned}
\]
Combining the two inequalities above proves Eq.~(\ref{eqn:gamma-lambda-sigma-inequality-1}) and concludes the proof. 
\end{proof}

\section{Proofs for Section~\ref{sec:Ising-new}: Ising models}
\label{app:Ising-new}

\subsection{Proof of Theorem~\ref{thm:Ising-score-approximation}}

\subsubsection*{Approximate the minimizer of the free energy via an iterative algorithm}

We first show that we can approximate the minimizer of $\cF_t^{\rm VI}$ using a simple iterative algorithm. Calculating the Hessian of $\cF_t^{\rm VI}$, we obtain
\begin{align*}
\nabla^2_{\bbm} \cF_t^{\rm VI} (\bbm; \bz) = \diag\{(1 - m_i^2)_{i \in [d]}\} - \bA + \bK \succeq (1 - A) \cdot \id_d \succ 0,~~~~ \forall \bbm \in [-1, 1]^d,
\end{align*}
where the inequalities are due to the fact that $\diag\{(1 - m_i^2)_{i \in [d]}\} \succeq \id_d$ and the assumption that $\|\bK - \bA\|_{\op} \leq A < 1$. Therefore, $\cF_t^{\rm VI}(\cdot, \bz)$ is strongly convex in its first coordinate for all $\bz \in \R^d$, hence the critical equation
\begin{align*}
\nabla_{\bbm} \cF_t^{\rm VI}(\bbm; \bz) = \tanh^{-1} (\bbm) - \bA \bbm - \lambda_t \sigma_t^{-2} \bz + \bK \bbm = \bzero, 
\end{align*}
can have at most one solution on $[-1, 1]^d$. Furthermore, $\nabla_{\bbm} \cF_t^{\rm VI} (\bbm; \bz) = \bzero$ is equivalent to the fixed point equation
\[
\bbm = \tanh((\bA - \bK) \bbm + \lambda_t \sigma_t^{-2} \bz),
\]
and $T(\bbm) = \tanh((\bA - \bK) \bbm + \lambda_t \sigma_t^{-2} \bz)$ is a continuous mapping from $[-1, 1]^d$ to itself. Therefore, there exists a solution of $\bbm = T(\bbm)$ by Brouwer's fixed-point theorem. This implies that the above fixed point equation has a unique solution $\hat\bbm_t(\bz) \in [-1, 1]^d$.

Take $f: \R \to \R$ to be the function as derived by Lemma~\ref{lemma:approx-tanh} achieving $\zeta$-uniform approximation to $\tanh(\cdot)$. We write $f(x) = \sum_{j = 1}^{\lceil 2\zeta^{-1} \rceil - 1} a_j \relu (x - w_j) + a_0$. Define iterative algorithm $\{ \tilde \bbm^\ell \}_{\ell \ge 0}$ by
\begin{align}\label{eqn:iterative-algorithm-Ising-approximate}
\tilde\bbm^0 = \bzero, \qquad \tilde\bbm^\ell(\bz) =  \tilde\bbm^\ell =  f((\bA - \bK) \tilde\bbm^{\ell - 1} + \lambda_t \sigma_t^{-2} \bz). 
\end{align}
Then by Lemma~\ref{lem:approximate_TAP_iteration_Ising} with $\Pi = 1$, we obtain that 
\begin{align}\label{eqn:m-approximation-bound-in-proof-Ising}
\|\tilde \bbm^\ell(\bz) - \hat\bbm_t(\bz)\|_2 / \sqrt{d} \leq A^\ell + \zeta \cdot (1 - A)^{-1}.
\end{align}

\subsubsection*{Represent the iterative algorithm as a ResNet}

Next, we show that $\tilde\bbm^{\ell}(\bz)$ defined as above takes the form of a  ResNet. 

\begin{lemma}\label{lem:alg-to-net}
For all $\ell \in \N_+$ and $\delta \leq t \leq T$, there exists $\bW \in \cW_{d, D, \ell, M, B}$ with 
\[
\begin{aligned}
&~D = 3d, ~~~~~M = (\lceil 2\zeta^{-1} \rceil + 3)d, \\
&~B = (\lceil 2 \zeta^{-1} \rceil - 1)(4 + \log \lceil \zeta^{-1} \rceil ) + 8 + (1 - e^{-2\delta})^{-1} + \sqrt{d},
\end{aligned}
\]
such that $(\lambda_t \tilde \bbm^{\ell}(\bz) - \bz) / \sigma_t^2 = \resnet_{\bW}(\bz)$, where $\tilde \bbm^\ell$ is as defined in Eq.~(\ref{eqn:iterative-algorithm-Ising-approximate}). 
\end{lemma}

\begin{proof}[Proof of Lemma~\ref{lem:alg-to-net}]
Recall the definition of $f$ as an approximation of $\tanh$ as in Lemma \ref{lemma:approx-tanh}. Recall that a ResNet takes the form \eqref{eqn:relu-resnet}. We shall choose the weight matrices appropriately such that $\bu^{(\ell)} = [\tilde \bbm^\ell;  \sigma_t^{-2} \bz;  \ones_d]^{\sT} \in \R^{3d}$. In particular, for $\ell = 0$, we set
\begin{align*}
\bW_{\inside} = \left[ \begin{array}{ccc}
\bzero_{d \times d}  & \sigma_t^{-2} \id_d & \bzero_{d \times d}  \\
\bzero_{1 \times d}  & \bzero_{1 \times d} & \mathbf{1}_{1 \times d}
\end{array} \right]^{\sT} \in \R^{3d \times (d + 1)}. 
\end{align*}
For $\ell \geq 1$, we set 
\begin{align*}
& \bW_1^{(\ell)} = \left[ \begin{array}{ccccccc}
a_i \id_d  & \cdots & a_{\lceil 2 \zeta^{-1} \rceil - 1} \id_d & -\id_d & \id_d & a_0 \id_d & -a_0 \id_d \\
\bzero_{d \times d} & \cdots & \bzero_{d \times d} & \bzero_{d \times d} & \bzero_{d \times d} & \bzero_{d \times d} & \bzero_{d \times d} \\
\bzero_{d \times d} & \cdots  & \bzero_{d \times d} & \bzero_{d \times d} & \bzero_{d \times d} & \bzero_{d \times d} & \bzero_{d \times d}
\end{array} \right] \in \R^{3d \times (\lceil 2\zeta^{-1} \rceil + 3)d}, \\
& \bW_2^{(\ell)} = \left[ \begin{array}{ccccccc}
\bA - \bK &  \cdots &  \bA - \bK & \id_d & -\id_d &  \bzero_{d \times d} &  \bzero_{d \times d} \\
\lambda_t \id_d     &  \cdots &  \lambda_t \id_d & \bzero_{d \times d} &  \bzero_{d \times d} &  \bzero_{d \times d} &  \bzero_{d \times d} \\
-w_1 \id_d & \cdots &  -w_{\lceil 2\zeta^{-1} \rceil - 1} \id_d &  \bzero_{d \times d} &  \bzero_{d \times d} & \id_d & -\id_d
\end{array} \right]^{\sT} \in \R^{(\lceil 2\zeta^{-1} \rceil + 3)d \times 3d}. 
\end{align*}
Finally, we take $\bW_{\out} = [\lambda_t \sigma_t^{-2} \id_d, -\id_d, \bzero_{d \times d}] \in \R^{d \times 3d}$.

By Lemma~\ref{lemma:approx-tanh} and Remark~\ref{rmk:approx-tanh-lem}, we have $\sum_{j = 1}^{\lceil 2 \zeta^{-1} \rceil - 1} |a_j| \leq 2$, $|a_0| \leq 1$, and $|w_j| \leq \log \lceil \zeta^{-1} \rceil$. Therefore, $\|\bW_{\inside}\|_{\op} \leq \sqrt{d} + \sigma_t^{-2}$, $\|\bW_{\out}\|_{\op} \leq 1 + \lambda_t \sigma_t^{-2}$,  $\|\bW_1^{(\ell)}\|_{\op} \leq 2\lceil 2 \zeta^{-1} \rceil + 2$ and $\|\bW_2^{(\ell)}\|_{\op} \leq (\lceil 2\zeta^{-1} \rceil - 1)(2 + \log \lceil \zeta^{-1} \rceil) + 4$. Hence, $\nrmps{\bW} \leq (\lceil 2\zeta^{-1} \rceil - 1)(4 + \log \lceil \zeta^{-1} \rceil ) + 8 + \sigma_t^{-2} + \sqrt{d}$. Note that for $\delta \leq t \leq T$, it holds that $\sigma_t^{-2} \leq (1 - e^{-2\delta})^{-1}$. Therefore, we have 
%
\begin{align*}
\nrmps{\bW} \le B = (\lceil2 \zeta^{-1} \rceil - 1)(3 + \log \lceil \zeta^{-1} \rceil ) + 8 + (1 - e^{-2\delta})^{-1} + \sqrt{d}. 
\end{align*}
This completes the proof of Lemma~\ref{lem:alg-to-net}. 
\end{proof}

\subsubsection*{Proof of Theorem~\ref{thm:Ising-score-approximation}}

Recall that we have $\hat \bs_t(\bz) = \sP_t[\ResN_{\what \bW}](\bz)$, where $\what \bW = \argmin_{\bW \in \cW} \hat \E[\| \sP_t [\ResN_{\bW}](\bz) + \sigma_t^{-1} \bg  \|_2^2 ]$ for $\cW = \cW_{d, D, L, M, B}$. Here, $\hat \E$ denotes averaging over the empirical data distribution. By standard error decomposition analysis in empirical risk minimization theory, we have:
\[
\begin{aligned}
&~\E[\| \sP_t [\ResN_{\what \bW}](\bz)  + \sigma_t^{-1} \bg\|_2^2] / d 
\le  \inf_{\bW \in \cW} \E[\| \sP_t [\ResN_{\bW}](\bz) + \sigma_t^{-1} \bg  \|_2^2] / d \\
&~~~~~+ 2 \sup_{\bW \in \cW} \Big| \hat \E [ \| \sP_t [\ResN_{\bW}](\bz) + \bsigma_t^{-1} \bg \|_2^2] / d - \E [\| \sP_t [\ResN_{\bW}](\bz) + \bsigma_t^{-1} \bg \|_2^2] / d \Big|. 
\end{aligned}
\]
Furthermore, a standard identity in diffusion model theory shows:
\[
\begin{aligned}
\E[\| \hat \bs_t(\bz)  - \bs_t(\bz)\|_2^2 ] / d = \E[\| \hat \bs_t(\bz)  + \sigma_t^{-1} \bg \|_2^2] / d  + C,~~~~ C = \E[\| \bs_t(\bz) \|_2^2] / d - \E[\| \sigma_t^{-1} \bg \|_2^2]/d.
\end{aligned}
\]
Combining the above yields:
\begin{equation}\label{eqn:error-decomp-Ising-proof}
\E [\| \hat \bs_t(\bz) - \bs_t(\bz) \|_2^2] / d  \le \bar\eps_{\rm app}^2 + \bar \eps_{\rm gen}^2, 
\end{equation}
where $\bar\eps_{\rm app}^2$ is the approximation error and $\bar \eps_{\rm gen}^2$ is the generalization error, 
\[
\begin{aligned}
\bar\eps_{\rm app}^2 =&~ \inf_{\bW \in \cW} \E[\| \sP_t [\ResN_{\bW}](\bz) - \bs_t(\bz)  \|_2^2] / d, \\
\bar \eps_{\rm gen}^2 =&~ 2 \sup_{\bW \in \cW} \Big| \hat \E [ \| \sP_t [\ResN_{\bW}](\bz) + \bsigma_t^{-1} \bg \|_2^2] / d - \E [\| \sP_t [\ResN_{\bW}](\bz) + \bsigma_t^{-1} \bg \|_2^2] / d \Big|. 
\end{aligned}
\]
By Proposition~\ref{prop:uniform2} and take $D = 3d$ and $m = 0$, with probability at least $1 - \eta$, simultaneously for any $t \in \{ T - t_k\}_{0 \le k \le N-1}$, we have
\begin{align}\label{eqn:gen-error-bound-Ising-proof}
\bar \eps_{\rm gen}^2
\lesssim \frac{ \lambda_t^2}{\sigma_t^4} \cdot  \sqrt{\frac{[ d^2 + L d M ] \cdot [ L \cdot \log ( L B ) + \log(\lambda_t^{-1})] + \log (N / \eta)}{n}}. 
\end{align}
To bound $\bar\eps_{\rm app}^2$, by the identity that $\bs_t(\bz) = (\lambda_t \bbm_t(\bz) - \bz) / \sigma_t^2$ and $\sP_t \ResN_{\bW}(\bz) = \proj_{\lambda_t \sigma_t^{-2} \sqrt{d}}(\ResN_{\bW}(\bz) + \sigma_t^{-2} \bz ) - \sigma_t^{-2} \bz$, recalling $\tilde \bbm^L(\bz)$ as defined in Eq.~(\ref{eqn:iterative-algorithm-Ising-approximate}), and by Lemma~\ref{lem:alg-to-net}, we have
\begin{equation}\label{eqn:app-error-bound-Ising-proof}
\begin{aligned}
\bar\eps_{\rm app}^2 =&~ \inf_{\bW \in \cW} \E[\| \sP_t [\ResN_{\bW}](\bz) - \bs_t(\bz)  \|_2^2] / d \\
= &~ \inf_{\bW \in \cW} \E[\| \proj_{\lambda_t \sigma_t^{-2} \sqrt{d}} ( \ResN_{\bW}(\bz) + \sigma_t^{-2} \bz )  - \lambda_t \sigma_t^{-2} \bbm_t(\bz)  \|_2^2] / d \\
\le&~ \E[\| \proj_{\lambda_t \sigma_t^{-2} \sqrt{d}} ( \lambda_t \sigma_t^{-2} \tilde \bbm^L(\bz) )  - \lambda_t \sigma_t^{-2} \bbm_t(\bz)  \|_2^2] / d \\
\lesssim&~ \E[\| \proj_{\lambda_t \sigma_t^{-2} \sqrt{d}} ( \lambda_t \sigma_t^{-2} \tilde \bbm^L(\bz) )  - \proj_{\lambda_t \sigma_t^{-2} \sqrt{d}} ( \lambda_t \sigma_t^{-2} \hat \bbm (\bz) ) \|_2^2 ] / d \\
&~ + \E[\| \proj_{\lambda_t \sigma_t^{-2} \sqrt{d}} ( \lambda_t \sigma_t^{-2} \hat \bbm (\bz) ) - \lambda_t \sigma_t^{-2} \bbm_t(\bz) \|_2^2 ] / d
\end{aligned}
\end{equation}
where the last inequality uses the triangle inequality. By Eq.~(\ref{eqn:m-approximation-bound-in-proof-Ising}) and the $1$-Lipschitzness of $\proj_{\lambda_t \sigma_t^{-2} \sqrt{d}}$, the first quantity in the right-hand side is controlled by 
\begin{equation}\label{eqn:app-error-bound-Ising-proof-first}
\begin{aligned}
&~\E[\| \proj_{\lambda_t \sigma_t^{-2} \sqrt{d}} ( \lambda_t \sigma_t^{-2} \tilde \bbm^L(\bz) )  - \proj_{\lambda_t \sigma_t^{-2} \sqrt{d}} ( \lambda_t \sigma_t^{-2} \hat \bbm (\bz) ) \|_2^2 ] / d \\
\lesssim&~ \frac{\lambda_t^2}{\sigma_t^4} \cdot (A^{2L} + \zeta^2 (1 - A)^{-2})\lesssim \frac{\lambda_t^2}{\sigma_t^4} \cdot \Big(A^{2L} + \frac{d^2}{(1 - A)^2 M^2}\Big),
\end{aligned}
\end{equation}
where the last inequality is by the fact that we can choose $\zeta$ such that $M = d \cdot (\lceil 2 \zeta^{-1} \rceil + 3)$, which gives $\zeta \leq  6d / M$. Furthermore, by Assumption~\ref{ass:Ising-free-energy} and by $\| \hat \bbm(\bz) \|_2 \le \sqrt{d}$, the second quantity in the right-hand side is controlled by 
\begin{equation}\label{eqn:app-error-bound-Ising-proof-second}
\E[\| \proj_{\lambda_t \sigma_t^{-2} \sqrt{d}} ( \lambda_t \sigma_t^{-2} \hat \bbm (\bz) ) - \lambda_t \sigma_t^{-2} \bbm_t(\bz) \|_2^2 ] / d \lesssim  \frac{\lambda_t^2}{\sigma_t^4} \cdot \eps_{{\rm VI}, t}^2(\bA).
\end{equation}
Combining Eq.~(\ref{eqn:error-decomp-Ising-proof}), (\ref{eqn:gen-error-bound-Ising-proof}),  (\ref{eqn:app-error-bound-Ising-proof}), (\ref{eqn:app-error-bound-Ising-proof-first}), (\ref{eqn:app-error-bound-Ising-proof-second}) completes the proof of Theorem~\ref{thm:Ising-score-approximation}.

\subsection{Proof of Corollary~\ref{cor:KL-sampling-bound-Ising}}

Corollary~\ref{cor:KL-sampling-bound-Ising} is a direct consequence of Theorem~\ref{thm:Ising-score-approximation}, Theorem~\ref{thm:benton}, and Lemma~\ref{lem:gamma-lambda-sigma-inequality}.

\subsection{Proofs for Section \ref{sec:Ising-examples}}\label{app:Ising-examples}

\subsubsection{Proof of Lemma~\ref{lem:Ising-VB-consistency}} 

Lemma~\ref{lem:Ising-VB-consistency} is a direct consequence of Lemma~\ref{lem:Ising-VB-consistenct-prelim} below. Given Lemma~\ref{lem:Ising-VB-consistenct-prelim}, Lemma~\ref{lem:Ising-VB-consistency} holds by observing that when $\| \bA \|_{\op} < 1/2$, we have $(1 - \| \bA \|_{\op})^{-2} \le 4$. 

\begin{lemma}\label{lem:Ising-VB-consistenct-prelim}
Let $\bh \in \R^d$, $\bA \in \R^{d \times d}$ be symmetric with $\| \bA \|_{\op} < 1/2$. Consider the Ising model $\mu(\sigma) \propto \exp\{ \< \bsigma, \bA \bsigma\>/2 + \< \bsigma, \bh\> \}$ and denote $\bbm = \E_{\bsigma \sim \mu}[\bsigma]$. Let $\hat \bbm$ be the unique minimizer of the naive VB free energy 
\[
\hat \bbm = \argmin_{\bbm \in [-1, 1]^d} \Big\{ \sum_{i = 1}^d - \sh_{\rm bin}(m_i) - \< \bbm, \bA \bbm\>/2 - \< \bbm, \bh\> \Big\}. 
\]
Then we have 
\[
\frac{1}{d} \| \bbm - \hat \bbm \|_2^2 \le \frac{1}{(1 - 2 \| \bA \|_{\op}) (1 - \| \bA \|_{\op})^2} \frac{\| \bA \|_F^2}{d}. 
\]
\end{lemma}

\begin{proof}[Proof of Lemma~\ref{lem:Ising-VB-consistenct-prelim}]

Denote $\ell_i(\bsigma) = \sum_{j \neq i} A_{ij} \sigma_j + h_i$. Simple calculations yields 
$\E_\mu[\sigma_i | \{ \sigma_j\}_{j \neq i}] = \tanh(\ell_i(\bsigma))$, which implies that 
\[
\E_\mu[ \sigma_i ] = \E_\mu[ \tanh(\ell_i(\bsigma)) ]. 
\]
By the fact that $\sup_{x \in \R} |(\de^2 / \de x^2) \tanh(x) | \le 1$ and by Taylor's expansion, we have
\[
| \E_\mu[ \tanh(\ell_i(\bsigma))] - \tanh(\E_\mu[\ell_i(\bsigma)] ) |^2 \le \Var_\mu( \ell_i(\bsigma) ). 
\]
By Theorem 1 of \cite{eldan2022spectral}, the Ising model satisfies a Poincare's Inequality with Poincare's coefficient to be $1/(1 - 2 \| \bA \|_{\op})$ (we need to translate the Ising model to their setting, which leads to an additional $2$ coefficient in front of $\| \bA \|_{\op}$). Therefore, the Poincare's inequality implies that
\[
\Var_\mu( \ell_i(\bsigma) ) \le \frac{1}{1 - 2 \| \bA \|_{\op}} \sum_{j \neq i} A_{ij}^2. 
\]
Combining the equations above, we get 
\[
\frac{1}{d} \Big\| \bbm - \tanh(\bA \bbm + \bh) \Big\|_2^2 = \frac{1}{d} \sum_{i = 1}^d \big( \E_\mu[\sigma_i] - \tanh(\E_\mu[\ell_i(\bsigma)])\big)^2 \le \frac{1}{1 - 2 \| \bA \|_{\op}} \frac{\| \bA \|_F^2}{d} \equiv \eps^2. 
\]
Furthermore, notice that $\hat \bbm$ is the unique minimizer of the naive VB free energy implies that $\hat \bbm = \tanh(\bA \hat \bbm + \bh)$. Therefore, by the equation above, we get 
\[
\begin{aligned}
\eps \ge&~ \frac{1}{\sqrt{d}} \big\| (\bbm - \hat \bbm) - (\tanh(\bA \bbm + \bh) - \tanh(\bA \hat \bbm + \bh) ) \big\|_2 \\
\ge&~ \frac{1}{\sqrt{d}} \Big( \| \bbm - \hat \bbm  \|_2 - \| \tanh(\bA \bbm + \bh) - \tanh(\bA \hat \bbm + \bh) \|_2 \Big)\\
\ge&~ (1 - \| \bA \|_{\op}) \cdot \frac{1}{\sqrt{d}} \| \bbm - \hat \bbm  \|_2. 
\end{aligned}
\]
Combining the equations above concludes the proof of Lemma~\ref{lem:Ising-VB-consistenct-prelim}. 
\end{proof}

\subsubsection{Proof of Lemma~\ref{lem:SK-TAP-consistency}} 

Lemma~\ref{lem:SK-TAP-consistency} is a direct consequence of the lemma below. 

\begin{lemma}[Lemma 4.10 and Proposition 4.2 of \cite{el2022sampling}]\label{lem:EAMS22-consistency}
Let $\bJ \sim \GOE(d)$ and $\beta < 1/2$. Let $\bx \sim \mu(\bx) \propto \exp\{ \beta \< \bx, \bJ \bx\> / 2 \}$ on $\{ \pm 1\}^d$ and $\bg \sim \cN(\bzero, \id_d)$ independently. Let $\bz = \lambda \bx + \sigma \bg$. Consider the posterior measure 
\[
\mu(\bx | \bz) \propto \exp\{ \beta \< \bx, \bJ \bx\> / 2 + (\lambda /  \sigma^2) \<\bx, \bz \> \}, 
\]
and define $\bbm(\bz) = \sum_{\bx \in \{ \pm 1\}^d} \bx \mu(\bx | \bz)$. Furthermore, consider the TAP free energy 
\[
\cF_{\TAP}(\bbm; \bz, q) = \sum_{i = 1}^d -\sh_{\rm bin}(m_i) - \frac{\beta}{2} \< \bbm, \bJ \bbm\>  - \frac{\lambda}{\sigma^2} \< \bz, \bbm\> + \frac{\beta^2 (1 - q)}{2} \| \bbm \|_2^2,
\]
take $q_\star = q_\star(\beta, \lambda, \sigma)$ to be the unique solution of 
\[
q_\star = \E_{G \sim \cN(0, 1)}\big[\tanh^2(\beta^2 q_\star + (\lambda^2 / \sigma^2) + \sqrt{\beta^2 q_\star + (\lambda^2 / \sigma^2)} G) \big],
\]
and define $\hat \bbm(\bz) = \argmin_{\bbm \in [-1, 1]^d} \cF_{\TAP}(\bbm; \bz, q)$ to be the unique minimizer. Then we have 
\[
\| \bbm(\bz) - \hat \bbm(\bz) \|_2^2 / d \gotop 0.
\]
\end{lemma}

\begin{remark}
We discuss the several seeming differences between Lemma~\ref{lem:EAMS22-consistency} and \cite[Lemma 4.10]{el2022sampling}. 
\begin{itemize}
\item The parameter $\lambda^2 / \sigma^2$ in Lemma~\ref{lem:EAMS22-consistency} maps to the parameter $t$ in  \cite[Lemma 4.10]{el2022sampling}. The variable $(\lambda / \sigma^2) \bz = (\lambda^2 / \sigma^2) \bx + (\lambda / \sigma) \bg$ in Lemma~\ref{lem:EAMS22-consistency} maps to the variable $\by \stackrel{d}{=} t \bx + \sqrt{t} \cdot \bg$ in \cite[Lemma 4.10]{el2022sampling}. 
\item Lemma~\ref{lem:EAMS22-consistency} takes $\hat \bbm(\bz)$ to be the unique global minimizer of $\cF_{\TAP}(\bbm; \bz, q_\star)$, whereas \cite[Lemma 4.10]{el2022sampling} takes $\hat \bbm(\bz)$ to be a particular local minimizer of $\cF_{\TAP}(\bbm; \bz, q_\star)$. However, when $\beta < 1/2$, it can be shown that $\cF_{\TAP}$ is strongly convex with high probability, and hence the local minimizer is the global minimizer with high probability. 
\item \cite[Lemma 4.10]{el2022sampling} is proven under a different joint distribution of $(\bJ, \bz)$. However, \cite[Proposition 4.2]{el2022sampling} shows that the distribution for \cite[Lemma 4.10]{el2022sampling} is contiguous to the distribution for Lemma~\ref{lem:EAMS22-consistency}, and hence the high probability event under the sampling distribution of \cite[Lemma 4.10]{el2022sampling} can be translated to the corresponding high probability event under the sampling distribution of Lemma~\ref{lem:EAMS22-consistency}. 
\end{itemize}
\end{remark}
To prove Lemma~\ref{lem:SK-TAP-consistency}, we take $c_t = \beta^2 (1 - q_t)$ where $q_t$ is the unique solution of 
\[
q_t = \E_{G \sim \cN(0, 1)}\big[\tanh^2(\beta^2 q_t + (\lambda_t^2 / \sigma_t^2) + \sqrt{\beta^2 q_t + (\lambda_t^2 / \sigma_t^2)} G) \big]. 
\]
Hence by Lemma~\ref{lem:EAMS22-consistency}, for any $\beta < 1/2$, we have
\[
\E_{\bz \sim \mu_t} [\| \hat \bbm_t(\bz) - \bbm_t(\bz) \|_2^2 ] / d \gotop 0,~~~~ d \to \infty.
\]
Furthermore, note that $c_t \le \beta^2$ and $\| \beta \bJ \|_{\op} \le 2 \beta + \eps$ with high probability for arbitrarily small $\eps$. This ensures that $\| \beta \bJ - c_t \id_d \|_{\op} \le \| \beta \bJ \|_{\op} + \beta^2 < 1$ when $\beta \le 1/4$. This proves Lemma~\ref{lem:SK-TAP-consistency}.

\section{Proofs for Section~\ref{sec:latent-variable-ising}: Generalization to other models}\label{app:latent-variable-ising}

\subsection{Proof of Theorem~\ref{thm:Ising-score-approximation-marginal}}\label{app:Ising-score-approximation-marginal}

\subsubsection*{Approximate the minimizer of the free energy via iterative algorithms}

Once again, we first prove that we can approximately minimize the free energy by implementing a simple iterative algorithm. Recall that 
\begin{align*}
& \hat \bomega_t(\bz) = \argmin_{\bomega \in [-1, 1]^{d+m}} \cF_t^{\mathrm{marginal}}(\bomega; \bz), \\
& \cF_t^{\mathrm{marginal}}(\bomega; \bz) := \Big\{ \sum_{i = 1}^{d+m} -\sh_{\rm bin}(\omega_i) - \frac{1}{2} \< \bomega, \bA \bomega\>  - \frac{\lambda_t}{\sigma_t^2} \< \bz, \bomega_{1:d}\> + \frac{1}{2} \< \bomega, \bK \bomega \>\Big\}. 
\end{align*}
Taking the gradient and the Hessian of $\cF_t^{\mathrm{marginal}}(\bomega; \bz)$, we obtain
\begin{align*}
& \nabla_{\bomega} \cF_t^{\mathrm{marginal}}(\bomega; \bz) = \tanh^{-1} (\bomega) + (\bK -  \bA )\bomega - \frac{\lambda_t}{\sigma_t^2} [\bz; \bzero_m]^{\sT }, \\
&  \nabla^2_{\bomega} \cF_t^{\mathrm{marginal}}(\bomega; \bz) = \diag\{(1 - \omega_i^2)^{-1}\}_{i \in [d + m]} + \bK - \bA. 
\end{align*}
Since $\|\bK - \bA\|_{\op} \leq A < 1$, we can then conclude that $\nabla^2_{\bw} \cF_t^{\mathrm{marginal}}(\bw; \bz) \succeq (1 - A) \id_{d+m}$ for all $\bz \in \R^d$, hence $\cF_t^{\mathrm{marginal}}(\cdot; \bz)$ is strongly-convex for all $\bz \in \R^d$. This further implies that the fixed-point equation below
\begin{align*}
\bomega = \tanh\left(  (\bA - \bK)\bomega + \frac{\lambda_t}{\sigma_t^2} [\bz; \bzero_m]^{\sT } \right)
\end{align*}
has a unique solution. By Lemma~\ref{lem:approximate_TAP_iteration_Ising}, we obtain that if we run the iteration 
\begin{align}\label{eqn:iterative-algorithm-marginal-approximate}
\tilde\bomega^0(\bz) = \bzero, \qquad \tilde \bomega^k(\bz) = f((\bA - \bK)\tilde\bomega^{k - 1}(\bz) + \lambda_t \sigma_t^{-2} [\bz; \bzero_m]^{\sT }), 
\end{align}
where $\|f(\cdot) - \tanh(\cdot)\|_{\infty} \leq \zeta$, then 
\begin{align}\label{eqn:m-approximation-bound-in-proof-Ising-marginal}
\frac{1}{\sqrt{d + m}} \|\tilde \bomega^k(\bz) - \hat\bomega_t(\bz) \|_2 \leq A^k + \zeta (1 - A)^{-1}. 
\end{align}
In particular, we require that $f(\cdot)$ is the function that we construct in Lemma~\ref{lemma:approx-tanh}. 

\subsubsection*{Represent the iterative algorithm as a ResNet}

Recall that $\hat\bbm_t(\bz) = [\hat\bomega_t(\bz)]_{1:d}$. We define $\tilde \bbm^{\ell}(\bz) := [\tilde \bomega^{(\ell)}(\bz)]_{1:d}$. In what follows, we show that $(\lambda_t \tilde \bbm^{\ell}(\bz) - \bz) / \sigma_t^2$ can be expressed as a ResNet that takes input $\bz$.

\begin{lemma}\label{lem:alg-to-net-marginal}
For all $\ell \in \N_+$ and $\delta \leq t \leq T$, there exists $\bW \in \cW_{d, D, \ell, M, B}$ with 
\[
\begin{aligned}
&~D = 3(d + m), ~~~~~M = ( \lceil  2\zeta^{-1} \rceil + 1)(d + m), \\
&~B = (\lceil 2\zeta^{-1} \rceil - 1) \cdot (\log \lceil \zeta^{-1} \rceil + 4) + 8 + \sqrt{d + m} + (1 - e^{-2\delta})^{-1},
\end{aligned}
\]
such that $(\lambda_t \tilde \bbm^{\ell}(\bz) - \bz) / \sigma_t^2 = \resnet_{\bW}(\bz)$, where $\tilde \bbm^\ell$ is as defined in Eq.~(\ref{eqn:iterative-algorithm-marginal-approximate}). 

\end{lemma}
\begin{proof}[Proof of Lemma~\ref{lem:alg-to-net-marginal}]
Recall the definition of $f$ as an approximation of $\tanh$ as in Lemma \ref{lemma:approx-tanh}. The proof of this lemma is similar to that of Lemma~\ref{lem:alg-to-net}. To be specific, we will select the weight matrices $\{\bW_1^{(\ell)}, \bW_2^{(\ell)}, \bW_{\inside}, \bW_{\out}\}$ appropriately such that $\bu^{(\ell)} = [\tilde \bomega^\ell(\bz);  \sigma_t^{-2} [\bz; \bzero_m]^{\sT}; \ones_{d + m}]^{\sT} \in \R^{3(d + m)}$. When $\ell = 0$, this can be achieved by setting 
\begin{align*}
\bW_{\inside} = \left[ \begin{array}{ccc}
\bzero_{d \times (d + m)}  &  \sigma_t^{-2} [\id_d, \bzero_{d \times m}] & \bzero_{d \times (d + m)} \\
\bzero_{1 \times (d + m)}  & \bzero_{1 \times (d + m)} & \ones_{1 \times (d + m)}
\end{array} \right] \in \R^{(d + 1) \times 3(d + m)}. 
\end{align*}
Also, recall that $f(x) = \sum_{j = 1}^{\lceil 2\zeta^{-1} \rceil - 1} \relu (x - w_j) + a_0$. Therefore, for $\ell \in \N_+$, we simply set
\begin{align*}
& \bW_1^{(\ell)} = \left[ \begin{array}{ccccccc}
a_i \id_d  & \cdots & a_{\lceil 2\zeta^{-1} \rceil - 1} \id_{d + m} & -\id_{d + m} & \id_{d + m} & a_0 \id_{d + m} & -a_0 \id_{d + m} \\
\bzero_{{(d + m)} \times {(d + m)}} & \cdots & \bzero_{{(d + m)} \times {(d + m)}} & \bzero_{{(d + m)} \times {(d + m)}} & \bzero_{(d + m) \times (d + m)} & \bzero_{(d + m) \times (d + m)} & \bzero_{(d + m) \times (d + m)} \\
\bzero_{(d + m) \times (d + m)} & \cdots  & \bzero_{(d + m) \times (d + m)} & \bzero_{(d + m) \times (d + m)} & \bzero_{(d + m) \times (d + m)} & \bzero_{(d + m) \times (d + m)} & \bzero_{(d + m) \times (d + m)}
\end{array} \right] \\
& \in \R^{3(d + m) \times (\lceil 2\zeta^{-1} \rceil + 3)(d + m)}, \\
& \bW_2^{(\ell)} = \left[ \begin{array}{ccccccc}
\bA - \bK &  \cdots &  \bA - \bK & \id_{d + m} & -\id_{d + m} &  \bzero_{(d + m) \times (d + m)} &  \bzero_{(d + m) \times (d + m)} \\
\lambda_t \id_{d + m}     &  \cdots &  \lambda_t \id_{d + m} & \bzero_{(d + m) \times (d + m)} &  \bzero_{(d + m) \times (d + m)} &  \bzero_{(d + m) \times (d + m)} &  \bzero_{(d + m) \times (d + m)} \\
-w_1 \id_{d + m} & \cdots &  -w_{\lceil 2\zeta^{-1} \rceil - 1} \id_{d + m} &  \bzero_{(d + m) \times (d + m)} &  \bzero_{(d + m) \times (d + m)} & \id_{d + m} & -\id_{d + m}
\end{array} \right]^{\sT} \\
& \in \R^{(\lceil 2\zeta^{-1} \rceil + 3)(d + m) \times 3(d + m)}. 
\end{align*}
Finally, we take $\bW_{\out} = [\lambda_t \sigma_t^{-2} \id_d, \bzero_{d \times m},  -\id_d, \bzero_{d \times (d + 2m)}] \in \R^{d \times 3(d + m)}$. 

Next, we upper bound the norm of the residual network. By Lemma~\ref{lemma:approx-tanh} and Remark~\ref{rmk:approx-tanh-lem}, we have $\sum_{j = 1}^{ \lceil 2\zeta^{-1} \rceil - 1} |a_j| \leq 2$, $|a_0| \leq 1$, $|w_j| \leq \log \lceil \zeta^{-1} \rceil$. Therefore, 
\begin{align*}
& \|\bW_{\inside}\|_{\op} \leq \sqrt{d + m} + \sigma_t^{-2}, \qquad \|\bW_{\out}\|_{\op} \leq 1 + \lambda_t \sigma_t^{-2}, \\
& \|\bW_1^{(\ell)}\|_{\op} \leq 2\lceil 2\zeta^{-1} \rceil + 2, \qquad \|\bW_2^{(\ell)}\|_{\op} \leq (\lceil 2\zeta^{-1} \rceil - 1) \cdot (\log \lceil \zeta^{-1} \rceil + 2) + 4. 
\end{align*}
This implies that 
\begin{align*}
\nrmps{\bW} \leq B = (\lceil 2\zeta^{-1} \rceil - 1) \cdot (\log \lceil \zeta^{-1} \rceil + 4) + 8 + \sqrt{d + m} + (1 - e^{-2\delta})^{-1}.  
\end{align*}
This completes the proof of Lemma~\ref{lem:alg-to-net-marginal}. 
\end{proof}


\subsubsection*{Proof of Theorem~\ref{thm:Ising-score-approximation-marginal}}

Similar to the proof of Theorem~\ref{thm:Ising-score-approximation}, we obtain
\begin{equation}\label{eqn:error-decomp-Ising-marginal-proof}
\E [\| \hat \bs_t(\bz) - \bs_t(\bz) \|_2^2] / d  \le \bar\eps_{\rm app}^2 + \bar \eps_{\rm gen}^2, 
\end{equation}
where $\bar\eps_{\rm app}^2$ is the approximation error and $\bar \eps_{\rm gen}^2$ is the generalization error, 
\[
\begin{aligned}
\bar\eps_{\rm app}^2 =&~ \inf_{\bW \in \cW} \E[\| \sP_t \ResN_{\bW}(\bz) - \bs_t(\bz)  \|_2^2] / d, \\
\bar \eps_{\rm gen}^2 =&~ 2 \sup_{\bW \in \cW} \Big| \hat \E [ \| \sP_t \ResN_{\bW}(\bz) + \bsigma_t^{-1} \bg \|_2^2] / d - \E [\| \sP_t \ResN_{\bW}(\bz) + \bsigma_t^{-1} \bg \|_2^2] / d \Big|. 
\end{aligned}
\]
By Proposition~\ref{prop:uniform2} and take $D = 3 (d + m)$, with probability at least $1 - \eta$, simultaneously for any $t \in \{ T - t_k\}_{0 \le k \le N-1}$, we have
\begin{align}\label{eqn:gen-error-bound-Ising-marginal-proof}
\bar \eps_{\rm gen}^2
\lesssim \frac{ \lambda_t^2}{\sigma_t^4} \cdot  \sqrt{\frac{[ M L + d ] (d + m) \cdot [ L \cdot \log ( L B ) + \log(\lambda_t^{-1})] + \log (N / \eta)}{n}}. 
\end{align}
To bound $\bar\eps_{\rm app}^2$, by the identity that $\bs_t(\bz) = (\lambda_t \bbm_t(\bz) - \bz) / \sigma_t^2$ and $\sP_t \ResN_{\bW}(\bz) = \proj_{\lambda_t \sigma_t^{-2} \sqrt{d}}(\ResN_{\bW}(\bz) + \sigma_t^{-2} \bz ) - \sigma_t^{-2} \bz$, recalling $\tilde \bbm^L(\bz) = \tilde \bomega^L_{1:d}(\bz)$ as defined in Eq.~(\ref{eqn:iterative-algorithm-marginal-approximate}), and by Lemma~\ref{lem:alg-to-net-marginal}, we have
\begin{equation}\label{eqn:app-error-bound-Ising-marginal-proof}
\begin{aligned}
\bar\eps_{\rm app}^2 =&~ \inf_{\bW \in \cW} \E[\| \sP_t \ResN_{\bW}(\bz) - \bs_t(\bz)  \|_2^2] / d \\
\lesssim&~ \E[\| \proj_{\lambda_t \sigma_t^{-2} \sqrt{d}} ( \lambda_t \sigma_t^{-2} \tilde \bbm^L(\bz) )  - \proj_{\lambda_t \sigma_t^{-2} \sqrt{d}} ( \lambda_t \sigma_t^{-2} \hat \bbm (\bz) ) \|_2^2 ] / d \\
&~ + \E[\| \proj_{\lambda_t \sigma_t^{-2} \sqrt{d}} ( \lambda_t \sigma_t^{-2} \hat \bbm (\bz) ) - \lambda_t \sigma_t^{-2} \bbm_t(\bz) \|_2^2 ] / d. 
\end{aligned}
\end{equation}
By Eq.~(\ref{eqn:m-approximation-bound-in-proof-Ising-marginal}), the $1$-Lipschitzness of $\proj_{\lambda_t \sigma_t^{-2} \sqrt{d}}$, and the definition that $\hat\bbm_t(\bz) = [\hat\bomega_t(\bz)]_{1:d}$ and $\tilde \bbm^{\ell}(\bz) = [\tilde \bomega^{(\ell)}(\bz)]_{1:d}$, the first quantity on the right-hand side is controlled by 
\begin{equation}\label{eqn:app-error-bound-Ising-marginal-proof-first}
\begin{aligned}
&~\E[\| \proj_{\lambda_t \sigma_t^{-2} \sqrt{d}} ( \lambda_t \sigma_t^{-2} \tilde \bbm^L(\bz) )  - \proj_{\lambda_t \sigma_t^{-2} \sqrt{d}} ( \lambda_t \sigma_t^{-2} \hat \bbm (\bz) ) \|_2^2 ] / d \\
\lesssim&~ \frac{d + m}{d} \cdot \frac{\lambda_t^2}{\sigma_t^4} \cdot (A^{2L} + \zeta^2 (1 - A)^{-2})\lesssim \frac{d + m}{d} \cdot \frac{\lambda_t^2}{\sigma_t^4} \cdot \Big(A^{2L} + \frac{(d + m)^2}{(1 - A)^2 M^2}\Big),
\end{aligned}
\end{equation}
where the last inequality is by the fact that we can choose $\zeta$ such that $\zeta \leq  6(d + m) / M$. Furthermore, by Assumption~\ref{ass:Ising-free-energy-marginal} and by $\| \hat \bbm(\bz) \|_2 \le \sqrt{d}$, the second quantity in the right-hand side is controlled by 
\begin{equation}\label{eqn:app-error-bound-Ising-marginal-proof-second}
\E[\| \proj_{\lambda_t \sigma_t^{-2} \sqrt{d}} ( \lambda_t \sigma_t^{-2} \hat \bbm (\bz) ) - \lambda_t \sigma_t^{-2} \bbm_t(\bz) \|_2^2 ] / d \lesssim  \frac{\lambda_t^2}{\sigma_t^4} \cdot \eps_{{\rm VI}, t}^2(\bA).
\end{equation}
Combining Eq.~(\ref{eqn:error-decomp-Ising-marginal-proof}), (\ref{eqn:gen-error-bound-Ising-marginal-proof}),  (\ref{eqn:app-error-bound-Ising-marginal-proof}), (\ref{eqn:app-error-bound-Ising-marginal-proof-first}), (\ref{eqn:app-error-bound-Ising-marginal-proof-second}) completes the proof of the score estimation result in Theorem~\ref{thm:Ising-score-approximation-marginal}. The KL divergence bound is a direct consequence of score estimation error, Theorem~\ref{thm:benton}, and Lemma~\ref{lem:gamma-lambda-sigma-inequality}. This concludes the proof.

\subsection{Proof of Theorem~\ref{thm:Ising-score-approximation-conditional}}\label{app:Ising-score-approximation-conditional}

\subsubsection*{Approximate the minimizer of the free energy via iterative algorithm}

We define
\begin{align*}
\cF_t^{\rm{cond}} (\bbm; \bz, \btheta) := \sum_{i = 1}^{d} -\sh_{\rm bin}(m_i) - \frac{1}{2} \< \bbm, \bA_{11} \bbm \> - \< \bbm, \bA_{12} \btheta\>- \frac{\lambda_t}{\sigma_t^2} \< \bz, \bbm \> + \frac{1}{2} \< \bbm, \bK \bbm \>. 
\end{align*}
Taking the gradient and the Hessian of $\cF_t^{\rm{cond}}$, we obtain
\begin{align*}
& \nabla_{\bbm} \cF_t^{\rm{cond}} (\bbm; \bz, \btheta) = \tanh^{-1} (\bbm) + (\bK - \bA_{11}) \bbm - \bA_{12} \btheta - \frac{\lambda_t}{\sigma_t^2} \bz, \\
& \nabla_{\bbm}^2 \cF_t^{\rm{cond}} (\bbm; \bz, \btheta) = \diag\{((1 - m_i^2)^{-1})_{i \in [d]}\} + \bK - \bA_{11}. 
\end{align*}
When $\|\bK - \bA_{11}\|_{\op} \leq A < 1$, we always have $\nabla_{\bbm}^2 \cF_t^{\rm{cond}} (\bbm; \bz, \btheta) \succeq (1 - A) \id \succ 0$. That is to say, $\cF_t^{\rm{cond}}(\cdot; \bz, \btheta)$ is strongly convex, hence 
\begin{align*}
\bbm =  \tanh\left( (\bA_{11} - \bK) \bbm + \bA_{12} \btheta + \frac{\lambda_t}{\sigma_t^2} \bz \right)
\end{align*}
has a unique solution. We then can apply Lemma~\ref{lem:approximate_TAP_iteration_Ising}, and conclude that if we run iteration
\begin{align}\label{eqn:iterative-algorithm-conditional-approximate}
\tilde \bbm^0(\bz; \btheta) = \bzero, \qquad \tilde \bbm^{\ell}(\bz; \btheta) = f((\bA_{11} - \bK) \tilde \bbm^{\ell - 1}(\bz; \btheta) + \bA_{12} \btheta + \lambda_t \sigma_t^{-2} \bz) 
\end{align}
for some $\|f - \tanh\|_{\infty} \leq \zeta$,  it then holds that
\begin{align}\label{eqn:m-approximation-bound-in-proof-Ising-conditional}
\frac{1}{\sqrt{d}} \|\tilde \bbm^{\ell}(\bz; \btheta) - \hat\bbm_t(\bz; \btheta)\|_2 \leq A^{\ell} + \zeta (1 - A)^{-1}.  
\end{align}
As usual, we require $f(\cdot)$ satisfies all other conditions from Lemma~\ref{lemma:approx-tanh}. 

\subsubsection*{Represent the iterative algorithm as a ResNet}

Next, we show that $(\lambda_t \tilde \bbm^{\ell} (\bz; \btheta) - \bz) / \sigma_t^2$ can be expressed as a ResNet as in \eqref{eqn:relu-resnet-conditional} that has input $(\bz, \btheta)$.

\begin{lemma}\label{lem:alg-to-net-conditional}
For all $\ell \in \N_+$ and $\delta \leq t \leq T$, there exists $\bW \in \cW_{d, m, D, \ell, M, B}$ with 
\[
\begin{aligned}
&~D = 4d, ~~~~~ M = (\lceil 2\zeta^{-1} \rceil + 3)d, \\
&~B = ( \lceil 2 \zeta^{-1} \rceil - 1) \cdot (\log \lceil \zeta^{-1} \rceil + 4 + \|\bA_{12}\|_{\op}) + 8 + (1 - e^{-2\delta})^{-1} + \|\bA_{12}\|_{\op} + \sqrt{d},
\end{aligned}
\]
such that $(\lambda_t \tilde \bbm^{\ell}(\bz; \btheta) - \bz) / \sigma_t^2 = \resnet_{\bW}(\bz, \btheta)$, where $\tilde \bbm^\ell$ is as defined in Eq.~(\ref{eqn:iterative-algorithm-conditional-approximate}). 
\end{lemma}
\begin{proof}[Proof of Lemma~\ref{lem:alg-to-net-conditional}]
Recall the definition of $f$ as an approximation of $\tanh$ as in Lemma \ref{lemma:approx-tanh}. We shall choose the weight matrices such that $\bu^{(\ell)} = [\tilde\bbm^{\ell} (\bz; \btheta);  \sigma_t^{-2} \bz; \bA_{12} \btheta; \ones_d] \in \R^{4d}$. For $\ell = 0$, we simply set 
\begin{align*}
\bW_{\inside} = \left[ \begin{array}{ccc}
\bzero_{d \times d}  & \bzero_{d \times m} & \bzero_{d \times 1}\\
\sigma_t^{-2} \id_d  &  \bzero_{d \times m} & \bzero_{d \times 1} \\
\bzero_{d \times d} & \bA_{12} & \bzero_{d \times 1} \\
\bzero_{d \times d} & \bzero_{d \times m} & \mathbf{1}_{d \times 1}
\end{array} \right] \in \R^{4d \times (d + m + 1)}. 
\end{align*}
For $\ell \geq 1$, we let 
\begin{align*}
& \bW_1^{(\ell)} = \left[ \begin{array}{ccccccc}
a_i \id_d  & \cdots & a_{\lceil 2 \zeta^{-1} \rceil - 1} \id_d & -\id_d & \id_d & a_0 \id_d & -a_0 \id_d \\
\bzero_{d \times d} & \cdots & \bzero_{d \times d} & \bzero_{d \times d} & \bzero_{d \times d} & \bzero_{d \times d} & \bzero_{d \times d} \\
\bzero_{d \times d} & \cdots  & \bzero_{d \times d} & \bzero_{d \times d} & \bzero_{d \times d} & \bzero_{d \times d} & \bzero_{d \times d} \\
\bzero_{d \times d} & \cdots  & \bzero_{d \times d} & \bzero_{d \times d} & \bzero_{d \times d} & \bzero_{d \times d} & \bzero_{d \times d}
\end{array} \right] \in \R^{4d \times (\lceil 2\zeta^{-1} \rceil + 3)d}, \\
& \bW_2^{(\ell)} = \left[ \begin{array}{ccccccc}
\bA_{11} - \bK &  \cdots &  \bA_{11} - \bK & \id_d & -\id_d &  \bzero_{d \times d} &  \bzero_{d \times d} \\
\lambda_t \id_d     &  \cdots &  \lambda_t \id_d & \bzero_{d \times d} &  \bzero_{d \times d} &  \bzero_{d \times d} &  \bzero_{d \times d} \\
\bA_{12} & \cdots & \bA_{12} & \bzero_{d \times d} & \bzero_{d \times d} & \bzero_{d \times d} & \bzero_{d \times d} \\
-w_1 \id_d & \cdots &  -w_{\lceil 2 \zeta^{-1} \rceil - 1} \id_d &  \bzero_{d \times d} &  \bzero_{d \times d} & \id_d & -\id_d
\end{array} \right]^{\sT} \in \R^{(\lceil 2\zeta^{-1} \rceil + 3)d \times 4d}. 
\end{align*}
Finally, we let $\bW_{\out} = [\lambda_t \sigma_t^{-2} \id_d, - \id_d, \bzero_{d \times d}, \bzero_{d \times d}] \in \R^{d \times 4d}$. 
By Lemma~\ref{lemma:approx-tanh} and Remark~\ref{rmk:approx-tanh-lem}, we have $\sum_{j = 1}^{ \lceil 2\zeta^{-1} \rceil - 1} |a_j| \leq 2$, $|a_0| \leq 1$, $|w_j| \leq \log \lceil \zeta^{-1} \rceil$. Therefore, 
\begin{align*}
& \|\bW_{\out}\|_{\op} \leq \lambda_t \sigma_t^{-2} + 1, \qquad \|\bW_{\inside}\|_{\op} \leq \sqrt{d} + \sigma_t^{-2} + \|\bA_{12}\|_{\op}, \\
& \|\bW_1^{(\ell)}\|_{\op} \leq 2\lceil 2 \zeta^{-1} \rceil + 2, \qquad \|\bW_2^{(\ell)}\|_{\op} \leq ( \lceil 2\zeta^{-1} \rceil - 1) \cdot (\log \lceil \zeta^{-1} \rceil + 2 + \|\bA_{12}\|_{\op}) + 4. 
\end{align*}
As a result, we conclude that 
\begin{align*}
\nrmps{\bW} \leq  B = ( \lceil 2 \zeta^{-1} \rceil - 1) \cdot (\log \lceil \zeta^{-1} \rceil + 4 + \|\bA_{12}\|_{\op}) + 8 + (1 - e^{-2\delta})^{-1} + \|\bA_{12}\|_{\op} + \sqrt{d}. 
\end{align*}
We have completed the proof of Lemma~\ref{lem:alg-to-net-conditional}. 
\end{proof}

\subsubsection*{Proof of Theorem~\ref{thm:Ising-score-approximation-conditional}}

Similar to the proof of Theorem~\ref{thm:Ising-score-approximation}, we obtain
\begin{equation}\label{eqn:error-decomp-Ising-conditional-proof}
\E_{\btheta, \bz} [\| \hat \bs_t(\bz;\btheta) - \bs_t(\bz;\btheta) \|_2^2] / d  \le \bar\eps_{\rm app}^2 + \bar \eps_{\rm gen}^2, 
\end{equation}
where $\bar\eps_{\rm app}^2$ is the approximation error and $\bar \eps_{\rm gen}^2$ is the generalization error, 
\[
\begin{aligned}
\bar\eps_{\rm app}^2 =&~ \inf_{\bW \in \cW} \E_{\btheta, \bz}[\| \sP_t [\ResN_{\bW}](\bz, \btheta) - \bs_t(\bz; \btheta)  \|_2^2] / d, \\
\bar \eps_{\rm gen}^2 =&~ 2 \sup_{\bW \in \cW} \Big| \hat \E [ \| \sP_t [\ResN_{\bW}](\bz, \btheta) + \bsigma_t^{-1} \bg \|_2^2] / d - \E_{\btheta, \bz} [\| \sP_t [\ResN_{\bW}](\bz, \btheta) + \bsigma_t^{-1} \bg \|_2^2] / d \Big|. 
\end{aligned}
\]
By Proposition~\ref{prop:uniform2} and take $D = 4d$, with probability at least $1 - \eta$, simultaneously for any $t \in \{ T - t_k\}_{0 \le k \le N-1}$, we have
\begin{align}\label{eqn:gen-error-bound-Ising-conditional-proof}
\bar \eps_{\rm gen}^2
\lesssim \frac{ \lambda_t^2}{\sigma_t^4} \cdot  \sqrt{\frac{( M d L + d (d + m) ) \cdot [ L \cdot \log ( L B d^{-1} (m + d) ) + \log(\lambda_t^{-1})] + \log (N / \eta)}{n}}. 
\end{align}
To bound $\bar\eps_{\rm app}^2$, by the identity that $\bs_t(\bz; \btheta) = (\lambda_t \bbm_t(\bz; \btheta) - \bz) / \sigma_t^2$ and $\sP_t \ResN_{\bW}(\bz, \btheta) = \proj_{\lambda_t \sigma_t^{-2} \sqrt{d}}(\ResN_{\bW}(\bz, \btheta) + \sigma_t^{-2} \bz ) - \sigma_t^{-2} \bz$, recalling $\tilde \bbm^L(\bz)$ as defined in Eq.~(\ref{eqn:iterative-algorithm-conditional-approximate}), and by Lemma~\ref{lem:alg-to-net-conditional}, we have
\begin{equation}\label{eqn:app-error-bound-Ising-conditional-proof}
\begin{aligned}
\bar\eps_{\rm app}^2 =&~ \inf_{\bW \in \cW} \E_{\btheta, \bz}[\| \sP_t [\ResN_{\bW}](\bz, \btheta) - \bs_t(\bz; \btheta)  \|_2^2] / d \\
\lesssim&~ \E_{\btheta, \bz}[\| \proj_{\lambda_t \sigma_t^{-2} \sqrt{d}} ( \lambda_t \sigma_t^{-2} \tilde \bbm^L(\bz; \btheta) )  - \proj_{\lambda_t \sigma_t^{-2} \sqrt{d}} ( \lambda_t \sigma_t^{-2} \hat \bbm (\bz; \btheta) ) \|_2^2 ] / d \\
&~ + \E_{\btheta, \bz}[\| \proj_{\lambda_t \sigma_t^{-2} \sqrt{d}} ( \lambda_t \sigma_t^{-2} \hat \bbm (\bz; \btheta) ) - \lambda_t \sigma_t^{-2} \bbm_t(\bz; \btheta) \|_2^2 ] / d. 
\end{aligned}
\end{equation}
By Eq.~(\ref{eqn:m-approximation-bound-in-proof-Ising-conditional}) and the $1$-Lipschitzness of $\proj_{\lambda_t \sigma_t^{-2} \sqrt{d}}$, the first quantity on the right-hand side is controlled by 
\begin{equation}\label{eqn:app-error-bound-Ising-conditional-proof-first}
\begin{aligned}
&~\E_{\btheta, \bz}[\| \proj_{\lambda_t \sigma_t^{-2} \sqrt{d}} ( \lambda_t \sigma_t^{-2} \tilde \bbm^L(\bz; \btheta) )  - \proj_{\lambda_t \sigma_t^{-2} \sqrt{d}} ( \lambda_t \sigma_t^{-2} \hat \bbm (\bz; \btheta) ) \|_2^2 ] / d \\
\lesssim&~ \frac{\lambda_t^2}{\sigma_t^4} \cdot (A^{2L} + \zeta^2 (1 - A)^{-2})\lesssim  \frac{\lambda_t^2}{\sigma_t^4} \cdot \Big(A^{2L} + \frac{d^2}{(1 - A)^2 M^2}\Big),
\end{aligned}
\end{equation}
where the last inequality is by the fact that we can choose $\zeta$ such that $\zeta \leq  6d / M$. Furthermore, by Assumption~\ref{ass:Ising-free-energy-conditional} and by $\| \hat \bbm(\bz; \btheta) \|_2 \le \sqrt{d}$, the second quantity in the right-hand side is controlled by 
\begin{equation}\label{eqn:app-error-bound-Ising-conditional-proof-second}
\E_{\btheta, \bz}[\| \proj_{\lambda_t \sigma_t^{-2} \sqrt{d}} ( \lambda_t \sigma_t^{-2} \hat \bbm (\bz; \btheta) ) - \lambda_t \sigma_t^{-2} \bbm_t(\bz; \btheta) \|_2^2 ] / d \lesssim  \frac{\lambda_t^2}{\sigma_t^4} \cdot \eps_{{\rm VI}, t}^2(\bA).
\end{equation}
Combining Eq.~(\ref{eqn:error-decomp-Ising-conditional-proof}), (\ref{eqn:gen-error-bound-Ising-conditional-proof}),  (\ref{eqn:app-error-bound-Ising-conditional-proof}), (\ref{eqn:app-error-bound-Ising-conditional-proof-first}), (\ref{eqn:app-error-bound-Ising-conditional-proof-second}) completes the proof of the score estimation result in Theorem~\ref{thm:Ising-score-approximation-conditional}. The KL divergence bound is a direct consequence of score estimation error, Theorem~\ref{thm:benton}, and Lemma~\ref{lem:gamma-lambda-sigma-inequality}. This concludes the proof.

To prove the second result of the bound of the expected KL divergence, we simply notice that by Theorem \ref{thm:benton}, conditioning on every $\btheta$ we have
\begin{align*}
\frac{1}{d} {\rm KL}( \mu_\delta(\cdot | \btheta), \hat \mu( \cdot | \btheta) ) \lesssim \varepsilon^2  + \kappa^2 N + \kappa T + e^{- 2 T}, 
\end{align*}
where 
\begin{align*}
\varepsilon^2 = \frac{1}{d} \sum_{k = 0}^{N - 1} \gamma_k \E \left[ \|\hat\bs_{T - t_k}(\bz; \btheta) - \bs_{T - t_k}(\bz; \btheta)\|_2^2 \mid \btheta \right]. 
\end{align*}
The proof is complete of Theorem~\ref{thm:Ising-score-approximation-conditional} by simply integrating over $\btheta$. 

\subsection{Proof of Theorem~\ref{thm:SC-score-approximation}}\label{app:SC-score-approximation}




\subsubsection*{Relationship of the score function $\bs_t$ to the denoiser $\be_t$}

We first compute the score function $\bs_t(\bz) = \nabla_{\bz} \log \mu_t(\bz)$, for $\bx = \bA \btheta + \beps$ and $\bz = \lambda_t \bx + \sigma_t \bg$, where $\bg \sim \cN(\mathbf{0}_d, \id_d)$ is independent of $(\btheta, \beps) \sim \pi_0^m \otimes \cN(\bzero, \tau^2 \id_d)$. Note that
\begin{align*}
\E[\bx \mid \bz] = &~ \E[\bA \btheta + \beps \mid \lambda_t \bA \btheta +  \lambda_t \beps + \sigma_t \bg] = \bA \E[\btheta \mid \bz] + \E[\beps \mid \lambda_t \bA \btheta +  \lambda_t \beps + \sigma_t \bg] \\
= &~ \bA \E[\btheta \mid \bz] + \frac{\lambda_t \tau^2}{\lambda_t^2 \tau^2 + \sigma_t^2} \E[\lambda_t \beps + \sigma_t \bg \mid \lambda_t \bA \btheta +  \lambda_t \beps + \sigma_t \bg] \\
= &~ \bA \E[\btheta \mid \bz] + \frac{\lambda_t \tau^2}{\lambda_t^2 \tau^2 + \sigma_t^2} \cdot \left( \bz - \lambda_t \bA \E[\btheta \mid \bz]\right) =  \frac{\sigma_t^2}{\lambda_t^2 \tau^2 + \sigma_t^2} \bA \E[\btheta \mid \bz] + \frac{\lambda_t \tau^2}{\lambda_t^2 \tau^2 + \sigma_t^2} \bz. 
\end{align*}
By \eqref{eqn:score-denoiser}, we obtain
\begin{align*}
\bs_t(\bz) = \frac{\lambda_t}{\sigma_t^2} \E[\bx \mid \bz] - \frac{1}{\sigma_t^2} \bz = - \frac{1}{\tau^2 \lambda_t^2 + \sigma_t^2} \bz + \frac{\lambda_t}{\tau^2 \lambda_t^2 + \sigma_t^2} \bA \cdot \E[\btheta \mid \bz].
\end{align*}
We notice the equality in distribution $\bz / \lambda_t \stackrel{d}{=} \bz_{\ast} = \bA \btheta + \bar \beps$ where $(\btheta, \bar \beps ) \sim \pi_0^m \otimes \cN(\bzero, \bar \tau_t^2 \id_d)$ (this $\bz_{\ast}$ is as defined in Assumption~\ref{ass:SC-free-energy}). This implies  
\begin{align}\label{eqn:denoiser-score-SC}
\bs_t(\bz) = - \frac{1}{\tau^2 \lambda_t^2 + \sigma_t^2} \bz + \frac{\lambda_t}{\tau^2 \lambda_t^2 + \sigma_t^2} \bA \cdot \be_t(\bz / \lambda_t),
\end{align}
where $\be_t$ is as defined in Eq.~(\ref{eqn:be-z-ast}). 

\subsubsection*{Existence of a unique minimizer of the VI free energy}
We analyze the VI free energy. We define 
\begin{align*}
\cF_t^{\rm{sparse}} (\be; \bz_{\ast}) :=   \sum_{i = 1}^m \max_{\lambda} \Big[  \lambda e_i - \log \E_{\beta \sim \pi_0}[e^{\lambda \beta - \beta^2 \nu_t / 2}] \Big] + \frac{1}{2 \bar \tau_t^2}\| \bz_{\ast} - \bA \be \|_2^2 -  \frac{1}{2} \< \be, \bK_t \be\>. 
\end{align*}
Let $G_t(\lambda) = \log \E_{\beta \sim \pi_0}[e^{\lambda \beta - {\beta^2 \nu_t} / {2}}]$, and $\lambda_i = \arg\max_{\lambda} [\lambda e_i - G_t(\lambda)]$, then $e_i = G_t'(\lambda_i)$. Therefore, 
\begin{align*}
& \frac{\de}{\de e_i} \left[ \lambda_i e_i - G_t(\lambda_i) \right] = \lambda_i + \frac{e_i}{G_t''(\lambda_i)} - \frac{G_t'(\lambda_i)}{G_t''(\lambda_i)} = \lambda_i, \\
& \frac{\de^2}{\de^2 e_i} \left[ \lambda_i e_i - G_t(\lambda_i) \right] =  \frac{1}{G_t''(\lambda_i)}. 
\end{align*}
Hence, we have
\begin{align*}
& \nabla_{\be} \cF_t^{\rm{sparse}} (\be; \bz_{\ast}) = (G_t')^{-1}(\be) - \frac{1}{\bar \tau_t^2}   \bA^{\sT} \bz_{\ast} + \frac{1}{\bar \tau_t^2}  \bA^{\sT} \bA \be -  \bK_t \be,  \\
& \nabla_{\be}^2 \cF_t^{\rm{sparse}} (\be; \bz_{\ast}) = \diag\{(G_t''(\lambda_i)^{-1})_{i \in [m]}\} + \frac{1}{\bar \tau_t^2} \bA^{\sT} \bA - \bK_t. 
\end{align*}
Note that $G_t''(\lambda_i) = \Var_{(\beta, z) \sim \pi_0 \otimes \cN(0,1)}[\, \beta \mid \beta + \nu_t^{-1/2} z = \lambda \nu_t^{-1}] \leq \Pi^2$. In addition, note that $|G_t'(\lambda)| = |\E[\beta \mid \beta + \nu_t^{-1/2} z = \lambda \nu_t^{-1}]| \leq \Pi$ for all $\lambda$. By assumption, $\|\bar \tau_t^{-2} \bA^{\sT} \bA - \bK_t\|_{\op} < {\Pi^{-2}}$, hence $\nabla_{\be}^2 \cF_t^{\rm{sparse}} (\be; \bz_{\ast})$ is positive-definite and $\cF_t^{\rm{sparse}} (\cdot; \bz_{\ast})$ as a function of $\be$ is strongly convex. That is to say, the equation 
\begin{align*}
\be = G_t'\left( (-\bar \tau_t^{-2} \bA^{\sT} \bA + \bK_t)\be + \bar \tau_t^{-2} \bA^{\sT} \bz_{\ast} \right)
\end{align*}
has a unique fixed point $\hat \be_t(\bz_\ast)$.

\subsubsection*{Approximate the minimizer of the free energy via iterative algorithm}

We denote by $f_t(\cdot)$ the function obtained from Lemma \ref{lemma:approx-tanh} that achieves $\zeta$-uniform approximation to $G_t'(\cdot)$. 
By Lemma \ref{lem:approximate_TAP_iteration_Ising}, we conclude that if we implement the following iteration 
\begin{align}\label{eqn:iterative-alg-SC}
\tilde \be^0 (\bz_{\ast}) = \bzero, \qquad \tilde\be^{\ell + 1}(\bz_{\ast}) = f_t\left( (-\bar \tau_t^{-2} \bA^{\sT} \bA + \bK_t) \tilde\be^\ell (\bz_{\ast}) + \bar \tau_t^{-2} \bA^{\sT} \bz_{\ast} \right), 
\end{align}
%
then for all $\ell \in \N_+$, we have
\begin{align}\label{eqn:m-approximation-bound-in-proof-Ising-4}
\frac{1}{\sqrt{m}} \|\tilde \be^{\ell}(\bz_{\ast}) - \hat\be_t(\bz_{\ast}) \|_2 \leq \Pi \cdot (\Pi^2 A)^{\ell} + \frac{\zeta}{1 - \Pi^2 A}. 
\end{align}

\subsubsection*{Represent the iterative algorithm as a ResNet}

We then show that $\bs_t(\bz) = (\lambda_t \bA \be_t(\bz/\lambda_t) - \bz) / (\tau^2 \lambda_t^2 + \sigma_t^2)$ (c.f. Eq.~(\ref{eqn:denoiser-score-SC})) can be expressed as a ResNet that takes input $\bz$. 

\begin{lemma}\label{lem:alg-to-net-sparse-coding}
For all $t \in \{ T - t_k\}_{0 \le k \le N-1}$ and $\ell \in \N_+$, there exists $\bW \in \cW_{d, D, \ell, M, B}$, with 
\begin{align*}
& D = 3m + d, \qquad M =  (\lceil 2 \Pi \zeta^{-1} \rceil + 3)m, \\
& B = \left( \lceil 2 \Pi \zeta^{-1} \rceil - 1 \right) \cdot \left( A + 1 + 2\Pi^2 + w_{\zeta} \right) + 2\Pi + 6  + (\|\bA\|_{\op} + 1) / (1 - e^{-2\delta}) +\bar \tau_t^{-2} \lambda_t^{-1} \|\bA\|_{\op} + \sqrt{m},
\end{align*}
such that $( \lambda_t \bA \tilde \be^{\ell} (\bz / \lambda_t) -  \bz)  / (\tau^2 \lambda_t^2 + \sigma_t^2) = \resnet_{\bW}(\bz)$. Here, $\tilde \be^{\ell}$ is as defined in Eq.~(\ref{eqn:iterative-alg-SC}), and $w_\zeta$ is given by 
\[
w_\zeta = \sup_{t \in \{ T - t_k\}_{0 \le k \le N-1}} \inf\Big\{w: \mbox{ for all $\lambda_1 > \lambda_2 \geq w$ or $\lambda_1 < \lambda_2 \leq -w$ we have $|G_t'(\lambda_1) - G_t'(\lambda_2)| < \zeta$} \Big\}.
\]
\end{lemma}

\begin{proof}[Proof of Lemma~\ref{lem:alg-to-net-sparse-coding}]
Recall that the ResNet is defined as \eqref{eqn:relu-resnet}. Recall the definition of $f_t$ as an approximation of $G_t'$ as in Lemma \ref{lemma:approx-tanh}. We shall choose the weight matrices appropriately, such that $\bu^{(\ell)} = [\tilde \be^{\ell}( \bz / \lambda_t); \bar \tau_t^{-2}  \bA^{\sT}  \bz / \lambda_t; \ones_m; \bz] \in \R^{3m + d}$. For $\ell = 0$, we set
\begin{align*}
\bW_{\inside} = \left[ \begin{array}{cccc}
\bzero_{d \times m}  & \bar \tau_t^{-2} \lambda_t^{-1} \bA & \bzero_{d \times m} & \id_d \\
\bzero_{1 \times m}  & \bzero_{1 \times m} & \ones_{1 \times m} & \bzero_{1 \times d}
\end{array} \right]^{\sT} \in \R^{(3m + d) \times (d + 1)}. 
\end{align*}
For $\ell \geq 1$, we set 
\begin{align*}
& \bW_1^{(\ell)} = \left[ \begin{array}{ccccccc}
a_i \id_m  & \cdots & a_{\lceil 2 \Pi \zeta^{-1}\rceil - 1} \id_m & -\id_m & \id_m & a_0 \id_m & -a_0 \id_m \\
\bzero_{m \times m} & \cdots & \bzero_{m \times m} & \bzero_{m \times m} & \bzero_{m \times m} & \bzero_{m \times m} & \bzero_{m \times m} \\
\bzero_{m \times m} & \cdots  & \bzero_{m \times m} & \bzero_{m \times m} & \bzero_{m \times m} & \bzero_{m \times m} & \bzero_{m \times m} \\
\bzero_{d \times m} & \cdots  & \bzero_{d \times m} & \bzero_{d \times m} & \bzero_{d \times m} & \bzero_{d \times m} & \bzero_{d \times m}
\end{array} \right] \in \R^{(3m + d) \times (\lceil 2 \Pi \zeta^{-1} \rceil + 3)m}, \\
& \bW_2^{(\ell)} = \left[ \begin{array}{ccccccc}
-\bar \tau_t^{-2} \bA^{\sT} \bA + \bK_t &  \cdots &  -\bar \tau_t^{-2} \bA^{\sT} \bA + \bK_t & \id_m & -\id_m &  \bzero_{m \times m} &  \bzero_{m \times m} \\
\id_m     &  \cdots &  \id_m & \bzero_{m \times m} &  \bzero_{m \times m} &  \bzero_{m \times m} &  \bzero_{m \times m} \\
-w_1 \id_m & \cdots &  -w_{\lceil 2 \Pi \zeta^{-1} \rceil - 1} \id_m &  \bzero_{m \times m} &  \bzero_{m \times m} & \id_m & -\id_m \\
\bzero_{d \times m} & \cdots  & \bzero_{d \times m} & \bzero_{d \times m} & \bzero_{d \times m} & \bzero_{d \times m} & \bzero_{d \times m}
\end{array} \right]^{\sT} \in \R^{(\lceil 2 \Pi \zeta^{-1} \rceil + 3)m \times (3m + d)}. 
\end{align*}
For the output layer, we let $\bW_{\out} = [\lambda_t \bA / (\sigma_t^2 + \tau^2 \lambda_t^2), \bzero_{d \times m}, \bzero_{d \times m}, -(\tau^2 \lambda_t^2 + \sigma_t^2)^{-1} \id_d] \in \R^{d \times (3m + d)}$. 

The following upper bounds are straightforward:
\begin{align*}
& \|\bW_{\inside}\|_{\op} \leq \bar \tau_t^{-2} \lambda_t^{-1} \|\bA\|_{\op} + \sqrt{m} + 1, \qquad \|\bW_{\out}\|_{\op} \leq (\|\bA\|_{\op} + 1) / (1 - e^{-2\delta}), \\
& \|\bW_1^{(\ell)}\|_{\op} \leq 2\Pi^2(\lceil 2\Pi \zeta^{-1} \rceil - 1) +  2\Pi + 2, \qquad \|\bW_2^{(\ell)}\|_{\op} \leq \left( \lceil 2 \Pi \zeta^{-1} \rceil - 1 \right) \cdot \left( A + 1 + w_{\zeta} \right) + 4. 
\end{align*}
In summary, we have 
$$\nrmps{\bW} \leq \left( \lceil 2 \Pi \zeta^{-1} \rceil - 1 \right) \cdot \left( A + 1 + 2\Pi^2 + w_{\zeta} \right) +  2\Pi + 6  + (\|\bA\|_{\op} + 1) / (1 - e^{-2\delta}) + \bar \tau_t^{-2} \lambda_t^{-1} \|\bA\|_{\op} + \sqrt{m}. $$ 
This concludes the proof of Lemma~\ref{lem:alg-to-net-sparse-coding}. 
\end{proof}

\subsubsection*{Proof of Theorem~\ref{thm:SC-score-approximation}}

%
Similar to the proof of Theorem~\ref{thm:Ising-score-approximation}, we obtain
\begin{equation}\label{eqn:error-decomp-Ising-proof-4}
\E [\| \hat \bs_t(\bz) - \bs_t(\bz) \|_2^2] / d  \le \bar\eps_{\rm app}^2 + \bar \eps_{\rm gen}^2, 
\end{equation}
where $\bar\eps_{\rm app}^2$ is the approximation error and $\bar \eps_{\rm gen}^2$ is the generalization error: 
\[
\begin{aligned}
\bar\eps_{\rm app}^2 =&~ \inf_{\bW \in \cW} \E[\| \bar\sP_t [\ResN_{\bW}](\bz) - \bs_t(\bz)  \|_2^2] / d, \\
\bar \eps_{\rm gen}^2 =&~ 2 \sup_{\bW \in \cW} \Big| \hat \E [ \| \bar \sP_t [\ResN_{\bW}](\bz) + \bsigma_t^{-1} \bg \|_2^2] / d - \E [\| \sP_t \ResN_{\bW}(\bz) + \bsigma_t^{-1} \bg \|_2^2] / d \Big|. 
\end{aligned}
\]

Applying Proposition~\ref{prop:uniform3} and taking $D = 3m + d$, we conclude that with probability at least $1 - \eta$, simultaneously for any $t \in \{ T - t_k\}_{0 \le k \le N-1}$, when $n \ge \log(2 / \eta)$, we have
\begin{equation}\label{eqn:gen-error-bound-Ising-proof-4}
\begin{aligned}
\bar \eps_{\rm gen}^2
\lesssim\, &~  \Big( \lambda_t^2\|\bA\|_{\op}^2 \Pi^2  (\tau^{-4} + 1)\frac{m}{d}  + \frac{\lambda_t^2}{\sigma_t^2} (1 + \tau^2) \Big)  \\
&~ \times \sqrt{\frac{(dD + LDM) \cdot \left[  T +  L\log (LB) + \log (n m T  (\tau + 1) (\|\bA\|_{\op} \Pi + 1) \tau^{-1}) \right] + \log (2N/\eta)}{n}}. 
\end{aligned}
\end{equation}
where we choose
\begin{equation}\label{eqn:w-star-definition-in-proof}
\begin{aligned}
B =&~  M / m \cdot \left( A + 1 + 2\Pi^2 + w_\star \right) + 2\Pi + 6  + (\|\bA\|_{\op} + 1) / (1 - e^{-2\delta}) + \tau^{-2} \|\bA\|_{\op} + \sqrt{m},\\
w_\star = &~ \sup_{t \in \{ T - t_k\}_{0 \le k \le N-1}} \inf\Big\{w: \mbox{ for all $\lambda_1 > \lambda_2 \geq w$ or $\lambda_1 < \lambda_2 \leq -w$ we have $|G_t'(\lambda_1) - G_t'(\lambda_2)| < M / (6m \Pi)$ } \Big\}.
\end{aligned}
\end{equation}

We next upper bound $\bar\eps_{\rm app}^2$. 
Recall Eq.~(\ref{eqn:denoiser-score-SC}) and $\bar \tau_t^2 = \tau^2 + \sigma_t^2 / \lambda_t^2$, we have $\bs_t(\bz) = - \lambda_t^{-2}\bar\tau_t^{-2} \bz + {\lambda_t}^{-1}\bar\tau_t^{-2} \bA \bar\be_t(\bz_{\ast})$ (recall that $\bz_{\ast} = \bz / \lambda_t$) and recall $\bar\sP_t [\ResN_{\bW}](\bz) = \proj_{\sqrt{m} \|\bA\|_{\op} \Pi \lambda_t^{-1} \bar\tau_t^{-2} }(\ResN_{\bW}(\bz) + \lambda_t^{-2}\bar\tau_t^{-2} \bz ) - \lambda_t^{-2}\bar\tau_t^{-2} \bz$. According to Lemma~\ref{lem:alg-to-net-sparse-coding}, recalling $\tilde \be^L$ as defined in Eq.~(\ref{eqn:iterative-alg-SC}), we have
\begin{equation}\label{eqn:app-error-bound-Ising-proof-4}
\begin{aligned}
\bar\eps_{\rm app}^2 =&~ \inf_{\bW \in \cW} \E[\| \bar\sP_t [\ResN_{\bW}](\bz) - \bs_t(\bz)  \|_2^2] / d \\
= &~ \inf_{\bW \in \cW} \E[\| \proj_{\sqrt{m} \|\bA\|_{\op} \Pi  \lambda_t^{-1} \bar\tau_t^{-2}} ( \ResN_{\bW}(\bz) + \lambda_t^{-2}\bar\tau_t^{-2} \bz )  - {\lambda_t}^{-1}\bar\tau_t^{-2} \bA \bar\be_t(\bz_{\ast})  \|_2^2] / d \\
\le&~ \E[\| \proj_{\sqrt{m} \|\bA\|_{\op} \Pi  \lambda_t^{-1} \bar\tau_t^{-2}} ( \lambda_t^{-1} \bar\tau_t^{-2} \bA \tilde \be^L(\bz_{\ast}) )  - {\lambda_t}^{-1}\bar\tau_t^{-2} \bA \bar\be_t(\bz_{\ast})  \|_2^2] / d \\
\lesssim&~ \E[\| \proj_{\sqrt{m} \|\bA\|_{\op} \Pi  \lambda_t^{-1} \bar\tau_t^{-2}} ( \lambda_t^{-1} \bar\tau_t^{-2} \bA \tilde \be^L(\bz_{\ast}) )  - \proj_{\sqrt{m} \|\bA\|_{\op} \Pi  \lambda_t^{-1} \bar\tau_t^{-2}} ( \lambda_t^{-1} \bar\tau_t^{-2} \bA \hat \be(\bz_{\ast}) ) \|_2^2 ] / d \\
&~ + \E[\| \proj_{\sqrt{m} \|\bA\|_{\op} \Pi  \lambda_t^{-1} \bar\tau_t^{-2}} ( \lambda_t^{-1} \bar\tau_t^{-2} \bA \hat \be (\bz_{\ast}) ) - \lambda_t^{-1} \bar\tau_t^{-2} \bA \bar\be_t(\bz_{\ast}) \|_2^2 ] / d
\end{aligned}
\end{equation}
where the last inequality is by the triangle inequality. By Eq.~(\ref{eqn:m-approximation-bound-in-proof-Ising-4}) and the $1$-Lipschitzness of $\proj(\cdot)$, we obtain that the first term in the right-hand side above is upper bounded by
\begin{equation}\label{eqn:app-error-bound-Ising-proof-first-4}
\begin{aligned}
&~\E[\| \proj_{\sqrt{m} \|\bA\|_{\op} \Pi  \lambda_t^{-1} \bar\tau_t^{-2}} ( \lambda_t^{-1} \bar\tau_t^{-2} \bA \tilde \be^L(\bz_{\ast}) )  - \proj_{\sqrt{m} \|\bA\|_{\op} \Pi  \lambda_t^{-1} \bar\tau_t^{-2}} ( \lambda_t^{-1} \bar\tau_t^{-2} \bA \hat \be (\bz_{\ast}) ) \|_2^2 ] / d \\
\lesssim&~ \frac{m \|\bA\|_{\op}^2}{d \lambda_t^2 \bar\tau_t^4} \cdot (\Pi^2 \cdot (\Pi^2 A)^{2L} + \zeta^2 (1 - \Pi^2 A)^{-2})\lesssim \frac{m \|\bA\|_{\op}^2}{d \lambda_t^2 \bar\tau_t^4} \cdot \Big(\Pi^2 \cdot (\Pi^2 A)^{2L} + \frac{m^2\Pi^2}{(1 - \Pi^2A)^2 M^2}\Big).
\end{aligned}
\end{equation}
In the above display, the last inequality is by the fact that we can choose $\zeta$ such that $M = m \cdot (\lceil 2 \Pi \zeta^{-1} \rceil + 3)$, which implies that $2m \Pi / M \leq \zeta \leq  6m \Pi / M$. Furthermore, by Assumption~\ref{ass:SC-free-energy} and by the fact that $\| \hat \be(\bz_{\ast}) \|_2 \le \sqrt{m}\,  \Pi$, we obtain that the second quantity in the right-hand side of Eq.~\eqref{eqn:app-error-bound-Ising-proof-4} is controlled by 
\begin{equation}\label{eqn:app-error-bound-Ising-proof-second-4}
\E[\| \proj_{\sqrt{m} \|\bA\|_{\op} \Pi  \lambda_t^{-1} \bar\tau_t^{-2}} ( \lambda_t^{-1} \bar\tau_t^{-2} \bA \hat \be (\bz_{\ast}) ) - \lambda_t^{-1} \bar\tau_t^{-2} \bA \bar\be_t(\bz_{\ast}) \|_2^2 ] / d \lesssim  \frac{m \|\bA\|_{\op}^2}{d\lambda_t^2\bar\tau_t^4} \cdot \eps_{{\rm VI}, t}^2(\bA).
\end{equation}
Finally, we combine Eq.~(\ref{eqn:error-decomp-Ising-proof-4}), (\ref{eqn:gen-error-bound-Ising-proof-4}),  (\ref{eqn:app-error-bound-Ising-proof-4}), (\ref{eqn:app-error-bound-Ising-proof-first-4}), (\ref{eqn:app-error-bound-Ising-proof-second-4}). This completes the proof of Theorem~\ref{thm:SC-score-approximation}.

\subsection{Proof of Lemma~\ref{lem:SC-TAP-consistency}}\label{app:proof-lemma-SC-orthogonal}

Consider the sparse coding problem $\bz_{\ast} = \bA \btheta + \bar \beps \in \R^d$ with dictionary $\bA \in \R^{d \times m}$, sparse representation $\btheta \in \R^m$, and noise $\bar \beps \in \R^d$. Assume that the model satisfies the following assumption. 

\begin{assumption}[Simplified version of Assumption 1 - 4 of \cite{li2023random}]\label{ass:SC-in-proof}
Assume that $\bA = \bQ \bD \bO^\sT$ is the singular value decomposition of $\bA$, where $\bQ \in \R^{d \times d}$ and $\bO \in \R^{m \times m}$ are orthogonal and $\bD \in \R^{d \times m}$ is diagonal with diagonal elements $\{ d_i \}_{i \in [\min\{d, m\}] }$. Assume that $\bQ$, $\bD$ are deterministic, $\bO$, $\btheta$, $\beps$ are mutually independent, and $\bO \sim {\rm Haar}({\rm SO}(m))$ is uniformly distributed on the special orthogonal group. As $d, m \to \infty$, we assume $\mu_{\bD} \stackrel{W}{\to} \sD$ where $\mu_{\bD}$ is the empirical distribution of coordinates of $\bD$, $\sD$ is a random variable with ${\rm supp}\{\sD^2\} \subseteq [d_-, d_+]$ and $0 < d_- < d_+ < \infty$, and $\stackrel{W}{\to}$ denotes Wasserstein-p convergence. Furthermore, $\min_{i}\{ d_i^2 \} \to d_-$ and $\max_{i}\{ d_i^2 \} \to d_+$. We further assume $\theta_i \sim_{iid} \pi_0$ with $\E_{\pi_0}[\theta] = 0$, $\E_{\pi_0}[\theta^2] > 0$, and $\pi_0$ is compactly supported. Finally, we have $\bar \eps_i \sim_{iid} \cN(0, \bar \tau^2)$.
\end{assumption}

Denote the posterior mean of $\btheta$ given $(\bA, \bz_{\ast})$ by $\be(\bz_{\ast}) = \E[\btheta | \bz_{\ast}]$. Theorem 1.11 of \cite{li2023random} proves the following. 

\begin{lemma}[Theorem 1.11 of \cite{li2023random}]\label{lem:SC-corollary-orthogonal-in-proof}
Let Assumption~\ref{ass:SC-in-proof} hold. There exists $\bar \tau_0^2$ that depends on $(\alpha, \pi_0, \sD)$, such that the following happens. For any $\bar \tau^2 \ge \bar \tau_0^2$, there exists $\nu_\star = (\alpha, \pi_0, \sD, \bar \tau^2)$ that depends on $(\alpha, \pi_0, \sD, \bar \tau^2)$ such that, taking $G(\lambda) = \log \E_{\beta \sim \pi_0}[e^{\lambda \beta - {\beta^2 \nu_\star} / {2}}]$ we have almost surely
\[
\lim_{d, m \to \infty} \E_{\bz_{\ast}}\Big[\Big\| \be(\bz_{\ast}) - G'\big( - \bar \tau^{-2} \big( (\bA^\sT \bA - \nu_\star \id_m) \be(\bz_{\ast}) - \bA^\sT \bz_{\ast} \big) \big) \Big\|_2^2  \Big| \bA \Big] = 0. 
\]
Furthermore, for any fixed $(\pi_0, \alpha, \sD)$, we have $\sup_{\bar \tau^2 \ge \bar \tau_0^2} 
\nu_\star(\bar \tau^2) < \infty$. 
\end{lemma}

We remark that Theorem 1.11 of \cite{li2023random} assumes the fixed noise level $\bar \tau^2 = 1$. However, a simple rescaling argument could extend the result to general $\bar \tau^2$.

Given Lemma~\ref{lem:SC-corollary-orthogonal-in-proof}, we are now ready to prove Lemma~\ref{lem:SC-TAP-consistency}. Taking $\bar \tau^2 = \bar \tau_t^2 
= \tau^2 + \sigma_t^2 / \lambda_t^2$, $\nu_t = \nu_\star(\bar \tau_t^2)$, $G_t = G$, and $\bK_t =  \bar \tau_t^{-2} \nu_\star(\bar \tau_t^2)$, we note that the minimizer of the VI free energy $\hat \be_t(\bz_{\ast}) \in [- \Pi, \Pi]^m$ should satisfy 
\[
\hat \be_t(\bz_{\ast}) = G_t'\left( - \bar \tau_t^{-2}\big( (\bA^{\sT} \bA - \nu_t \id_m ) \hat \be_t(\bz_{\ast}) - \bA^{\sT} \bz_{\ast} \big) \right). 
\]
For the posterior mean $\be_t(\bz_{\ast}) \in [-\Pi, \Pi]^m$, we have
\[
\begin{aligned}
&~ \Big\| \be_t(\bz_{\ast}) - G_t'\big( - \bar \tau_t^{-2} \big( (\bA^\sT \bA - \nu_t \id_m) \be_t(\bz_{\ast}) - \bA^\sT \bz_{\ast} \big) \big) \Big\|_2\\
\ge&~ \Big\| \be_t(\bz_{\ast}) - \hat \be_t(\bz_{\ast}) \Big\|_2 - \Big\|  G_t'\big( - \bar \tau_t^{-2} \big( (\bA^\sT \bA - \nu_t \id_m) \be_t(\bz_{\ast}) - \bA^\sT \bz_{\ast} \big) \big) - G_t'\big( - \bar \tau_t^{-2} \big( (\bA^\sT \bA - \nu_tr \id_m) \hat \be_t(\bz_{\ast}) - \bA^\sT \bz_{\ast} \big) \Big\|_2 \\
\ge &~ \Big(1 - \Pi^2 \bar \tau_t^{-2} \| \bA^\sT \bA - \nu_t \id_m  \|_{\op} \Big) \| \be_t(\bz_{\ast}) - \hat \be_t(\bz_{\ast}) \|_2,
\end{aligned}
\]
where the last inequality used the fact that $G_t'$ is $\Pi^2$-Lipschitz. Notice that by Lemma~\ref{lem:SC-corollary-orthogonal-in-proof}, $\sup_{\bar \tau^2 \ge \bar \tau_0^2} 
\nu_\star(\bar \tau^2) = \nu < \infty$, and $\| \bA^\sT \bA \|_{\op} = \max_{i} d_i^2$ bounded almost surely by some $D < \infty$ per Assumption~\ref{ass:SC-in-proof}. Therefore, when $\tau_0^2 \ge 2 \Pi^2 (D + \nu)$, we have $1 - \Pi^2 \bar \tau_t^{-2} \| \bA^\sT \bA - \nu_\star \id_m  \|_{\op} \ge 1/2$ for any $\tau^2 \ge \tau_0^2$ and any $t$. This gives 
\[
\Big\| \be_t(\bz_{\ast}) - G_t'\big( - \bar \tau_t^{-2} \big( (\bA^\sT \bA - \nu_t \id_m) \be_t(\bz_{\ast}) - \bA^\sT \bz_{\ast} \big) \big) \Big\|_2 \ge \| \be_t(\bz_{\ast}) - \hat \be_t(\bz_{\ast}) \|_2 / 2. 
\]
Furthermore, by Lemma~\ref{lem:SC-corollary-orthogonal-in-proof}, the posterior mean $\be_t(\bz_{\ast})$ satisfies
\[
\lim_{d, m \to \infty} \E_{\bz_{\ast}} \Big[\Big\| \be_t(\bz_{\ast}) - G_t'\big( - \bar \tau_t^{-2} \big( (\bA^\sT \bA - \nu_t \id_m) \be_t(\bz_{\ast}) - \bA^\sT \bz_{\ast} \big) \big) \Big\|_2^2  \Big| \bA \Big] = 0. 
\]
This implies that 
\[
\lim_{d, m \to \infty} \E_{\bz_{\ast}} [\| \be_t(\bz_{\ast}) - \hat \be_t(\bz_{\ast}) \|_2^2 | \bA ] = 0, 
\]
which concludes the proof of Lemma~\ref{lem:SC-TAP-consistency}.

\end{document}

%% file: preamble.tex
\usepackage{amsmath}
\usepackage{amsfonts,amscd,amssymb,bm,bbm,mathrsfs}
\usepackage{amsthm}
\usepackage{mathtools}
\usepackage{algorithm}
\usepackage[noend]{algorithmic}
\usepackage{comment}

\usepackage{xcolor}
\usepackage{hyperref}
\hypersetup{
    colorlinks,
    linkcolor={blue!50!black},
    citecolor={blue!50!black},
}
\colorlet{linkequation}{blue}

\def\NGD{\mathsf{NGD}}
\def\relu{\mathrm{ReLU}}
\def\resnet{\mathrm{ResN}}
\def\inside{\mathrm{in}}
\def\out{\mathrm{out}}
\def\score{\mathrm{score}}
\def\iidsim{\sim_{iid}}
\def\N{{\mathbb N}}
\def\Var{{\rm Var}}

\def\proj{{\rm proj}}
\def\sD{{\mathsf D}}
\def\P{{\mathbb P}}

\newtheorem{theorem}{Theorem}
\newtheorem*{theorem*}{Theorem}
\newtheorem{lemma}{Lemma}
\newtheorem{remark}{Remark}
\newtheorem*{remark*}{Remark}
\newtheorem*{lemma*}{Lemma}
\newtheorem{corollary}{Corollary}
\newtheorem{proposition}[theorem]{Proposition}
\newtheorem{definition}{Definition}

\newtheorem{assumption}{Assumption}

\def\argmin{\mathrm{argmin}}
\def\de{{\rm d}}
\def\R{{\mathbb R}}
\def\E{{\mathbb E}}
\def\bzero{{\boldsymbol 0}}
\def\ones{{\boldsymbol 1}}
\def\id{{\mathbf I}}
\def\cN{{\mathcal N}}
\def\sT{{\mathsf T}}
\def\GOE{{\rm GOE}}
\def\Unif{{\rm Unif}}
\def\sh{{\mathsf h}}
\def\ResN{{\rm ResN}}

\def\op{{\rm op}}
\newcommand{\nrmps}[1]
{{|\!|\!|{#1}|\!|\!|}}

\def\eps{{\varepsilon}}
\def\cW{{\mathcal W}}
\newcommand{\what}{\widehat}
\def\diag{{\rm diag}}

\def\Ent{{\rm Ent}}
\def\NN{{\rm NN}}
\def\Cov{{\rm Cov }}
\def\Tr{{\rm Tr}}
\def\AMP{{\sf{AMP}}}

\def\sP{{\mathsf P}}
\def\cP{{\mathcal P}}
\def\cF{{\mathcal F}}
\def\cD{{\mathcal D}}

\def\TAP{{\rm TAP}}
\def\VB{{\rm VB}}
\newcommand{\<}{\langle}
\renewcommand{\>}{\rangle}

\def\cU{{\mathcal U}}
\def\cO{{\mathcal O}}
\def\KL{\mathsf{KL}}

\def\gotop{\stackrel{p}{\longrightarrow}}

\def\bA{\boldsymbol{A}} 
\def\bB{\boldsymbol{B}} 
 
\def\bD{\boldsymbol{D}} 
\def\bE{\boldsymbol{E}} 
 
\def\bG{\boldsymbol{G}}

\def\bJ{\boldsymbol{J}} 
\def\bK{\boldsymbol{K}}

\def\bO{\boldsymbol{O}} 
 
\def\bQ{\boldsymbol{Q}}

\def\bU{\boldsymbol{U}} 
 
\def\bW{\boldsymbol{W}} 
 
\def\bY{\boldsymbol{Y}}

\def\bzeta{\boldsymbol{\zeta}}
\def\bsigma{\boldsymbol{\sigma}}

\def\be{\boldsymbol{e}} 
\def\bmf{\boldsymbol{f}} 
\def\bg{\boldsymbol{g}} 
\def\bh{\boldsymbol{h}}

\def\bbm{\boldsymbol{m}}

\def\bs{\boldsymbol{s}} 
 
\def\bu{\boldsymbol{u}} 
 
\def\bw{\boldsymbol{w}} 
\def\bx{\boldsymbol{x}} 
\def\by{\boldsymbol{y}} 
\def\bz{\boldsymbol{z}} 

\def\btheta{{\boldsymbol \theta}}

\def\bSigma{{\boldsymbol \Sigma}}
\def\bomega{{\boldsymbol \omega}}
\def\beps{{\boldsymbol \eps}}